\documentclass{article}

\usepackage{survey_style}
\usepackage[utf8]{inputenc}
\usepackage[T1]{fontenc}
\usepackage{newunicodechar}
\newunicodechar{±}{\ensuremath{\pm}}
\newunicodechar{×}{\ensuremath{\times}}
\newunicodechar{–}{\textendash}
\newunicodechar{—}{\textemdash}
\newunicodechar{→}{\ensuremath{\rightarrow}}
\newunicodechar{≤}{\ensuremath{\leq}}
\newunicodechar{≥}{\ensuremath{\geq}}

\usepackage[numbers,sort]{natbib}
\usepackage{graphicx}
\usepackage{booktabs}
\usepackage{amsfonts}
\usepackage{amssymb}
\usepackage{amsmath}
\usepackage{amsthm}
\usepackage{microtype}
\usepackage{nicefrac}
\usepackage{xspace}
\usepackage{float}
\usepackage{url}
\usepackage{xurl}
\usepackage{multirow}
\usepackage{subcaption}
\usepackage{makecell}
\usepackage{tabularx}
\usepackage{array}
\usepackage{pifont}
\usepackage[table,xcdraw,dvipsnames]{xcolor}
\usepackage[most]{tcolorbox}
\usepackage{tikz}
\usetikzlibrary{calc,positioning,shadows.blur}
\usepackage[edges]{forest}
\usepackage{enumitem}
\usepackage{caption}
\newcolumntype{P}[1]{>{\raggedright\arraybackslash}p{#1}}
\newcolumntype{Y}{>{\raggedright\arraybackslash}X}
\DeclareCaptionLabelSeparator{pipe}{ | }
\DeclareRobustCommand{\FullCov}{\tikz[baseline=-0.65ex]{\fill (0,0) circle (0.70ex);}}
\DeclareRobustCommand{\PartCov}{\tikz[baseline=-0.65ex]{\begin{scope}\clip (0,-0.70ex) rectangle (0.70ex,0.70ex);\fill (0,0) circle (0.70ex);\end{scope}\draw[line width=0.12ex] (0,0) circle (0.70ex);}}
\DeclareRobustCommand{\NoCov}{\tikz[baseline=-0.65ex]{\draw[line width=0.12ex] (0,0) circle (0.70ex);}}

\usepackage{hyperref}
\usepackage{cleveref}
\hypersetup{
  colorlinks=true,
  linkcolor=blue,
  urlcolor=blue,
  citecolor=blue,
  linkbordercolor=blue,
  urlbordercolor=blue,
  citebordercolor=blue,
  pdfborderstyle={/S/U/W 1},
}

\definecolor{abstractbg}{HTML}{F0F0F0}
\definecolor{abstractborder}{HTML}{D8D8D8}
\definecolor{abstractaccent}{HTML}{4F7957}
\newtcolorbox{abstractbox}{
  enhanced,
  breakable,
  colback=abstractbg,
  colframe=abstractbg,
  boxrule=0pt,
  arc=2.5mm,
  left=3.5mm,
  right=3.5mm,
  top=2.5mm,
  bottom=2.5mm,
  before skip=8pt,
  after skip=12pt
}
\renewenvironment{abstract}
{
  \begin{abstractbox}
  \begin{center}
    {\large\bfseries Abstract}
  \end{center}
  \vspace{-0.25\baselineskip}
}
{
  \par\medskip
  \end{abstractbox}
}

\setlist[itemize]{leftmargin=*}
\setlist[enumerate]{leftmargin=*}
\setlist[description]{leftmargin=*}

\title{Multi-Agent Embodied Autonomous Driving: From V2X Information Exchange to Shared World Models}

\renewcommand{\shorttitle}{MAEAD: From V2X Information Exchange to Shared World Models}

\author{
Senkang Hu\textsuperscript{1,2}, Zhengru Fang\textsuperscript{1,2}, Yihang Tao\textsuperscript{1,2}, Zihan Fang\textsuperscript{1,2}, Yiqin Deng\textsuperscript{3},
\textbf{Yuguang Fang\textsuperscript{1,2}} \\
\textsuperscript{1}Hong Kong JC STEM Lab of Smart City,
\textsuperscript{2}City University of Hong Kong,
\textsuperscript{3}Lingnan University. \\
\texttt{senkang.forest@my.cityu.edu.hk} \\
}

\begin{document}
\maketitle

\newif\ifpreprintvisualabstract
\preprintvisualabstracttrue
\ifpreprintvisualabstract
\begin{center}
  \vspace{-3em}
    \includegraphics[width=1\textwidth]{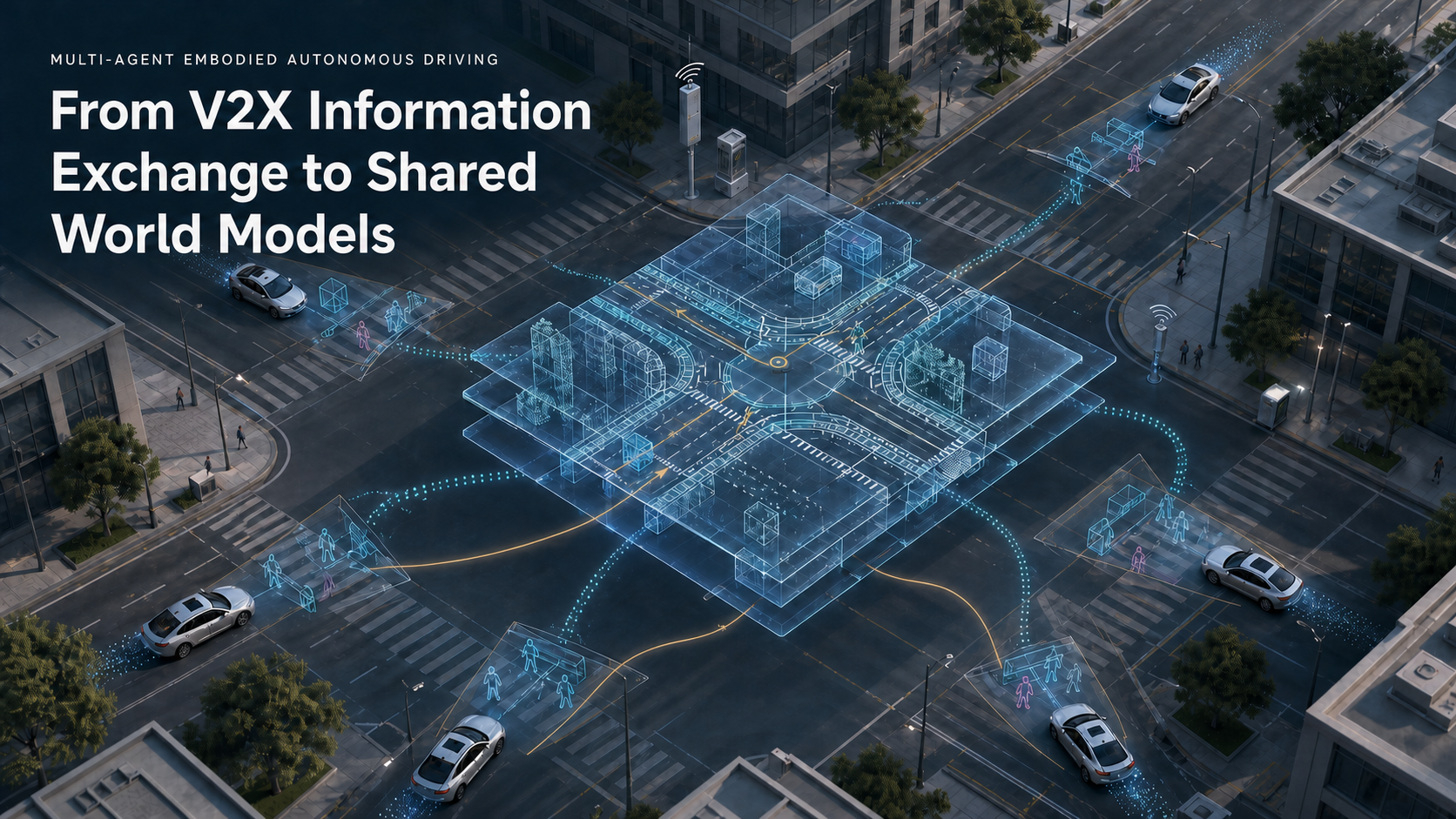}\\[0.4em]
    \parbox{1\textwidth}{%
        \centering\footnotesize
        \textit{Visual abstract.} Vehicles and roadside units exchange complementary observations through V2X links to construct and maintain a shared world model for scene understanding, intent-aware prediction, and coordinated action.%
    }
\end{center}
\fi

\begin{abstract}
Autonomous driving is shifting from isolated vehicle intelligence toward multi-agent embodied systems that share perception, infer intent, and coordinate action under uncertainty. This survey examines this transition through the lens of Shared World Models (SWMs): predictive cross-agent representations maintained across vehicles, infrastructure, and other traffic participants. We review approximately 400 publications covering vehicle-to-everything (V2X) communication, collaborative perception, inter-agent cognition, cooperative planning, end-to-end cooperative driving, and simulation and data engines for closed-loop validation. The organizing question is how exchanged observations become aligned state, intent-aware interaction, and coordinated downstream action. Across the surveyed literature, evaluation remains concentrated in simulation, curated benchmarks, and offline protocols. Foundation-model-based coordination also lacks verifiable real-time safety guarantees in open traffic. These gaps motivate key research priorities for multi-agent embodied autonomous driving (MAEAD): verifiable shared-state maintenance, robust intent and plan alignment, and safe coordinated action under communication and computing constraints in real-world deployment. We maintain an open-source project to continuously track the latest developments at \texttt{\url{https://github.com/dl-m9/Multi-Agent-Embodied-Autonomous-Driving}}.
\end{abstract}


\tableofcontents

\section{Introduction}

\subsection{Background and Motivation}

The pursuit of fully autonomous driving (AD) has historically centered on robust, self-contained vehicles that can perceive, predict, and plan without relying on other road users~\cite{chenMilestonesAutonomousDriving2023, chenEndendAutonomousDriving2024}. This {single-agent paradigm} has enabled major progress in structured settings, yet it remains constrained by partial observability, sensor occlusion, long-range prediction, and dense social interaction~\cite{caillotSurveyCooperativePerception2022, malikCollaborativeAutonomousDriving2021}. A vehicle that reasons only from its own viewpoint can miss hidden hazards, misread collaborative intent, and react too late in tightly coupled traffic. These limitations motivate a shift from isolated autonomy toward collaborative coalition.

The transition to multi-agent systems (MAS) has been accelerated in part by vehicle-to-everything (V2X) communication~\cite{balkusSurveyCollaborativeMachine2022, ansari2021joint}. V2X includes vehicle-to-vehicle (V2V), vehicle-to-infrastructure (V2I), and vehicle-to-pedestrian (V2P) links, allowing vehicles and infrastructure to exchange raw sensor data, processed perception outputs, and high-level intentions~\cite{ansari2021joint}. This cooperation has expanded the perceptual horizon beyond line of sight, leading to collaborative perception (CP) methods based on early, intermediate, and late fusion~\cite{liuVehicletoeverythingAutonomousDriving2023, caillotSurveyCooperativePerception2022}. Representative systems such as V2X-ViT~\cite{xuV2XViTVehicletoEverythingCooperative2022a} and CoBEVT~\cite{xuCoBEVTCooperativeBird2023}, together with datasets such as DAIR-V2X~\cite{yuDAIRV2XLargeScaleDataset2022} and OPV2V~\cite{xuOPV2VOpenBenchmark2022}, show that sharing information across agents can mitigate occlusion and improve scene understanding.

The V2X-centric view is foundational, but it mainly addresses the \textit{data} and \textit{communication} layers. Many systems still treat agents as loosely coupled decision-makers that exchange messages to refine separate local world views. The underlying assumption is that each vehicle maintains its own internal model, while communication acts as a corrective or supplementary service. We refer to this interaction pattern as \emph{Information Exchange}. It is often insufficient for dense and uncertain scenarios that require mutual predictability, coordinated intent, and joint future reasoning~\cite{malikCollaborativeAutonomousDriving2021, liKnowledgedrivenAutonomousDriving2023}. This survey uses \emph{multi-agent embodied autonomous driving (MAEAD)} to indicate the broader setting in which agents must perceive, reason, and act through shared physical and communicative context.

\begin{table}[t]
\centering
\caption{Summary of Related Surveys on Multi-Agent Autonomous Driving.}
\vspace{-0.5em}
\footnotesize
\label{tab:related_surveys}
\resizebox{0.8\textwidth}{!}{
\begin{tabular}{>{\raggedright\arraybackslash}m{3.5cm}>{\raggedright\arraybackslash}m{6cm}*{7}{>{\centering\arraybackslash}m{0.7cm}}}
\toprule
\textbf{Venue / Ref.} & \textbf{Survey Topic} & \textbf{CP} & \textbf{MAD} & \textbf{Comm.} & \textbf{FM} & \textbf{WM} & \textbf{E2E} & \textbf{SWM} \\ \midrule
2022 TITS~\cite{caillotSurveyCooperativePerception2022} & Surveys collaborative perception in automotive settings, including architectures, datasets, and challenges & \FullCov & \NoCov & \PartCov & \NoCov & \NoCov & \NoCov & \NoCov \\ \midrule
2022 COMST~\cite{balkusSurveyCollaborativeMachine2022} & Surveys collaborative machine learning enabled by 5G vehicular communication & \PartCov & \NoCov & \FullCov & \NoCov & \NoCov & \NoCov & \NoCov \\ \midrule
2022 AIS~\cite{dinnewethMultiagentReinforcementLearning2022} & Surveys MARL methods for autonomous vehicles and coordination challenges & \NoCov & \FullCov & \PartCov & \NoCov & \NoCov & \NoCov & \NoCov \\ \midrule
2023 arXiv~\cite{liuVehicletoeverythingAutonomousDriving2023} & Reviews V2X collaborative perception with fusion, bandwidth, robustness, and benchmark analysis & \FullCov & \NoCov & \FullCov & \NoCov & \NoCov & \NoCov & \NoCov \\ \midrule
2024 TITS~\cite{gao2024vehicle} & Reviews vehicle-road-cloud collaborative perception and deployment-oriented system design & \FullCov & \NoCov & \PartCov & \NoCov & \NoCov & \NoCov & \NoCov \\ \midrule
2024 TITS~\cite{bejarbanehExploringSharedPerception2024} & Reviews cooperative vehicle-intersection systems across shared perception, intersection control, and vehicle control & \FullCov & \FullCov & \PartCov & \NoCov & \NoCov & \NoCov & \NoCov \\ \midrule
2024 TIV~\cite{gao2024survey} & Surveys collaborative perception for intelligent vehicles at intersections & \FullCov & \NoCov & \PartCov & \NoCov & \NoCov & \NoCov & \NoCov \\ \midrule
2024 TPAMI~\cite{chenEndendAutonomousDriving2024} & Surveys end-to-end autonomous driving, including world-model-enabled extensions & \NoCov & \PartCov & \NoCov & \PartCov & \PartCov & \FullCov & \NoCov \\ \midrule
2024 WACVW~\cite{cuiSurveyMultimodalLarge2023} & Surveys multimodal LLM tools, datasets, benchmarks, and driving, transportation, and map applications & \PartCov & \PartCov & \NoCov & \FullCov & \NoCov & \PartCov & \NoCov \\ \midrule
2024 TIV~\cite{guanWorldModelsAutonomous2024} & Early survey of world models for autonomous driving & \NoCov & \PartCov & \NoCov & \PartCov & \FullCov & \PartCov & \NoCov \\ \midrule
2025 Proc. IEEE~\cite{huangVehicleEverythingCooperative2025} & Surveys V2X collaborative perception with collaboration models, fusion, alignment, uncertainty, and communication constraints & \FullCov & \PartCov & \FullCov & \NoCov & \NoCov & \NoCov & \NoCov \\ \midrule
2025 Engineering~\cite{zhuSurveyLargeLanguage2025} & Journal survey on large-language-model-powered autonomous driving across perception, reasoning, and planning & \PartCov & \PartCov & \NoCov & \FullCov & \NoCov & \PartCov & \NoCov \\ \midrule
2025 EMNLP Findings~\cite{wuMultiAgentAutonomousDrivingLLM2025} & Surveys large-language-model-based multi-agent autonomous driving systems, interaction modes, human-agent interaction, datasets, and challenges & \PartCov & \FullCov & \FullCov & \FullCov & \NoCov & \PartCov & \NoCov \\ \midrule
2025 ICCVW~\cite{jiangSurveyVisionLanguageAction2025} & Surveys vision-language-action models and unified reasoning-action pipelines for autonomous driving & \PartCov & \PartCov & \NoCov & \FullCov & \PartCov & \FullCov & \NoCov \\ \midrule
2026 Eng. Appl. AI~\cite{fouratiFoundationModelsAutonomous2026} & Surveys foundation models for ADS across model families, driving tasks, benchmarks, deployment, and safety & \PartCov & \PartCov & \PartCov & \FullCov & \PartCov & \PartCov & \NoCov \\ \midrule
\textbf{This work} & \textbf{Surveys MAEAD from V2X information exchange to shared-world-model-driven perception, communication, planning, and validation} & \FullCov & \FullCov & \FullCov & \FullCov & \FullCov & \FullCov & \FullCov \\ \bottomrule
\end{tabular}
}
\vspace{0.3em}

\begin{minipage}{0.95\textwidth}
\footnotesize \textbf{Note:} \FullCov{} denotes fully addressed, \PartCov{} denotes partially covered, and \NoCov{} denotes not addressed. CP: collaborative perception, MAD: multi-agent decision-making, Comm.: communication, FM: foundation models, WM: world models, E2E: end-to-end driving, SWM: shared world model.
\end{minipage}
\vspace{-1em}
\end{table}

\begin{figure}[t]
\centering
\includegraphics[width=0.7\textwidth]{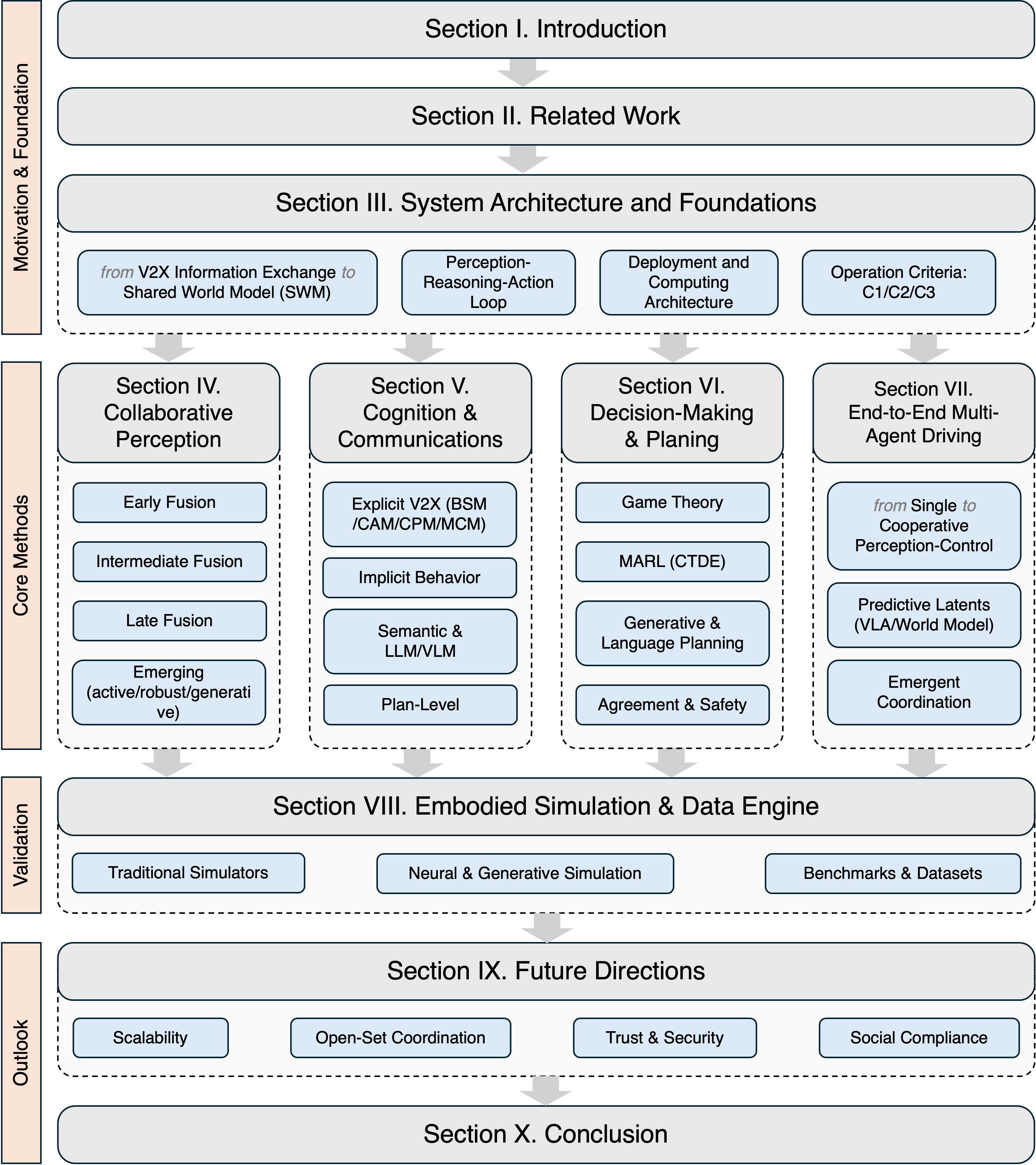}
\caption{Organization of the survey and dependencies among sections. The paper first defines the motivation and evidence scope, then builds architectural foundations, surveys core MAEAD methods, and finally reviews validation engines and future research directions.}
\label{fig:survey_organization}
\vspace{-1em}
\end{figure}

\subsection{From V2X Information Exchange to Shared World Models}

The limitations of \emph{Information Exchange} motivate an embodied view of autonomous driving. In this view, vehicles act in, sense from, and reshape a shared physical environment through a feedback relation among perception, action, interaction, and internal world modeling~\cite{liKnowledgedrivenAutonomousDriving2023, guanWorldModelsAutonomous2024}.

The core distinction is the transition from V2X-style \textit{Information Exchange} to coordination driven by shared world models (SWMs). This survey uses the transition as an organizing lens for a body of work in which multiple autonomous agents continuously interact with each other and with the environment to co-construct a predictive shared representation for collective decision-making and future-state prediction. The lens synthesizes mechanisms already discussed in the literature, including collaborative perception, shared world modeling, multi-agent world models, and language-based reasoning.

This shift requires a generative and predictive process beyond the averaging of local perception outputs. Agents must share \textit{what they see}, \textit{what they believe will happen}, and \textit{how their actions may jointly influence future states}~\cite{gaia2023, gaia2_2025}. This requirement points toward cognitive architectures that connect semantic grounding with predictive vision. Semantic grounding makes scene context, candidate intent, and negotiation constraints explicit. Predictive vision represents future scene evolution and the consequences of coordinated actions.

\emph{Large language models (LLMs)} and \emph{vision-language models (VLMs)} provide one route toward semantic grounding. When grounded in reliable scene representations, they may expose scene context, candidate intentions, negotiation constraints, and plan critiques in a form that downstream planners can inspect~\cite{xuDriveGPT4InterpretableEndtoend2023, simaDriveLMDrivingGraph2023}. Their useful role is to make high-level interpretation and coordination structure explicit while safety-critical control remains grounded in downstream planning and verification. Recent language-grounded driving systems make intent and plan alignment more inspectable, but reported evidence remains simulation-centered, and the latency, reliability, and safety implications of LLM-in-the-loop planning remain open (see Section~\ref{sec:fm-limitations}).

\emph{World models} extend this semantic grounding into predictive vision. They represent future scene evolution, uncertainty, and counterfactual consequences beyond current-scene description. Inspired by human cognition, these models allow agents to learn environmental dynamics, simulate possible outcomes of actions, and reason about other agents' intentions~\cite{guanWorldModelsAutonomous2024, liKnowledgedrivenAutonomousDriving2023}. In the multi-agent embodied context, the SWM is a dynamic representation of the environment that must be updated and harmonized across the fleet. Recent trajectory-generation and driving-simulation models illustrate how future dynamics can be synthesized for prediction and testing~\cite{jiang2023motiondiffuser, zhang2025drivegeninfinitediversetraffic}. These models support predictive SWM construction, while cross-agent alignment and safety-critical coordination remain open.

\subsection{Why This Survey: System-Level Evidence for MAEAD}

Existing surveys have clarified individual layers of collaborative autonomous driving, including V2X communication, collaborative perception, multi-agent decision-making, end-to-end driving, foundation models, and world models~\cite{balkusSurveyCollaborativeMachine2022, caillotSurveyCooperativePerception2022, liuVehicletoeverythingAutonomousDriving2023, chenEndendAutonomousDriving2024, guanWorldModelsAutonomous2024, zhuSurveyLargeLanguage2025}. In MAEAD, these layers become coupled through the cooperative driving loop: communication shapes the shared state, the shared state supports reasoning, and reasoning must affect planning, control, and validation. This survey therefore assesses each line of work by its downstream role in cooperative driving. Perception methods are most relevant when they build planner-usable shared states. Semantic or language-based coordination is most relevant when it is grounded in scene observations, vehicle dynamics, and traffic constraints. Generative models are most relevant when their rollouts support prediction, planning, safety checking, or validation under realistic communication limits.

This survey therefore reviews MAEAD as a system-level evidence chain. The central question is how exchanged messages and observations become shared predictive state, intent-aware interaction, and coordinated downstream behavior. This perspective connects the technical sections: collaborative perception studies shared-state construction, cognition and communication investigate intent and semantic exchange, decision-making and end-to-end driving study action selection, and simulation and data engines define which claims can be tested before deployment.

The review is organized around three evidence questions. First, what mechanisms maintain a shared or aligned representation across agents? Second, how do systems expose future behavior, roles, plans, or negotiation structure? Third, what evidence shows that these representations change planning, control, safety monitoring, or fallback behavior? These questions keep the survey focused on observable mechanisms while leaving formal convergence, deployment certification, and community-standard terminology as separate issues.

\subsection{Paper Roadmap}

The survey follows the transition from V2X Information Exchange to shared world models and then to the evidence criteria defined in Section~\ref{sec:ec-criteria}. Foundation models, validation engines, and open challenges are treated as supporting dimensions. Fig.~\ref{fig:survey_organization} summarizes the survey organization and section dependencies.

The remainder of the paper follows this progression. Section~\ref{sec:related_work} positions prior surveys and evidence families. Section~\ref{sec:architecture} defines the architectural foundations. Sections~\ref{sec:collab-perception} to~\ref{sec:e2e} move from collaborative perception to communication, planning, and end-to-end driving. Section~\ref{sec:simulation} reviews simulation and data engines, and Section~\ref{sec:future} discusses open deployment and trustworthiness gaps.

\section{Related Work}
\label{sec:related_work}

This section positions this paper relative to prior surveys and clarifies the evidence boundary used throughout the paper. Table~\ref{tab:related_surveys} compares representative surveys by the layers they cover and by whether they explicitly organize the field around shared-world-model-driven coordination.

\subsection{Scope of the Review}

This survey is a \emph{structured narrative review}. It is designed to synthesize mechanisms and evidence across the multi-agent autonomous-driving stack, not to perform a meta-analysis or pooled performance comparison. We focus on works that directly contribute to collaborative perception, V2X communication, interaction-aware prediction, cooperative planning, foundation-model-based reasoning, shared world modeling, simulation, benchmarking, and safety analysis. The main coverage period is from January 2019 to May 31, 2026, with earlier papers retained when they provide foundational concepts, standards, simulators, coordination mechanisms, or early autonomous-driving baselines.

Peer-reviewed journal and conference papers are used as the main basis for mature technical claims. Non-peer-reviewed technical sources, project pages, and industry reports are used for mechanism description, standards context, benchmark context, or frontier system examples. Single-vehicle methods and generic multi-agent robotics papers are included when their mechanisms are directly relevant to MAEAD.

\subsection{Evidence Families}

Within this scope, the literature is grouped into six evidence families. The purpose of these families is to support the paper's tutorial narrative from information exchange to shared world models and coordinated behavior. They are not intended as mutually exclusive bins, since many systems combine perception, communication, planning, and simulation components.
\begin{enumerate}
\item \emph{V2X, cooperative driving, and system surveys.}
This cluster covers platooning communication, infrastructure-assisted perception, connected and automated driving, vehicle-to-vehicle messaging, edge support, and broader intelligent transportation surveys~\cite{balador2022platoon, hu2024collab, baiInfrastructureBasedObjectDetection2022, dembaVehicletoVehicleCommunicationTechnology2018, duFederatedLearningVehicular2020, duttaComprehensiveReviewRecent2024, hobertEnhancementsV2XCommunication2015, huCommunicationEfficientCollaborativePerception2024, huPragmaticCommunicationMultiAgent2024, kimVehicletoVehicleV2VMessage2019, liMobilityawareDynamicOffloading2021, chenConnectedAutomatedVehicle2021, dingLearningHelpEmergency2023, liPrivacyPreservedFederatedLearning2022}.

\item \emph{Collaborative perception, fusion, and datasets.}
This cluster covers early collaborative perception systems, LiDAR-camera fusion, bird's eye view (BEV) perception, multi-domain datasets, simulation-to-reality perception transfer, and collaborative sensor fusion~\cite{albertiIDDALargeScaleMultiDomain2020, arrudaCrossDomainCarDetection2019, azfarDeepLearningBasedComputer2024, baiTransFusionRobustLiDARCamera2022, caesarNuScenesMultimodalDataset2020, chenCooperCooperativePerception2019, chenDeepNeuralNetwork2021, chittaTransFuserImitationTransformerBased2022, congSTCrowdMultimodalDataset2022, cuiDeepLearningImage2022, dingDODADataorientedSimtoReal2022, fantauzzoFedDriveGeneralizingFederated2022, geigerAreWeReady2012, guoDeepLearning3d2020, houMultilevelMultimodalFeature2022}.

\item \emph{Decision-making, trajectory prediction, and planning.}
This cluster covers interaction-aware guidance, uncertainty-aware trajectory calibration, overtaking, ramp merging, multi-agent reinforcement learning, cooperative decision-making, and trajectory forecasting~\cite{britoLearningInteractionawareGuidance2021, caoCCTRCalibratingTrajectory2024, chenAutomaticOvertakingTwoway2021, chenCooperativeRampMerging2024, chenDeepMultiagentReinforcement2022, chenLearningAllVehicles2022a, chibLGTrajLLMGuided2024, cuiMultimodalTrajectoryPredictions2019, hangCooperativeDecisionMaking2022, hangDecisionMakingConnected2021, hangDecisionMakingConnected2022, itoCoordinationConnectedVehicles2019, jiaAMPAutoregressiveMotion2024, jiaHDGTHeterogeneousDriving2023, karbalaiealiDynamicAdaptiveAlgorithm2020}.

\item \emph{Foundation models, language grounding, and world models.}
This cluster covers large multimodal models, vision-language-action (VLA)-style transfer, language-assisted driving, embodied multimodal agents, world-model-based simulation, and language-driven reasoning for driving~\cite{abdelnabiLLMDeliberationEvaluatingLLMs2023, baiQwenVLVersatileVisionLanguage2023, brohanRT2VisionLanguageActionModels2023, chenAgentVerseFacilitatingMultiAgent2023, chenAsynchronousLargeLanguage2024, chenDrivingLLMsFusing2023, choudharyTalk2BEVLanguageenhancedBird2023, cuiDriveLLMChartingPath2023, 10491134, dingHolisticAutonomousDriving2024, driessPaLMEEmbodiedMultimodal2023, duanPromptingMultiModalTokens2024, fuDriveHumanRethinking2023, gaia2023, huang2026skillconditionedgatedselfdistillationllm, hu2026selfinducedoutcomepotentialturnlevel, hu2026optimizingagenticreasoningretrieval, zhang2026pelicanunify10unifiedembodied}.

\item \emph{Safety, security, anomaly detection, and adversarial robustness.}
This cluster covers multimodal attacks, vehicular cybersecurity, anomaly detection, LiDAR spoofing, driving fuzzing, security testbeds, and adversarial robustness in cooperative settings~\cite{abdelfattahAdversarialAttacksCameraLiDAR2021, bendiabAutonomousVehiclesSecurity2023, boddupalliREDEMRealTimeDetection2020, boddupalliReplaceRealtimeSecurity2021, bogdollAnomalyDetectionAutonomous2022, bogdollPerceptionDatasetsAnomaly2023, caoAdversarialSensorAttack2019, caoEmergingThreatsDeep2022, caoInvisibleBothCamera2021, choADoPTLiDARSpoofing2023, gaoAutonomousDrivingSecurity2022, hallyburtonMultiAgentSecurityTestbed2024, hallyburtonPartialInformationLongitudinalCyber2023, kimDriveFuzzDiscoveringAutonomous2022, kimPGFUZZPolicyGuidedFuzzing2021}.

\item \emph{Benchmarks, robustness resources, and foundational perception methods.}
This cluster covers autonomous-driving benchmarks, corner-case datasets, robust perception resources, scenario-risk benchmarks, latency-aware collaboration, adversarial multi-agent learning, calibration surveys, and foundational BEV perception methods~\cite{courseyFTAEDBenchmarkDataset2024, gongEdgeIntelligenceIntelligent2023, gongSDACMultimodalSynthetic2024, halderPhysicsBasedRenderingImproving2019, hanADSLeadLifelongAnomaly2023, hanCollaborativePerceptionAutonomous2023, kungRiskBenchScenariobasedBenchmark2024, leiLatencyAwareCollaborativePerception2022a, liAttackingCooperativeMultiAgent2023, liAutomaticTargetlessLiDAR2023, liBEVFormerLearningBird2022, fengIntelligentDrivingIntelligence2021, chenEndendAutonomousDriving2024, liDelvingDevilsBird2024, huPlanningorientedAutonomousDriving2023}.

\end{enumerate}

\begin{figure}[t]
\centering
\resizebox{\textwidth}{!}{%
\begin{tikzpicture}[x=1cm,y=1cm,font=\sffamily]

\fill[black!3,rounded corners=3pt] (2.70,5.45) rectangle (18.00,6.75);
\fill[black!3,rounded corners=3pt] (2.70,4.00) rectangle (18.00,5.30);
\fill[black!3,rounded corners=3pt] (2.70,2.55) rectangle (18.00,3.85);

\node[anchor=east,align=right,font=\sffamily\tiny\bfseries,black!60] at (2.55,8.69) {conceptual\\progression};
\fill[RoyalBlue!6,draw=RoyalBlue!50,rounded corners=3pt] (2.70,8.35) rectangle (7.15,9.02);
\fill[TealBlue!7,draw=TealBlue!55,rounded corners=3pt] (7.55,8.35) rectangle (11.95,9.02);
\draw[Orange!65,dashed,fill=Orange!5,rounded corners=3pt] (12.35,8.35) rectangle (18.00,9.02);
\node[text width=4.0cm,align=center,font=\sffamily\scriptsize\bfseries] at (4.93,8.69) {V2X Data and Feature Exchange};
\node[text width=3.9cm,align=center,font=\sffamily\scriptsize\bfseries] at (9.75,8.69) {Shared State Representations};
\node[text width=5.1cm,align=center,font=\sffamily\scriptsize\bfseries] at (15.18,8.69) {Toward Predictive Shared World Models (SWM)};
\draw[->,RoyalBlue!75,line width=0.8pt] (7.15,8.69) -- (7.55,8.69);
\draw[->,RoyalBlue!75,line width=0.8pt] (11.95,8.69) -- (12.35,8.69);

\node[anchor=east,align=right,font=\sffamily\tiny\bfseries,black!60] at (2.55,7.55) {capability\\stages};
\fill[black!8,rounded corners=3pt] (2.70,6.95) rectangle (5.60,8.15);
\fill[RoyalBlue!10,rounded corners=3pt] (5.75,6.95) rectangle (8.65,8.15);
\fill[TealBlue!12,rounded corners=3pt] (8.80,6.95) rectangle (11.70,8.15);
\fill[Purple!9,rounded corners=3pt] (11.85,6.95) rectangle (14.75,8.15);
\fill[Orange!7,rounded corners=3pt] (14.90,6.95) rectangle (18.00,8.15);

\node[text width=2.55cm,align=center,font=\sffamily\scriptsize\bfseries] at (4.15,7.75) {V2X Message Exchange};
\node[font=\sffamily\scriptsize,black!60] at (4.15,7.18) {2010 to 2019};
\node[text width=2.55cm,align=center,font=\sffamily\scriptsize\bfseries] at (7.20,7.75) {Collaborative Perception};
\node[font=\sffamily\scriptsize,black!60] at (7.20,7.18) {2019 to 2022};
\node[text width=2.60cm,align=center,font=\sffamily\tiny\bfseries] at (10.25,7.76) {Selective, Robust, and Heterogeneous Collaboration};
\node[font=\sffamily\scriptsize,black!60] at (10.25,7.18) {2022 to 2024};
\node[text width=2.55cm,align=center,font=\sffamily\scriptsize\bfseries] at (13.30,7.75) {Cooperative Planning and Intent Alignment};
\node[font=\sffamily\scriptsize,black!60] at (13.30,7.18) {2022 to 2025};
\node[text width=2.75cm,align=center,font=\sffamily\scriptsize\bfseries] at (16.45,7.75) {Integrated and Trustworthy Systems};
\node[font=\sffamily\scriptsize,black!60] at (16.45,7.18) {2025 to 2026};

\node[anchor=east,align=right,font=\sffamily\scriptsize\bfseries] at (2.55,6.10) {Representation\\and Communication};
\node[anchor=east,align=right,font=\sffamily\scriptsize\bfseries] at (2.55,4.65) {Planning and\\Coordinated Action};
\node[anchor=east,align=right,font=\sffamily\scriptsize\bfseries] at (2.55,3.20) {Datasets, Simulation,\\and Deployment};

\node[anchor=west,align=left,text width=2.55cm,font=\sffamily\scriptsize] at (2.85,6.38) {IEEE 802.11p~\cite{ieee80211p2010}};
\node[anchor=west,align=left,text width=2.55cm,font=\sffamily\scriptsize] at (2.85,5.88) {ETSI CAM/DENM~\cite{etsiEN3026372CAM2019,etsiEN3026373DENM2019}};
\node[anchor=west,align=left,text width=2.55cm,font=\sffamily\scriptsize] at (5.90,6.35) {F-Cooper~\cite{chenFcooperFeatureBased2019}};
\node[anchor=west,align=left,text width=2.55cm,font=\sffamily\scriptsize] at (5.90,5.85) {V2X-ViT~\cite{xuV2XViTVehicletoEverythingCooperative2022a}};
\node[anchor=west,align=left,text width=2.55cm,font=\sffamily\scriptsize] at (8.95,6.50) {Where2comm~\cite{huWhere2commCommunicationefficientCollaborative2024}};
\node[anchor=west,align=left,text width=2.55cm,font=\sffamily\scriptsize] at (8.95,6.10) {ROBOSAC~\cite{liUsAdversariallyRobust2023}};
\node[anchor=west,align=left,text width=2.55cm,font=\sffamily\scriptsize] at (8.95,5.70) {HEAL~\cite{luExtensibleFrameworkOpen2024}};
\node[anchor=west,align=left,text width=2.75cm,font=\sffamily\scriptsize] at (15.05,6.35) {CP-Guard~\cite{huCPGuardMaliciousAgent2024}};
\node[anchor=west,align=left,text width=2.75cm,font=\sffamily\scriptsize] at (15.05,5.85) {GCP~\cite{taoGCPGuardedCollaborative2025}};

\node[anchor=west,align=left,text width=2.55cm,font=\sffamily\scriptsize] at (2.85,4.65) {PPBC~\cite{itoCoordinationConnectedVehicles2019}};
\node[anchor=west,align=left,text width=2.55cm,font=\sffamily\scriptsize] at (5.90,4.65) {Virtual Rotation~\cite{chenConnectedAutomatedVehicle2021}};
\node[anchor=west,align=left,text width=2.55cm,font=\sffamily\scriptsize] at (8.95,4.65) {IA-DMARL~\cite{wu2023intent}};
\node[anchor=west,align=left,text width=2.55cm,font=\sffamily\scriptsize] at (12.00,5.02) {COOPERNAUT~\cite{coopernaut2022}};
\node[anchor=west,align=left,text width=2.55cm,font=\sffamily\scriptsize] at (12.00,4.65) {MAPPO-PIS~\cite{guo2024mappo}};
\node[anchor=west,align=left,text width=2.55cm,font=\sffamily\scriptsize] at (12.00,4.28) {CoLMDriver~\cite{liu2025colmdriver}};
\node[anchor=west,align=left,text width=2.75cm,font=\sffamily\scriptsize] at (15.05,4.65) {UniV2X~\cite{yuEndtoEndAutonomousDriving2024}};

\node[anchor=west,align=left,text width=2.55cm,font=\sffamily\scriptsize] at (2.85,3.20) {CARLA~\cite{dosovitskiyCARLAOpenUrban2017}};
\node[anchor=west,align=left,text width=2.55cm,font=\sffamily\scriptsize] at (5.90,3.50) {OPV2V~\cite{xuOPV2VOpenBenchmark2022}};
\node[anchor=west,align=left,text width=2.55cm,font=\sffamily\scriptsize] at (5.90,2.90) {DAIR-V2X~\cite{yuDAIRV2XLargeScaleDataset2022}};
\node[anchor=west,align=left,text width=2.55cm,font=\sffamily\scriptsize] at (8.95,3.20) {V2V4Real~\cite{xuV2V4RealRealWorldLargeScale2023}};
\node[anchor=west,align=left,text width=2.55cm,font=\sffamily\scriptsize] at (12.00,3.20) {V2X-Seq~\cite{v2x-seq}};
\node[anchor=west,align=left,text width=2.75cm,font=\sffamily\scriptsize] at (15.05,3.20) {V2Xverse~\cite{10979246}};

\end{tikzpicture}%
}
\caption{Capability roadmap of multi-agent embodied autonomous driving. The top row traces the survey-level conceptual progression from V2X data and feature exchange, through shared-state representations, toward predictive shared world models. Below, representative milestones are arranged in five nonoverlapping capability stages across three technical tracks: representation and communication, planning and coordinated action, and datasets, simulation, and deployment.}
\label{fig:maead_roadmap}
\vspace{-1em}
\end{figure}
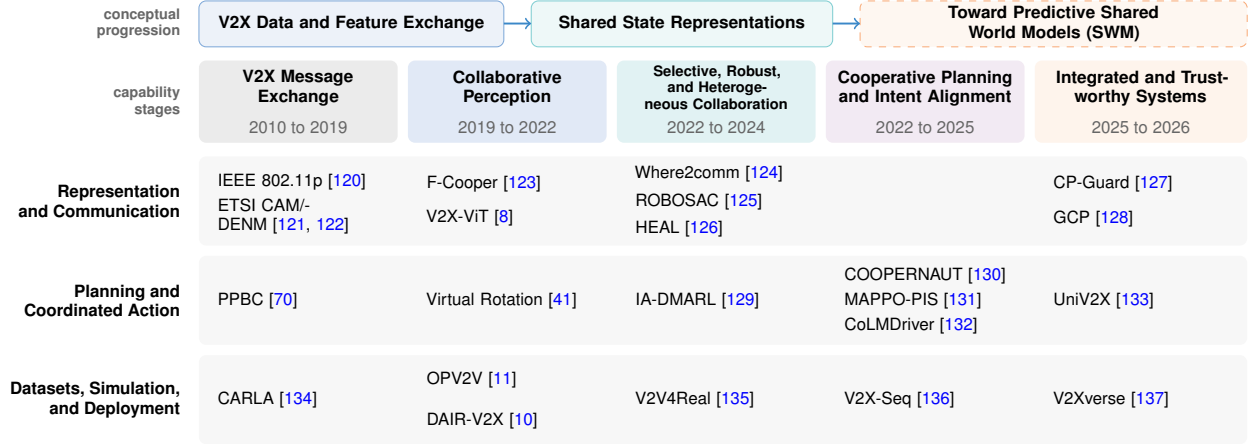

\subsection{Surveys on Collaborative Perception and V2X}

Published surveys primarily focus on the perception and communication layers. Caillot et al.~\cite{caillotSurveyCooperativePerception2022} provide an automotive-context survey of collaborative perception architectures, datasets, and deployment bottlenecks. Liu et al.~\cite{liuVehicletoeverythingAutonomousDriving2023} narrow the focus to V2X collaborative perception and analyze fusion stage, bandwidth-latency trade-offs, robustness to pose and communication noise, and benchmark design. Gao et al.~\cite{gao2024survey, gao2024vehicle} extend this line from two complementary deployment views: intelligent intersections and vehicle-road-cloud collaboration.

The communication substrate and system-level deployment perspective are covered by Balkus et al.~\cite{balkusSurveyCollaborativeMachine2022}, who survey collaborative machine learning over 5G vehicular communications, and Ji et al.~\cite{ji2024coop}, who review cooperative vehicle-infrastructure systems from perception to control. Collectively, these works give substantial coverage of sensing, communication constraints, and deployment settings. Their main limitation for our purposes is that they largely stop at perception-centric or communication-centric analysis, without treating downstream multi-agent reasoning, negotiation, and shared world models as first-class design objectives.

\subsection{Surveys on Multi-Agent Decision-Making and Planning}

Review coverage becomes more fragmented once the focus shifts from sensing to coordination. Malik et al.~\cite{malikCollaborativeAutonomousDriving2021} review collaborative autonomous driving from a broader systems perspective, while Gholamhosseinian and Seitz~\cite{gholamhosseinianComprehensiveSurveyCooperative2022} specialize in cooperative intersection management under heterogeneous connected-vehicle settings. On learning-based coordination, Dinneweth et al.~\cite{dinnewethMultiagentReinforcementLearning2022} survey multi-agent reinforcement learning (MARL) for autonomous vehicles, and Liu et al.~\cite{liu2022graph} review graph reinforcement learning for cooperative decision-making in mixed autonomy traffic.

These surveys clarify core coordination issues such as centralized-training-decentralized-execution, credit assignment, non-stationarity, graph-structured interactions, and intersection negotiation. However, they typically do not jointly cover collaborative perception, V2X communication, and foundation-model-based cognition under a single architectural framework.

\subsection{Surveys on Foundation Models and Emerging Technologies}

Foundation-model-oriented surveys are newer and more fragmented. LLM4Drive~\cite{yangLLM4DriveSurveyLarge2023} and the journal survey by Zhu et al.~\cite{zhuSurveyLargeLanguage2025} summarize LLM-powered driving stacks across perception, reasoning, planning, and interaction. Cui et al.~\cite{cuiSurveyMultimodalLarge2023} and Jiang et al.~\cite{jiangSurveyVisionLanguageAction2025} further extend the discussion to multimodal LLMs and vision-language-action (VLA) models, reflecting the shift from language-assisted modules to unified reasoning-action pipelines. Wu et al.~\cite{wuMultiAgentAutonomousDrivingLLM2025} focus specifically on LLM-based multi-agent autonomous driving systems (ADS), including interaction modes, human-agent interaction, resources, and challenges.

Some related surveys also cover emerging abstractions. Guan et al.~\cite{guanWorldModelsAutonomous2024} survey world models for autonomous driving, Chen et al.~\cite{chenEndendAutonomousDriving2024} review end-to-end driving pipelines, Li et al.~\cite{liKnowledgedrivenAutonomousDriving2023} survey knowledge-driven autonomous driving more broadly, and Zhao et al.~\cite{zhao2024sim} review frameworks and simulators. These works are important for understanding recent trends, but they remain largely single-agent, stack-specific, or tooling-centric. They have not jointly analyzed collaborative perception, V2X communication, negotiation, and coordinated downstream action under a unified multi-agent lens.

\subsection{Our Contributions and Distinctions}

Our survey is positioned as an integrated review of multi-agent embodied autonomous driving. As summarized in Table~\ref{tab:related_surveys}, prior surveys typically emphasize one or two layers of the system stack. Our goal is to examine how these layers interact in a single architectural narrative. The contributions of our survey are summarized as follows.

\begin{enumerate}
\item \emph{Holistic Integration.}
We provide an integrated review that jointly discusses collaborative perception, multi-agent decision-making, communication, computing architectures, deployment patterns, foundation models, world models, end-to-end approaches, and validation infrastructure in a unified framework. While existing surveys typically focus on one or two of these dimensions, our survey emphasizes how these components interact with and constrain each other at the system level.

\item \emph{SWM-driven coordination lens.}
We use the SWM-driven coordination lens as a paper-specific way to interpret the shift from traditional Information Exchange toward jointly maintained, predictive shared world models. We operationalize the lens through observable mechanisms such as cross-agent state alignment, intent and plan alignment, and coordinated downstream action.

\item \emph{Recent and Emerging Directions.}
We place a strong emphasis on recent technologies, including foundation models (LLMs and VLMs), world models, generative simulation, and emergent coordination behaviors. We connect mature building blocks with recently reported technical directions and state evidence boundaries for language-based negotiation, knowledge-pooling systems, emergent social driving behaviors, and compositional world models for multi-agent coordination.
\end{enumerate}

By providing this integrated overview, our survey aims to serve as a useful resource for researchers and practitioners working to advance multi-agent embodied autonomous driving.

\section{System Architecture and Foundations}
\label{sec:architecture}

Multi-agent embodied autonomous driving (MAEAD) can be viewed as an embodied multi-agent cyber-physical system that integrates perception, communication, computing, world modeling, and coordinated decision-making. This section bridges prior surveys and the technical sections. It establishes the shared vocabulary used later: V2X information exchange, shared world model (SWM)-driven coordination, a perception, reasoning, and action loop, deployment and computation topologies, and operational evidence criteria. The purpose is to compare perception, communication, planning, end-to-end driving, and simulation testbeds with a common language and conservative evidence rules. Fig.~\ref{fig:maead_roadmap} situates representative milestones along the capability progression from V2X exchange toward predictive shared-state coordination.

\subsection{From V2X Information Exchange to Shared World Models}

The evolution of multi-agent autonomous driving can be interpreted as a transition from simple \emph{Information Exchange} toward a more interactive model of SWM-driven coordination. We use the \emph{SWM-driven coordination lens} as a specific way to describe this transition, naming the agenda of moving beyond data sharing toward aligned predictive state and coordinated action. Section~\ref{sec:ec-criteria} operationalizes this lens as a set of observable evidence criteria.

\subsubsection{The V2X Information Exchange Paradigm}

The traditional V2X paradigm treats multi-agent driving as a networked communication and information-exchange substrate. Its primary role is to support message exchange among vehicles, infrastructure, pedestrians, and cloud or edge services, including safety beacons, kinematic states, sensor features, object lists, roadside observations, and maneuver intent. In collaborative perception, these exchanges can extend an agent's effective perceptual range and mitigate occlusion, while V2X also supports planning, traffic coordination, and safety services. This communication substrate faces three limits. First, {spectrum bandwidth and scalability}: raw or dense feature exchange can exceed practical communication budgets in crowded scenes. Second, {semantic poverty}: a shared box or velocity vector may not communicate intent, social context, or a planned maneuver. Third, {protocol rigidity}: fixed message formats can struggle with rare scenarios, heterogeneous agents, and degraded links.

\subsubsection{The SWM-Oriented Coordination View}

In contrast, the SWM-oriented coordination view treats agents as active participants in an interactive process. Agents may query, negotiate, and selectively share information according to context and collective need. Communication shifts from raw data toward semantic content, including intentions, goals, and beliefs. The central objective is to co-construct and maintain an SWM, a dynamic and probabilistic representation of the environment and its agents that supports collective decision-making.

This shift changes both communication structure and coordination representation, moving from loosely coupled local decisions toward aligned joint reasoning and actions.

\subsection{A Multi-Agent Perception, Reasoning, and Action Loop}

To frame the architecture, we use a compact perception, reasoning, and action loop. First, agents collect local observations, fuse sensory evidence, coordinate coverage, and query peers to reduce collective uncertainty. This stage determines what evidence enters the cooperative system and which blind spots remain. Second, fused observations update SWM, which represents agent states, map context, occluded regions, likely goals, and anticipated actions. SWM therefore acts as the shared representational workspace between communication and planning. Third, aligned SWM supports future-intent reasoning, plan alignment, and coordinated planning, possibly assisted by foundation models for high-level scene interpretation or proposal generation. Finally, agents execute coordinated maneuvers and feed outcomes back into the next sensing cycle, where prediction errors, communication failures, and new interactions update the shared representation. The loop highlights the recurrent nature of MAEAD: communication shapes shared state, shared state shapes reasoning, reasoning shapes action, and action creates the next round of evidence. The functional loop depends on where sensing, fusion, inference, planning, and control are executed. The next subsection maps these functions to recurring deployment patterns.

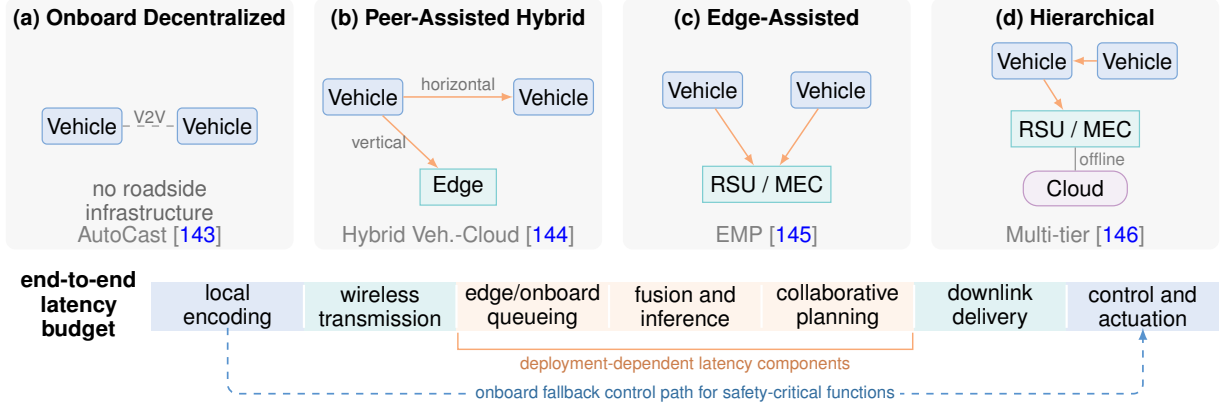
\begin{figure}[t]
\centering
\resizebox{\textwidth}{!}{%
\begin{tikzpicture}[
    x=1cm,y=1cm,font=\sffamily,
    veh/.style={rounded corners=2pt,draw=RoyalBlue!50,fill=RoyalBlue!10,line width=0.5pt,
                minimum width=1.15cm,minimum height=0.52cm,inner sep=0pt,
                font=\sffamily\footnotesize,align=center},
    infra/.style={draw=TealBlue!55,fill=TealBlue!10,line width=0.5pt,
                  minimum width=1.10cm,minimum height=0.52cm,inner sep=0pt,
                  font=\sffamily\footnotesize,align=center},
    cld/.style={rounded corners=6pt,draw=Purple!50,fill=Purple!7,line width=0.5pt,
                minimum width=1.15cm,minimum height=0.50cm,inner sep=0pt,
                font=\sffamily\footnotesize,align=center},
    link/.style={-,draw=black!40,line width=0.55pt},
    off/.style={->,>=latex,draw=Orange!65,line width=0.62pt},
    seg/.style={draw=none,minimum height=0.70cm,inner sep=0pt}
]

\path[use as bounding box] (0,1.04) rectangle (18,7.35);

\foreach \x in {0.25,4.70,9.15,13.60}
    {\fill[black!3,rounded corners=4pt] (\x+0.05,3.42) rectangle (\x+4.20,7.05);}

\node[font=\sffamily\footnotesize\bfseries] at (2.375,6.74) {(a) Onboard Decentralized};
\node[veh,fill=RoyalBlue!10] (a1) at (1.40,5.20) {Vehicle};
\node[veh,fill=RoyalBlue!10] (a2) at (3.35,5.20) {Vehicle};
\draw[link,dashed] (a1) -- node[above,font=\sffamily\scriptsize,black!55,inner sep=1pt] {V2V} (a2);
\node[font=\sffamily\footnotesize,black!60,align=center] at (2.375,4.13)
    {no roadside\\infrastructure};
\node[font=\sffamily\footnotesize,black!50] at (2.375,3.61) {AutoCast~\cite{qiuAutoCastScalableInfrastructureless2022}};

\node[font=\sffamily\footnotesize\bfseries] at (6.825,6.74) {(b) Peer-Assisted Hybrid};
\node[veh,fill=RoyalBlue!10] (b1) at (5.45,5.62) {Vehicle};
\node[veh] (b2) at (8.20,5.62) {Vehicle};
\draw[off] (b1) -- node[above=1.5pt,font=\sffamily\scriptsize,black!55,inner sep=1pt] {horizontal} (b2);
\node[infra] (b3) at (6.825,4.32) {Edge};
\draw[off] (b1) -- node[left,font=\sffamily\scriptsize,black!55,inner sep=1.5pt] {vertical} (b3);
\node[font=\sffamily\footnotesize,black!50] at (6.825,3.61) {Hybrid Veh.-Cloud~\cite{krijestoracHybridVehicularCloud2020}};

\node[font=\sffamily\footnotesize\bfseries] at (11.275,6.74) {(c) Edge-Assisted};
\node[veh] (c1) at (10.35,5.72) {Vehicle};
\node[veh] (c2) at (12.20,5.72) {Vehicle};
\node[infra,minimum width=1.85cm] (c3) at (11.275,4.36) {RSU / MEC};
\draw[off] (c1) -- (c3);
\draw[off] (c2) -- (c3);
\node[font=\sffamily\footnotesize,black!50] at (11.275,3.61) {EMP~\cite{zhangEMPEdgeassistedMultivehicle2021}};

\node[font=\sffamily\footnotesize\bfseries] at (15.725,6.74) {(d) Hierarchical};
\node[veh,fill=RoyalBlue!10] (d1) at (15.10,6.14) {Vehicle};
\node[veh] (d2) at (16.62,6.14) {Vehicle};
\draw[off] (d2) -- (d1);
\node[infra,fill=TealBlue!10,minimum width=1.85cm] (d3) at (15.725,5.16) {RSU / MEC};
\node[cld,minimum width=1.5cm] (d4) at (15.725,4.30) {Cloud};
\draw[off] (d1) -- (d3);
\draw[link] (d3) -- node[right,font=\sffamily\scriptsize,black!50,inner sep=1.5pt] {offline} (d4);
\node[font=\sffamily\footnotesize,black!50] at (15.725,3.61) {Multi-tier~\cite{xiaoPerceptionTaskOffloading2023}};

\node[align=center,text width=1.85cm,inner sep=0pt,font=\sffamily\footnotesize\bfseries] at (1.35,2.60)
    {end-to-end\\latency budget};

\node[seg,fill=RoyalBlue!10,minimum width=2.20cm,align=center,font=\sffamily\footnotesize] at (3.50,2.60) {local\\encoding};
\node[seg,fill=TealBlue!12,minimum width=2.20cm,align=center,font=\sffamily\footnotesize] at (5.70,2.60) {wireless\\transmission};
\node[seg,fill=Orange!7,minimum width=2.20cm,align=center,font=\sffamily\footnotesize] at (7.90,2.60) {edge/onboard\\queueing};
\node[seg,fill=Orange!7,minimum width=2.20cm,align=center,font=\sffamily\footnotesize] at (10.10,2.60) {fusion and\\inference};
\node[seg,fill=Orange!7,minimum width=2.20cm,align=center,font=\sffamily\footnotesize] at (12.30,2.60) {collaborative\\planning};
\node[seg,fill=TealBlue!12,minimum width=2.20cm,align=center,font=\sffamily\footnotesize] at (14.50,2.60) {downlink\\delivery};
\node[seg,fill=RoyalBlue!10,minimum width=2.20cm,align=center,font=\sffamily\footnotesize] at (16.70,2.60) {control and\\actuation};
\foreach \x in {4.60,6.80,9.00,11.20,13.40,15.60}
    {\draw[white,line width=0.9pt] (\x,2.29) -- (\x,2.91);}
\draw[Orange!65,line width=0.6pt] (6.82,2.22) -- (6.82,1.98) -- (13.38,1.98) -- (13.38,2.22);
\node[font=\sffamily\scriptsize,Orange!80!black] at (10.10,1.76) {deployment-dependent latency components};

\draw[->,>=latex,draw=RoyalBlue!65,line width=0.7pt,dashed,rounded corners=4pt]
    (3.50,2.29) -- (3.50,1.34) -- (16.70,1.34) -- (16.70,2.29);
\node[fill=white,inner sep=2pt,font=\sffamily\scriptsize,text=RoyalBlue!72!black] at (10.10,1.34) {onboard fallback control path for safety-critical functions};

\end{tikzpicture}%
}\vspace{-1em}

\caption{Deployment patterns and end-to-end latency architecture for MAEAD. The upper panels compare onboard decentralized, peer-assisted, edge-assisted, and hierarchical vehicle-edge-cloud computing. The lower band illustrates a schematic end-to-end latency budget. Orange marks the deployment-dependent latency components associated with queueing, inference, and planning. The dashed path denotes an onboard fallback control path for safety-critical functions.}
\vspace{-1em}
\label{fig:deployment_computing}
\end{figure}

\subsection{Deployment and Computing Architecture}
\label{sec:deployment-computing}

MAEAD computation can be described through four recurring deployment patterns. Fig.~\ref{fig:deployment_computing} summarizes these patterns and the associated end-to-end latency path. First, onboard decentralized computing keeps local encoding, inference, planning, and control on each vehicle. AutoCast~\cite{qiuAutoCastScalableInfrastructureless2022} demonstrates that direct V2V exchange and local processing can support collaborative perception and planning without roadside infrastructure. Second, peer-assisted vehicular computing allows nearby vehicles to contribute spare processing capacity. Hybrid Vehicular and Cloud Distributed Computing~\cite{krijestoracHybridVehicularCloud2020} combines horizontal offloading to neighboring vehicles with vertical offloading to an edge server. Third, edge-assisted computing assigns regional aggregation or collaborative inference to roadside units or multi-access edge computing (MEC) nodes. EMP~\cite{zhangEMPEdgeassistedMultivehicle2021} illustrates this mode through edge-assisted point-cloud partitioning and fusion. Fourth, hierarchical vehicle-edge-cloud computing combines onboard, peer, roadside, and remote cloud resources. Multi-tier perception task offloading~\cite{xiaoPerceptionTaskOffloading2023} provides direct evidence for collaborative computation across vehicles and roadside units, while cloud resources can support offline model training and lifecycle management~\cite{maSense4FLVehicularCrowdsensing2026}.

The selected deployment pattern and communication scheduling jointly determine whether shared information can be processed and translated into actuation within the control deadline. The end-to-end latency budget includes local encoding, wireless transmission, edge or onboard queueing, fusion and inference, collaborative planning, downlink delivery, control computation, and actuator response. Richer messages may improve shared-state quality while increasing transmission and processing delay. Mobility-aware dynamic offloading~\cite{liMobilityawareDynamicOffloading2021} and broader edge-intelligence studies~\cite{gongEdgeIntelligenceIntelligent2023} show that server load, vehicle mobility, handover, task partitioning, energy use, and link reliability must be considered together.

Because these operating conditions vary over time, safety-critical functions retain an onboard fallback control path. Edge resources can accelerate cross-agent fusion and regional shared-state updates when coverage and service availability are adequate and end-to-end latency meets the control deadline. Existing deployments primarily use edge resources for collaborative perception and computation offloading, while persistent edge-hosted SWM maintenance and closed-loop multi-agent planning remain open system problems. Together, these patterns define the physical execution and timing context for the operational criteria below, while deployment readiness remains a separate evidence claim.

\subsection{Operational Criteria for SWM-Driven Coordination}
\label{sec:ec-criteria}

This subsection explains how this survey analyzes the literature following the architectural loop above. Papers are analyzed for different artifacts, including messages, fused features, trajectories, language plans, control actions, and simulation outputs. The three criteria below convert these heterogeneous artifacts into a common evidence-coding scheme, so later tables can compare methods while separating message passing, shared state, intent reasoning, and control.

A \emph{shared world model (SWM)} is the intermediate representational target for multi-agent collaborative driving. It aligns what agents believe about scene geometry, dynamic actors, occlusions, goals, and likely futures into a representation that can support cooperative reasoning. An SWM may appear as fused BEV features, graph states, latent maps, symbolic scene descriptions, or a time-indexed V2X knowledge pool. For this survey, the key question is whether the representation becomes shared enough to affect downstream planning, regardless of the specific data structure.

The \emph{SWM-driven coordination lens} functions here as a paper-specific organizing lens for the transition from exchanged information to operationally aligned behavior. The term organizes the literature around observable evidence and keeps terminology standardization, communication-protocol design, mathematical convergence, and system-level realization as separate questions. The lens is useful because it makes three observable criteria explicit: aligned state, intent and plan alignment, and coordinated action. These criteria provide a staged reading of the field: first shared state, then aligned future behavior, and finally action selection shaped by the shared representation. We formalize them below for evidence coding in later tables.

\begin{itemize}
    \item \emph{C1: Cross-agent state alignment.} A method satisfies C1 when it maintains an explicit shared or aligned state representation across agents, such as fused BEV features, graph states, shared latent maps, or a maintained V2X knowledge pool. A weaker form appears when alignment is limited to object association or local message aggregation. Simple broadcasting of basic safety messages (BSMs), cooperative awareness messages (CAMs), detections, or raw features is insufficient for C1.
    Adjacent spatial-memory work, such as Shared Spatial Memory Through Predictive Coding~\cite{fangSharedSpatialMemory2025}, provides a mechanistic analogy for shared spatial state under partial observability and limited bandwidth.
    \item \emph{C2: Intent and plan alignment.} A method satisfies C2 when it explicitly models, communicates, retrieves, or negotiates future behavior, goals, roles, or planned actions. A weaker form appears when intent is latent in trajectory forecasting or language reasoning. Perception-only collaboration and reactive control without future-intent representation are insufficient for C2.
    \item \emph{C3: Coordinated downstream action.} A method satisfies C3 when shared state or modeled intent directly changes joint policy selection, trajectory generation, waypoint planning, or multi-agent control. A weaker form appears when a method produces forecasts that could support planning while leaving the loop open. Offline perception, detection, or open loop prediction alone is insufficient for C3.
\end{itemize}

For the cross-section map below, we apply a conservative, source-grounded coding rule. For each row, we ask three questions. First, what observable artifact does a paper expose, such as a message, fused BEV feature, graph state, trajectory, language plan, control action, or simulator output? Second, how is that artifact used downstream, for state construction, intent or plan reasoning, action selection, safety checking, or benchmark evaluation? Third, does a cited paper provide explicit, partial, or no support for the criterion under the definitions above? Explicit support requires an architectural component, learning objective, message type, benchmark output, or control interface that directly supports the criterion. Partial support denotes an implicit, task-specific, or intermediate capability. No support implies that a paper mainly exchanges data, predicts ego behavior, or evaluates a downstream task without exposing the corresponding cross-agent mechanism. The coding is a survey guide only, with deployment readiness, theoretical convergence, and safety certification left as separate claims.

\subsection{Cross-Section Navigation Map}
\label{sec:method-taxonomy}

Table~\ref{tab:method_taxonomy} connects the operational criteria above to representative methods discussed in later technical sections. It serves as a navigation map, with benchmarking and performance comparison treated as separate tasks. Rows are grouped by dominant technical role, and the C1/C2/C3 entries summarize observable evidence in the cited papers. The main-section column points readers to the detailed discussions.

\begin{table}[t]
\centering
\caption{Representative Cross-Section Navigation Map for MAEAD Methods.}
\vspace{-0.5em}
\footnotesize
\label{tab:method_taxonomy}
\begingroup\resizebox{0.9\textwidth}{!}{%
\begin{tabular}{llcccll}
\toprule
\textbf{Method} & \textbf{Primary Mechanism} & \textbf{C1} & \textbf{C2} & \textbf{C3} & \textbf{Evaluation Setting} & \textbf{Main Section} \\
\midrule
\multicolumn{7}{@{}l}{\emph{Communication Baselines}} \\
BSM/CAM (V2X) & Status message exchange & \NoCov & \NoCov & \NoCov & SAE BSM, ETSI CAM & \S5 \\
\midrule
\multicolumn{7}{@{}l}{\emph{Collaborative Perception and Shared-State Construction}} \\
Late Fusion & Object-level fusion & \PartCov & \NoCov & \NoCov & OPV2V, DAIR-V2X & \S4 \\
F-Cooper~\cite{chenFcooperFeatureBased2019} & Feature-level edge fusion & \FullCov & \NoCov & \NoCov & KITTI, T\&J campus data & \S3, \S4 \\
EMP~\cite{zhangEMPEdgeassistedMultivehicle2021} & Edge point-cloud partitioning & \FullCov & \NoCov & \NoCov & GTA V/PreSIL, LTE testbed & \S3, \S4 \\
VIPS~\cite{shiVIPSRealtimePerception2022} & Infrastructure fusion & \PartCov & \NoCov & \NoCov & Campus lamppost, simulator & \S3, \S4 \\
When2com~\cite{liuWhen2comMultiAgentPerception2020} & Learned message selection & \PartCov & \NoCov & \NoCov & AirSim-MAP, ModelNet40 & \S4 \\
V2VNet~\cite{wangV2VNetVehicletoVehicleCommunication2020a} & Graph feature fusion & \FullCov & \PartCov & \NoCov & V2V-Sim & \S4 \\
DiscoNet~\cite{liLearningDistilledCollaboration2021} & Distilled collaboration graph & \FullCov & \NoCov & \NoCov & V2X-Sim 1.0 & \S4 \\
V2X-ViT~\cite{xuV2XViTVehicletoEverythingCooperative2022a} & Transformer feature fusion & \FullCov & \NoCov & \NoCov & V2XSet & \S4 \\
CoBEVT~\cite{xuCoBEVTCooperativeBird2023} & Collaborative BEV fusion & \FullCov & \NoCov & \NoCov & OPV2V & \S4 \\
Where2Comm~\cite{huWhere2commCommunicationefficientCollaborative2024} & Sparse feature exchange & \FullCov & \NoCov & \NoCov & \makecell[l]{OPV2V, V2X-Sim, DAIR-V2X\\CoPerception-UAVs} & \S4 \\
PragComm~\cite{huPragmaticCommunicationMultiAgent2024} & Task-pragmatic messaging & \FullCov & \NoCov & \NoCov & V2V4Real, OPV2V, V2X-Sim 2.0 & \S4, \S5 \\
S2R-ViT~\cite{liS2RViTMultiAgentCooperative2023} & Sim-to-real feature adaptation & \FullCov & \NoCov & \NoCov & OPV2V, V2V4Real & \S4, \S8 \\
HEAL~\cite{luExtensibleFrameworkOpen2024} & Open heterogeneous alignment & \FullCov & \NoCov & \NoCov & OPV2V-H, DAIR-V2X & \S4 \\
Confidence-V2X~\cite{confidencev2x2025} & Confidence-gated exchange & \FullCov & \NoCov & \NoCov & OPV2V, V2XSet, V2V4Real & \S4, \S5 \\
Hyper-V2X~\cite{hyperv2x2026} & Uncertainty-aware BEV fusion & \FullCov & \NoCov & \NoCov & OPV2V & \S4 \\
AFFormer~\cite{afformer2026} & Channel-robust feature fusion & \FullCov & \NoCov & \NoCov & V2XSet, DAIR-V2X & \S4 \\
\midrule
\multicolumn{7}{@{}l}{\emph{Semantic Knowledge and Intent Communication}} \\
V2X-UniPool~\cite{v2xunipool2025} & V2X knowledge pooling & \PartCov & \PartCov & \PartCov & DAIR-V2X & \S5, \S6 \\
\midrule
\multicolumn{7}{@{}l}{\emph{Policy Learning and Generative Planning}} \\
MAPPO~\cite{yu2021mappo} & CTDE policy learning & \NoCov & \NoCov & \PartCov & MPE, SMAC, Hanabi, GRF & \S6 \\
Communication-efficient MARL~\cite{hua2024communication} & Bandwidth-aware MARL & \NoCov & \NoCov & \PartCov & CACC platoon, OpenACC & \S6 \\
RSU Hybrid RL~\cite{marlsmart2025} & RSU-assisted policy learning & \PartCov & \NoCov & \PartCov & CARLA intersections & \S6 \\
MotionDiffuser~\cite{jiang2023motiondiffuser} & Diffusion trajectory generation & \NoCov & \PartCov & \PartCov & Waymo Open Motion Dataset & \S6 \\
SCORP~\cite{scorp2026} & Diffusion planning with RL & \NoCov & \PartCov & \PartCov & WOMD closed-loop benchmark & \S6 \\
\midrule
\multicolumn{7}{@{}l}{\emph{End-to-End Cooperative Driving}} \\
UniV2X (E2E)~\cite{yuEndtoEndAutonomousDriving2024} & Sparse-dense V2X E2E & \FullCov & \NoCov & \FullCov & DAIR-V2X, V2X-Sim & \S7 \\
COOPERNAUT (E2E)~\cite{coopernaut2022} & V2V latent sharing & \FullCov & \NoCov & \FullCov & AutoCastSim/CARLA & \S7 \\
CoDriving~\cite{10979246} & Driving-request V2X E2E & \FullCov & \NoCov & \FullCov & V2Xverse & \S7, \S8 \\
V2X-VLM~\cite{youV2XVLMEndtoEndV2X2024} & VLM-based V2X planning & \FullCov & \PartCov & \FullCov & DAIR-V2X & \S7 \\
\midrule
\multicolumn{7}{@{}l}{\emph{Language and Negotiation Agents}} \\
AgentsCoDriver~\cite{huAgentsCoDriverLargeLanguage2024} & LLM social reasoning & \PartCov & \FullCov & \PartCov & HighwayEnv & \S5, \S6 \\
CoLMDriver~\cite{liu2025colmdriver} & LLM negotiation and waypoints & \PartCov & \FullCov & \PartCov & InterDrive/CARLA & \S5, \S6 \\
CoopReflect~\cite{coopreflect2026} & V2V message refinement & \PartCov & \FullCov & \PartCov & TalkingVehiclesGym & \S5, \S6 \\
CoMAL~\cite{yao2024comal} & Role memory and reasoning & \PartCov & \FullCov & \PartCov & Flow/SUMO & \S5, \S6 \\
KoMA~\cite{koma2024} & Knowledge-driven LLM planning & \PartCov & \FullCov & \PartCov & HighwayEnv & \S5, \S6 \\
AgentsCoMerge~\cite{huAgentsCoMergeLargeLanguage2024} & LLM cooperative merging & \PartCov & \FullCov & \PartCov & SUMO/LimSim++, nuScenes, HighD & \S5, \S6 \\
\midrule
\multicolumn{7}{@{}l}{\emph{Generative Simulation and World Models}} \\
DriveGen~\cite{zhang2025drivegeninfinitediversetraffic} & Generative traffic simulation & \NoCov & \PartCov & \NoCov & Argoverse2, SMARTS & \S6, \S8 \\
GAIA-2~\cite{gaia2_2025} & Controllable world modeling & \NoCov & \PartCov & \NoCov & Internal UK/US/Germany videos & \S5, \S8 \\
\bottomrule
\end{tabular}%
}
\endgroup
\vspace{0.3em}

\begin{minipage}{0.95\textwidth}
\footnotesize \textbf{Note:} Rows are grouped by dominant technical role. C1/C2/C3 follow Section~\ref{sec:ec-criteria}. Symbols use \FullCov{} for explicit support, \PartCov{} for partial support, and \NoCov{} for no support based on observable mechanisms in the cited papers. \textbf{Coding basis:} The Primary Mechanism column records the observable artifact, the Evaluation Setting column records the source context, and the C1/C2/C3 entries record whether that artifact is used for shared-state construction, intent or plan alignment, or coordinated downstream action under the support rules above. The coding is a survey guide, while deployment readiness, theoretical convergence, and safety certification require separate evidence.
\end{minipage}
\vspace{-1em}
\end{table}

\begin{figure}[t]
\centering
\includegraphics[width=0.5\textwidth]{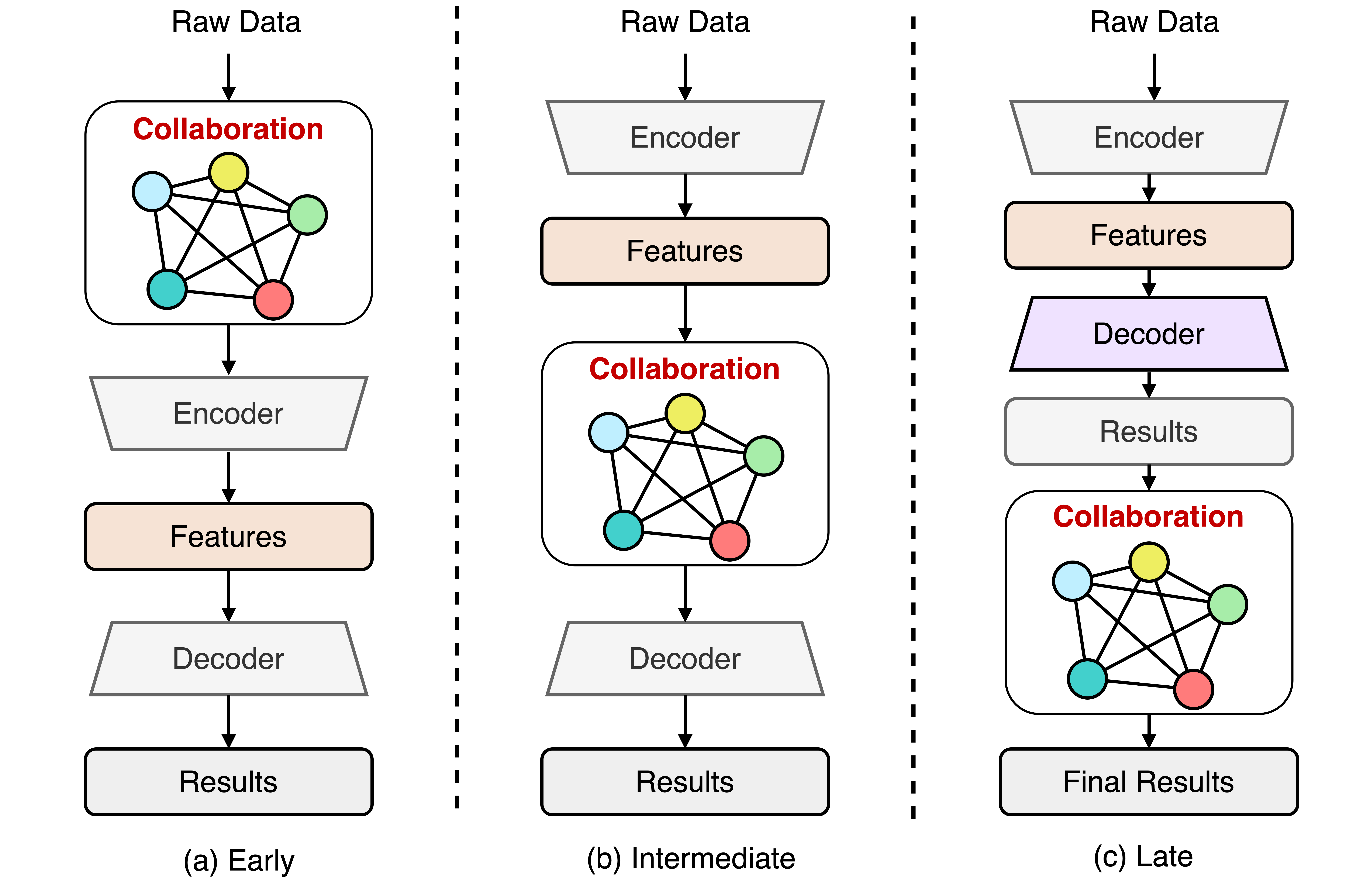}
\caption{Early, intermediate, and late collaborative perception fusion paradigms. Early fusion shares raw observations, intermediate fusion shares BEV or latent features, and late fusion shares object-level results.}
\label{fig:cp_fusion_paradigms}
\vspace{-1em}
\end{figure}

\section{Multi-Agent Collaborative Perception}
\label{sec:collab-perception}

Multi-agent collaborative perception (CP) is a central component of autonomous driving systems that move beyond single-agent ego-centric sensing. By enabling vehicles and roadside units (RSUs) to share sensory data and processed information through vehicle-to-everything (V2X) communication, CP extends the effective perception range, mitigates occlusions, and improves environmental understanding. This section reviews three primary paradigms, early fusion, intermediate fusion, and late fusion, and analyzes how they support shared scene construction under bandwidth, latency, and robustness constraints~\cite{liuVehicletoeverythingAutonomousDriving2023, gao2024vehicle}.

While early V2X-based CP relies on passive broadcast exchange, recent methods increasingly incorporate adaptive and goal-oriented collaboration. These systems select what to share and which collaborators to involve, moving the perception stack closer to explicit cross-agent state alignment. Fig.~\ref{fig:cp_fusion_paradigms} sketches the fusion comparison that motivates the following taxonomy.

\subsection{Paradigm 1: Early Fusion (Raw Data Fusion)}
Early fusion is the most information-rich approach to collaborative perception. It directly fuses raw sensor data streams, such as LiDAR point clouds or camera images, from multiple agents before feature extraction.

\subsubsection{Technical Mechanism and Advantages}

In this paradigm, raw data from collaborating agents is transformed into a common coordinate system using extrinsic calibration and motion compensation~\cite{chenCooperCooperativePerception2019, ahmed2022joint}. The aligned data is then aggregated into a dense representation and passed to a unified perception network.

The main advantage of early fusion is information preservation. By operating on the rawest available observations, the perception model can use fine spatial detail that may be lost after compression or abstraction. This is especially relevant for small, distant, or heavily occluded objects, where observations from another viewpoint can provide missing evidence~\cite{chenCooperCooperativePerception2019, arnold2020cooperative, zhangEMPEdgeassistedMultivehicle2021}.

\subsubsection{Challenges and Implementations}

Early fusion faces two practical challenges. First, raw point clouds and images impose heavy spectrum bandwidth and latency demands on current V2X links~\cite{chenCooperCooperativePerception2019, zhangEMPEdgeassistedMultivehicle2021, qiuAutoCastScalableInfrastructureless2022}. Second, fusion is sensitive to time synchronization and pose estimation errors. Even modest misalignment in timestamps or extrinsic calibration can produce inconsistent geometry in the fused representation~\cite{weiAsynchronyRobustCollaborativePerception2023b, luRobustCollaborative3D2023}.

Because raw-data sharing strains communication links, pure early fusion mainly serves as a near-upper-bound reference under ideal synchronization, accurate calibration, and generous bandwidth~\cite{chenCooperCooperativePerception2019, zhangEMPEdgeassistedMultivehicle2021, qiuAutoCastScalableInfrastructureless2022}. Representative systems instantiate this raw-data paradigm in different deployment settings: connected-vehicle raw LiDAR sharing with compression~\cite{ahmed2022joint}, infrastructure-based cooperative 3D detection~\cite{arnold2020cooperative}, infrastructure-less V2V scheduling~\cite{qiuAutoCastScalableInfrastructureless2022}, and edge-assisted point-cloud partitioning~\cite{zhangEMPEdgeassistedMultivehicle2021}. Together they show the perception gain of rich sharing and the bandwidth pressure it creates. Table~\ref{tab:early_fusion} lists representative early-fusion methods and core ideas.

\begin{table}[t]
\centering
\caption{Representative Early Fusion (Raw Data Fusion) methods for multi-agent collaborative perception.}
\vspace{-0.5em}
\footnotesize
\label{tab:early_fusion}

\begin{tabular}{lll}
\toprule
\textbf{Method} & \textbf{Venue} & \textbf{Primary Mechanism} \\
\midrule
AVR~\cite{qiuAVRAugmentedVehicular2018} & 2018 MobiSys & V2V stereo point-cloud sharing \\
Cooper~\cite{chenCooperCooperativePerception2019} & 2019 ICDCS & Point-cloud V2V fusion \\
Infrastructure CP~\cite{arnold2020cooperative} & 2020 TITS & Infrastructure point-cloud fusion \\
EMP~\cite{zhangEMPEdgeassistedMultivehicle2021} & 2021 MobiCom & Edge point-cloud partitioning \\
AutoCast~\cite{qiuAutoCastScalableInfrastructureless2022} & 2022 MobiSys & Scheduled raw sensor sharing \\
Joint Perception~\cite{ahmed2022joint} & 2022 Sensors & Raw LiDAR sharing with compression \\
Safety Field CP~\cite{zhaoSafetyFieldVehicle2025} & 2025 TITS & Safety-field point-cloud downsampling \\
\bottomrule
\end{tabular}
\vspace{-1em}
\end{table}

\subsection{Paradigm 2: Intermediate Fusion (Feature-Level Fusion)}

Intermediate fusion, also known as feature-level fusion, has become a mainstream direction because it balances perception quality and communication efficiency. Each agent locally extracts compact features, exchanges selected representations, and then fuses received features with its own local state.

\subsubsection{Mechanism and Key Methods}

Agents commonly exchange BEV feature maps, which provide a top-down representation of surrounding geometry across cameras, LiDAR, or multimodal inputs~\cite{liuBEVFusionMultiTaskMultiSensor2023}. The fusion process aligns received features into the ego-agent coordinate system, aggregates them with using attention, graph, or transformer modules, and passes the fused representation to a detection or segmentation head.

Two representative examples are V2X-ViT~\cite{xuV2XViTVehicletoEverythingCooperative2022a} and CoBEVT~\cite{xuCoBEVTCooperativeBird2023}. V2X-ViT introduces a transformer architecture for V2X collaborative perception and uses spatial attention to prioritize informative agents under misalignment. CoBEVT independently studies collaborative BEV perception with multi-camera inputs and fused BEV features.

Beyond these baselines, the intermediate-fusion literature has diversified along several directions. Bandwidth-efficient methods use distilled collaboration graphs and redundancy-aware feature exchange~\cite{liLearningDistilledCollaboration2021, luo2022complementarity}. Robustness-oriented methods address asynchrony, pose noise, domain shift, and heterogeneous agent configurations~\cite{wang2023core, dao2024practical, hong2024multi, luExtensibleFrameworkOpen2024, huAdaptiveCommunicationsCollaborative2023, huFullSceneDomainGeneralization2024}. Multimodal variants extend intermediate fusion from LiDAR-only settings to joint camera-LiDAR cooperation~\cite{yinV2VFormerMultiModalVehicletoVehicle2023}. Confidence-driven sparse communication further shows that local uncertainty can improve feature refinement, outbound gating, and adaptive fusion in V2X collaborative perception~\cite{confidencev2x2025}. Recent reports extend this reliability thread from confidence gating to explicit uncertainty decomposition. Hyper-V2X~\cite{hyperv2x2026} estimates epistemic and aleatoric uncertainty for collaborative BEV semantic segmentation, making shared-state construction more auditable when agents disagree or observe different parts of the scene. AFFormer~\cite{afformer2026} studies feature fusion under channel impairments such as noise, fading, and interference. Together, these reports suggest that CP reliability work is moving from clean-message fusion toward uncertainty-aware fusion, degraded-channel robustness, and heterogeneous participants.

\subsubsection{Handling Communication Challenges}

Intermediate fusion must manage feature compression and latency compensation. Compact BEV maps, quantization, sparsity, and learned message selection reduce the communication load~\cite{huWhere2commCommunicationefficientCollaborative2024, huCommunicationEfficientCollaborativePerception2024, liuWhen2comMultiAgentPerception2020}. Temporal compensation predicts the current location of delayed features using motion models or recurrent integration, but residual asynchrony can still degrade performance in fast-changing scenes~\cite{weiAsynchronyRobustCollaborativePerception2023b}.

Communication-critical benchmarks show that packet loss, asynchronous clocks, and topology-dependent bottlenecks can change the apparent ranking of methods relative to idealized evaluations~\cite{liLearningVehicletoVehicleCooperative2023, glaser2023we}. Recent work therefore optimizes both \textit{who} communicates and \textit{what} is transmitted through priority-aware scheduling, directed information flow, channel-aware graph control, confidence-aware gating, and domain adaptation for cooperative data fusion~\cite{fangPACPPriorityAwareCollaborative2024, taoDirectedCPEfficientCollaborative2025, anChannelAwareThroughputCooperative2024, confidencev2x2025, huAdaptiveCommunicationsCollaborative2023, huFullSceneDomainGeneralization2024}. Table~\ref{tab:intermediate_fusion} lists representative intermediate-fusion methods and core mechanisms.

\begin{table}[t]
\centering
\caption{Representative Intermediate Fusion (Feature-Level Fusion) methods for multi-agent collaborative perception.}
\vspace{-0.5em}
\footnotesize
\label{tab:intermediate_fusion}
\resizebox{0.6\textwidth}{!}{%
\begin{tabular}{lll}
\toprule
\textbf{Method} & \textbf{Venue} & \textbf{Primary Mechanism} \\
\midrule
\multicolumn{3}{@{}l}{\emph{Core Feature-Fusion Architectures}} \\
F-Cooper~\cite{chenFcooperFeatureBased2019} & 2019 SEC & Feature-level edge fusion \\
V2VNet~\cite{wangV2VNetVehicletoVehicleCommunication2020a} & 2020 ECCV & Graph feature fusion \\
V2X-ViT~\cite{xuV2XViTVehicletoEverythingCooperative2022a} & 2022 ECCV & Transformer feature fusion \\
CoBEVT~\cite{xuCoBEVTCooperativeBird2023} & 2022 CoRL & Cooperative BEV fusion \\
\midrule
\multicolumn{3}{@{}l}{\emph{Representation and Task Extensions}} \\
CoCa3D~\cite{huCollaborationHelpsCamera2023} & 2023 CVPR & Camera-only collaboration \\
HM-ViT~\cite{xiangHMViTHeteroModalVehicletoVehicle} & 2023 ICCV & Hetero-modal fusion \\
CORE~\cite{wang2023core} & 2023 ICCV & Reconstruction-guided feature fusion \\
V2VFormer++~\cite{yinV2VFormerMultiModalVehicletoVehicle2023} & 2023 TITS & Global-local BEV transformer \\
TransIFF~\cite{chenTransIFFInstanceLevel2023} & 2023 ICCV & Instance-level feature fusion \\
CoHFF~\cite{songCollaborativeSemanticOccupancy2024} & 2024 CVPR & Hybrid occupancy feature fusion \\
CoopTrack~\cite{zhongCoopTrackExploring2025} & 2025 ICCV & Sparse instance feature tracking \\
HeCoFuse~\cite{weiHeCoFuseCrossModal2025} & 2025 ITSC & Cross-modal feature fusion \\
SCOPE~\cite{yangSpatioTemporalDomainAwareness2023} & 2023 ICCV & Spatio-temporal aggregation \\
V2XPnP~\cite{zhouV2XPnPVehicletoEverything2025} & 2025 ICCV & Spatio-temporal transformer fusion \\
Long-SCOPE~\cite{wangLongSCOPEFully2026} & 2026 CVPR & Sparse long-range query fusion \\
CEST~\cite{chenCESTEnhancing2026} & 2026 TITS & Query-aware spatial-temporal fusion \\
\midrule
\multicolumn{3}{@{}l}{\emph{Communication-Efficient Feature Exchange}} \\
When2com~\cite{liuWhen2comMultiAgentPerception2020} & 2020 CVPR & Learned message selection \\
Who2com~\cite{liuWho2comCollaborativePerception2020} & 2020 ICRA & Handshake-based feature exchange \\
DiscoNet~\cite{liLearningDistilledCollaboration2021} & 2021 NeurIPS & Distilled collaboration graph \\
Where2Comm~\cite{huWhere2commCommunicationefficientCollaborative2024} & 2022 NeurIPS & Sparse feature exchange \\
S-AdaFusion~\cite{qiaoAdaptiveFeatureFusion2023} & 2023 WACV & Adaptive feature selection \\
How2comm~\cite{yangHow2commCommunicationEfficient2023} & 2023 NeurIPS & Sender-centric selection \\
What2comm~\cite{yangWhat2commCommunicationefficientCollaborative2023} & 2023 MM & Feature-decoupling exchange \\
PACP~\cite{fangPACPPriorityAwareCollaborative2024} & 2024 TMC & Priority-aware BEV exchange \\
Fusion2comm~\cite{chuOcclusionGuidedMultimodal2025} & 2025 PR & Occlusion-guided foreground features \\
FocalComm~\cite{shenkutFocalCommHard2026} & 2026 WACV & Hard-instance feature exchange \\
What2Keep~\cite{zhangWhat2KeepCommunicationEfficient2026} & 2026 CVIU & Consensus-based feature selection \\
Map4comm~\cite{qiuMap4commMapAware2026} & 2026 Inf. Fusion & Map-masked voxel feature selection \\
UMC~\cite{wangUMCUnified2023} & 2023 ICCV & Multi-resolution feature exchange \\
CodeFilling~\cite{huCommunicationEfficientCollaborativePerception2024} & 2024 CVPR & Codebook message coding \\
ERMVP~\cite{zhangERMVPCommunicationEfficientCollaborationRobust} & 2024 CVPR & Hierarchical feature sampling \\
SparseComm~\cite{liuSparseCommEfficient2025} & 2025 PR & Sparse instance feature exchange \\
SparseAlign~\cite{yuanSparseAlignFully2025} & 2025 CVPR & Fully sparse feature alignment \\
QCTF~\cite{chenQCTFQuantized2025} & 2025 TITS & Quantized transferable fusion \\
CoRange~\cite{shuCoRangeCollaborative2025} & 2025 TIV & Range-aware adaptive fusion \\
AccBEV~\cite{shiV2VCooperativePerception2025} & 2025 TITS & Adaptive lossy BEV fusion \\
PragComm~\cite{huPragmaticCommunicationMultiAgent2024} & 2026 TPAMI & Task-pragmatic messaging \\
CoGMoE~\cite{liCoGMoESparse2026} & 2026 KBS & Sparse GraphMoE feature fusion \\
InfoCom~\cite{weiInfoComKilobyteScale2026} & 2026 AAAI & Information-bottleneck feature coding \\
V2X-DSC~\cite{zengV2XDSCMultiAgent2026} & 2026 arXiv & Distributed source feature coding \\
\midrule
\multicolumn{3}{@{}l}{\emph{Robust, Temporal, and Heterogeneous Fusion}} \\
SyncNet~\cite{leiLatencyAwareCollaborativePerception2022a} & 2022 ECCV & Feature-level latency compensation \\
CoBEVFlow~\cite{weiAsynchronyRobustCollaborativePerception2023b} & 2023 NeurIPS & BEV-flow temporal alignment \\
FFNet~\cite{yuFlowBasedFeatureFusion2023} & 2023 NeurIPS & Feature-flow prediction \\
MRCNet~\cite{hong2024multi} & 2024 CVPR & Motion-aware robust fusion \\
SCOPE++~\cite{yangRobustMultiAgent2025} & 2025 TCSVT & Spatio-temporal aware fusion \\
CoAlign~\cite{luRobustCollaborative3D2023} & 2023 ICRA & Pose-robust feature fusion \\
HEAL~\cite{luExtensibleFrameworkOpen2024} & 2024 ICLR & Heterogeneous alignment \\
STAMP~\cite{gaoSTAMPScalableTask2025} & 2025 ICLR & Adapter-reverter BEV alignment \\
CoBEVMoE~\cite{kongCoBEVMoEHeterogeneityAware2026} & 2026 ICRA & Dynamic MoE BEV fusion \\
Hyper-V2X~\cite{hyperv2x2026} & 2026 arXiv & Uncertainty-aware BEV fusion \\
AFFormer~\cite{afformer2026} & 2026 arXiv & Channel-robust fusion \\
\bottomrule
\end{tabular}
}
\vspace{-1em}
\end{table}

\subsection{Paradigm 3: Late Fusion (Object-Level Fusion)}

Late fusion exchanges final perception outputs after local processing. Each agent performs local detection independently and shares lightweight object-level messages, such as 3D bounding boxes, velocities, and class labels.

\subsubsection{Mechanism and Limitations}

The core operation is object-level association and refinement. The receiving agent matches object hypotheses that correspond to the same physical entity, then combines position, size, velocity, and confidence estimates into a consolidated result. This approach minimizes communication overhead and is comparatively robust to dropped messages because transmitted packets are compact.

The main limitation is that late fusion cannot recover objects missed by  local detectors. If no agent detects an occluded object, object-level aggregation will not be able to refine the detection. As a result, late fusion is attractive for deployment but offers a smaller perception gain than raw-data or feature-level collaboration~\cite{xuOPV2VOpenBenchmark2022, yuDAIRV2XLargeScaleDataset2022}.

\subsubsection{Advanced Late Fusion Methods}

Recent late-fusion methods improve object association across agents. VIPS~\cite{shiVIPSRealtimePerception2022} matches detections through graph association for real-time vehicle-to-infrastructure fusion, and model-agnostic aggregation merges object outputs without assuming a shared detector backbone~\cite{xu2023model}. These designs share a recurring tradeoff: object-level fusion is the easiest paradigm to deploy across heterogeneous agents, but it stays bottlenecked by upstream detection and association quality.

Table~\ref{tab:late_fusion} lists representative late-fusion systems and benchmark baselines whose exchanged units are final detections, object states, or tracks.

\begin{table}[t]
\centering
\caption{Representative Late Fusion (Object-Level Fusion) methods for multi-agent collaborative perception.}
\vspace{-0.5em}
\footnotesize

\label{tab:late_fusion}
\begin{tabular}{lll}
\toprule
\textbf{Method} & \textbf{Venue} & \textbf{Primary Mechanism} \\
\midrule
AutoC2X/RSPU~\cite{tsukadaNetworkedRoadsidePerception2020} & 2020 Sensors & RSU object-message sharing \\
OPV2V Late Fusion~\cite{xuOPV2VOpenBenchmark2022} & 2022 ICRA & Box broadcast with NMS \\
TCLF~\cite{yuDAIRV2XLargeScaleDataset2022} & 2022 CVPR & Time-compensated V2I box fusion \\
VIPS~\cite{shiVIPSRealtimePerception2022} & 2022 MobiCom & Object graph matching for V2I fusion \\
MAPF~\cite{xu2023model} & 2023 ICRA & Calibrated bounding-box aggregation \\
Late Collaborative Framework~\cite{fadiliLateCollaborativePerception2025} & 2025 ICRAS & Multi-source 3D box attribute fusion \\
CPM Track-to-Track Fusion~\cite{castelinoTracktoTrackFusion2026} & 2026 Sensors & CPM track association and fusion \\
\bottomrule
\end{tabular}
\vspace{-1em}
\end{table}

\subsection{Paradigm-Level Trade-offs in Fusion Strategies}
\label{sec:cp-comparison}

After the per-method tables, this subsection summarizes paradigm-level trade-offs across collaborative perception fusion strategies. The main verifiable conclusion is that feature-level methods often improve perception over object-level fusion under matched benchmark protocols, while sparse and learned communication schemes reduce message load, saving spectrum bandwidth compared with dense feature exchange. Early fusion provides a near-upper-bound reference under bandwidth-rich and tightly synchronized settings, with raw-data exchange creating high communication and calibration burdens. Intermediate fusion is the central research direction when accuracy and limited V2X bandwidth must be balanced, especially through sparse BEV or latent-feature exchange. Late fusion remains attractive for heterogeneous deployment and unreliable communication because compact object-level messages are easier to exchange and fuse, and its gains remain constrained by upstream detection and association quality.

\subsection{Emerging Directions in Collaborative Perception}

Collaborative perception is moving beyond fixed fusion toward adaptive collaboration driven by task uncertainty, semantic reasoning, and robustness requirements. Active collaboration and uncertainty-aware feature fusion are indexed in Table~\ref{tab:intermediate_fusion}. Table~\ref{tab:emerging_cp_methods} focuses on complementary emerging directions: security and trust, language grounding, and generative world models.

\begin{table}[t]
\centering
\caption{Representative emerging-direction papers for collaborative perception.}
\vspace{-0.5em}
\footnotesize

\label{tab:emerging_cp_methods}
\begin{tabular}{lll}
\toprule
\textbf{Method} & \textbf{Venue} & \textbf{Primary Mechanism} \\
\midrule
\multicolumn{3}{@{}l}{\emph{Security and trust}} \\
ROBOSAC~\cite{liUsAdversariallyRobust2023} & 2023 ICCV & Sampling consensus defense \\
CP-Guard~\cite{huCPGuardMaliciousAgent2024} & 2025 AAAI & Collaborator consistency filtering \\
GCP~\cite{taoGCPGuardedCollaborative2025} & 2026 TDSC & Spatio-temporal attack detection \\
CP-Guard+~\cite{huCPGuardPlusNewParadigm2025} & 2025 arXiv & Enhanced malicious-agent detection \\
CP-UniGuard~\cite{huCPUniGuardUnified2026} & 2026 TMC & Unified malicious-agent defense \\
MVIG~\cite{tao2026learningmutualviewinformation} & 2026 arXiv & Adaptive trust weighting \\
\midrule
\multicolumn{3}{@{}l}{\emph{Language and semantic grounding}} \\
DriveGPT4~\cite{xuDriveGPT4InterpretableEndtoend2023} & 2024 RA-L & VLM driving explanations \\
Talk2BEV~\cite{choudharyTalk2BEVLanguageenhancedBird2023} & 2024 ICRA & Language-grounded BEV querying \\
DriveLM~\cite{simaDriveLMDrivingGraph2023} & 2024 ECCV & Graph VQA grounding \\
AgentsCoDriver~\cite{huAgentsCoDriverLargeLanguage2024} & 2024 arXiv & LLM lifelong collaboration \\
KoMA~\cite{koma2024} & 2024 TIV & Knowledge-driven agent reasoning \\
V2X-UniPool~\cite{v2xunipool2025} & 2025 arXiv & Time-indexed V2X knowledge pooling \\
\midrule
\multicolumn{3}{@{}l}{\emph{Generative world models}} \\
GAIA-2~\cite{gaia2_2025} & 2025 arXiv & Controllable multi-view generation \\
MiLA~\cite{mila2025} & 2025 arXiv & Long-term multi-view generation \\
Epona~\cite{epona2025} & 2025 ICCV & Autoregressive diffusion world model \\
DrivingGen~\cite{drivinggen2026} & 2026 ICLR & Generative world-model benchmark \\
WorldDrive~\cite{worlddrive2026} & 2026 arXiv & Generation-planning representation \\
ShareVerse~\cite{shareverse2026} & 2026 arXiv & Multi-agent consistent video generation \\
V2XCrafter~\cite{taoV2XCrafterGenerate2026} & 2026 arXiv & Cross-agent scene generation \\
\bottomrule
\end{tabular}
\vspace{-1em}
\end{table}

\subsubsection{Active Collaboration}

Traditional CP is largely passive, with agents broadcasting or receiving information regardless of immediate utility. Active collaboration introduces a task-oriented strategy where agents request occluded regions or specific missing evidence only when needed. The mechanism is confidence-driven: an agent identifies a region of interest with low confidence, then requests support from nearby vehicles or RSUs that are predicted to observe the region more clearly. This can reduce unnecessary communication and prioritize evidence that resolves immediate uncertainty~\cite{huWhere2commCommunicationefficientCollaborative2024}. Methods such as When2com~\cite{liuWhen2comMultiAgentPerception2020} and What2comm~\cite{yangWhat2commCommunicationefficientCollaborative2023} learn policies for deciding when to communicate and what content to request. Confidence-V2X~\cite{confidencev2x2025} gives a recent example by using confidence maps to prepare sparse features, gate outbound communication, and adaptively fuse received information. Hyper-V2X~\cite{hyperv2x2026} and AFFormer~\cite{afformer2026} recommend a complementary trend: collaboration should expose calibrated uncertainty, degraded-channel robustness, fallback behavior, and fused labels or features. Viewed at the shared-state level, these methods make message selection part of scene maintenance: agents communicate when new evidence can reduce uncertainty in the fused representation.

\subsubsection{Robustness against Malicious Agents}

Shared perception messages create an attack surface. Malicious or faulty agents can inject misleading features, object lists, or pose information. Recent defenses detect suspicious collaborators and enforce consistency among trustworthy agents. ROBOSAC~\cite{liUsAdversariallyRobust2023} uses a sampling-based consensus defense that compares collaborative results from random teammate subsets with individual perception to identify attacker-free collaborator subsets. Representative methods also include CP-Guard~\cite{huCPGuardMaliciousAgent2024}, CP-Guard+~\cite{huCPGuardPlusNewParadigm2025}, guarded collaborative perception with spatio-temporal malicious detection~\cite{taoGCPGuardedCollaborative2025}, and CP-UniGuard~\cite{huCPUniGuardUnified2026}. Direct data-fabrication and LiDAR-spoofing attacks further show that robust aggregation and cross-agent consistency checks should be treated as core systems requirements~\cite{zhangDataFabricationCollaborative2023, zhangCooperativePerceptionSafe2023}. Beyond fixed-defense designs, recent work models inter-agent trust adaptively: a mutual view information graph learns pairwise observation consistency so that the fusion stage can down-weight adversarial collaborators under changing attack patterns~\cite{tao2026learningmutualviewinformation}.

\subsubsection{LLM-Augmented Perception}

Large language models (LLMs) and vision-language models (VLMs) can enrich perception outputs with semantic context. They can translate numerical detections into structured scene descriptions, support high-level anomaly detection, and help align shared world models across agents~\cite{xuDriveGPT4InterpretableEndtoend2023}. For example, a perception system may report a vehicle pose, while a language-grounded module can connect that pose to a contextual description such as an illegally stopped vehicle obstructing a bike lane.

Representative single-agent or scene-grounded systems illustrate this role. DriveGPT4~\cite{xuDriveGPT4InterpretableEndtoend2023} maps visual driving context into interpretable explanations and driving-oriented answers. DriveLM~\cite{simaDriveLMDrivingGraph2023} formulates perception, prediction, and planning as graph visual question answering, while Talk2BEV~\cite{choudharyTalk2BEVLanguageenhancedBird2023} grounds free-form language queries in BEV maps. Their main contribution here is semantic grounding for perception outputs, with multi-agent fusion still requiring cross-view consistency and timing validation.

In the longer term, LLMs may serve as a semantic consensus layer that summarizes multimodal evidence from multiple agents into a shared narrative of the scene. AgentsCoDriver~\cite{huAgentsCoDriverLargeLanguage2024} and KoMA~\cite{koma2024} point toward language-grounded coordination, and V2X-UniPool~\cite{v2xunipool2025} connects this direction to time-indexed V2X knowledge pooling. Existing language-grounded driving systems and V2X knowledge-pooling frameworks therefore point toward translating shared scene understanding into coordinated driving decisions.

\subsubsection{Generative World Models for Predictive Collaborative Perception}

A complementary direction is to treat collaborative perception as predictive shared-state construction that extends beyond current-frame fusion. Driving video and sensor world models can generate plausible future observations, expose rare visibility conditions, and test whether a fused scene remains consistent across views and time~\cite{drivedreamer2024, zhao2024drivedreamer2world, gaia2_2025, unidrivedreamer2026}. This is especially relevant to MAEAD because shared-state alignment depends on maintaining a coherent scene representation when agents move, occlusions change, and communication is delayed.

Recent systems sharpen this point from different angles. DrivingGen~\cite{drivinggen2026} provides an evaluation benchmark for generative driving world models, WorldDrive~\cite{worlddrive2026} connects future scene generation with planning representations, and ShareVerse~\cite{shareverse2026} explicitly targets multi-agent consistent video generation for shared world modeling. V2XCrafter~\cite{taoV2XCrafterGenerate2026} similarly generates cross-agent-consistent collaborative driving scenes through staged diffusion with a collaboration-view-graph attention module, aiming to augment collaborative 3D detection data. Long-horizon video models such as MiLA~\cite{mila2025} and Epona~\cite{epona2025} further stress temporal consistency and planning-coupled future prediction. Waymo's World Model~\cite{waymoworldmodel2026} provides an external technical example for controllable multi-sensor rare-event simulation. Its evidence boundary should be kept separate from academic benchmarks. Overall, these models support prediction, data augmentation, closed-loop testing, and SWM validation. Real-time safe collaborative perception still depends on calibrated uncertainty, physical plausibility checks, and downstream safety validation.

Together, active collaboration, LLM-augmented perception, and generative world models point toward a setting in which agents build, predict, and reason over a more explicit shared semantic world model. Recent surveys also distinguish intersection-centric collaborative perception from the wider vehicle-edge-cloud setting, underscoring that topology and infrastructure participation affect which fusion assumptions remain valid in practice~\cite{gao2024survey, gao2024vehicle}. The synthesis from this section is that collaborative perception provides the clearest current progress on shared-state construction, especially through intermediate and sparse feature fusion. Its downstream value still depends on converting perception outputs into communicable intent, plans, or policy constraints.

\section{Inter-Agent Cognition \& Communication}
\label{sec:cognition}
This section examines the communication layer that connects shared state, intent inference, and coordinated downstream behavior. Explicit V2X messages provide structured information that onboard systems can directly process, behavior observation supports implicit intent inference, semantic communication compresses meaning, and language or plan-level exchange exposes proposed roles and maneuvers.

\begin{table}[t]
\centering
\caption{Representative Inter-Agent Cognition and Communication Standards and Methods for MAEAD.}
\vspace{-0.5em}
\footnotesize

\label{tab:cognition_comm_methods}
\begin{tabular}{lll}
\toprule
\textbf{Method} & \textbf{Venue} & \textbf{Primary Mechanism} \\ \midrule
\multicolumn{3}{@{}l}{\emph{Explicit V2X Communication Standards and Data Exchange}} \\
IEEE 802.11p/WAVE~\cite{ieee80211p2010, ieee1609wave2019} & 2010/2019 IEEE Std. & DSRC access stack \\
ETSI CAM/DENM~\cite{etsiEN3026372CAM2019, etsiEN3026373DENM2019} & 2019 ETSI EN & Awareness and event messages \\
3GPP C-V2X/NR-V2X~\cite{threegppTS23285V2X2019, threegppTS23287V2X2022} & 2019/2022 3GPP TS & PC5/Uu V2X architecture \\
SAE J2735 BSM~\cite{saeJ2735V2X2020} & 2020 SAE Std. & V2X message dictionary \\
MISO-V~\cite{liuMISOMisbehaviorDetection2021} & 2021 IV & CPM trust checking \\
ETSI CPS~\cite{etsiTS103324CollectivePerception2023} & 2023 ETSI TS & CPM object sharing \\
MCM analysis~\cite{oliveiraManeuverCoordinationAnalysis2024} & 2024 Electronics & Maneuver exchange \\
\midrule
\multicolumn{3}{@{}l}{\emph{Communication-Computing Co-Design}} \\
EdgeCooper~\cite{luoEdgeCooperNetworkAwareCooperative2023} & 2023 JSAC & Network-aware edge fusion \\
Harbor~\cite{zhuHarborCollaborativeVehicular2024} & 2024 SenSys & V2I/V2V workload partition \\
Q-CPTO~\cite{zakiQualityAwareTask2024} & 2024 TVT & Value-aware perception offloading \\
ITORA~\cite{dongTaskOffloadingResource2025} & 2025 TITS & Joint sensing and resource control \\
\midrule
\multicolumn{3}{@{}l}{\emph{Implicit Communication and Learned Signaling}} \\
CommNet~\cite{sukhbaatarLearningMultiagentCommunication2016} & 2016 NeurIPS & Learned continuous messages \\
DIAL~\cite{foersterLearningCommunicateDeep2016} & 2016 NeurIPS & Differentiable messaging \\
IntentNet~\cite{casasIntentNetLearning2018} & 2018 CoRL & Visual intent prediction \\
MA-MeSN~\cite{bhallaDeepMultiAgent2020} & 2020 Advances in AI & Communication-aware MARL \\
HiVT~\cite{zhou2022hivt} & 2022 CVPR & Hierarchical interaction encoding \\
EqMotion~\cite{eqmotion2023} & 2023 CVPR & Equivariant motion prediction \\
MTR++~\cite{shi2024mtr++} & 2024 TPAMI & Guided intention queries \\
\midrule
\multicolumn{3}{@{}l}{\emph{Semantic and Knowledge Communication}} \\
V2X-UniPool~\cite{v2xunipool2025} & 2025 arXiv & V2X knowledge pooling \\
Confidence-V2X~\cite{confidencev2x2025} & 2026 Adv. Eng. Inf. & Confidence-gated features \\
\midrule
\multicolumn{3}{@{}l}{\emph{Language, VLM, and Plan-Level Communication}} \\
Agent-Driver~\cite{maoLanguageAgentAutonomous2023} & 2024 CoLM & Tool-using LLM agent \\
DriveGPT4~\cite{xuDriveGPT4InterpretableEndtoend2023} & 2024 RA-L & VLM driving explanations \\
DriveLM~\cite{simaDriveLMDrivingGraph2023} & 2024 ECCV & Graph VQA grounding \\
AgentsCoDriver~\cite{huAgentsCoDriverLargeLanguage2024} & 2024 arXiv & Lifelong LLM driving \\
CoLMDriver~\cite{liu2025colmdriver} & 2025 ICCV & LLM negotiation \\
AgentsCoMerge~\cite{huAgentsCoMergeLargeLanguage2024} & 2025 TMC & Ramp-merging dialogue \\
CMP~\cite{wang2025cmp} & 2025 RA-L & Motion-prediction messages \\
CoopReflect~\cite{coopreflect2026} & 2026 AAMAS & Natural-language V2V \\ \bottomrule
\end{tabular}%
\vspace{-1em}
\end{table}

\begin{figure}[t]
\centering
\resizebox{\textwidth}{!}{%
\begin{tikzpicture}[
    x=1cm,y=1cm,font=\sffamily,
    lay/.style={rounded corners=4pt,line width=0.5pt,inner sep=0pt},
    flowarr/.style={->,>=latex,draw=black!50,line width=0.8pt}
]

\path[use as bounding box] (0,0) rectangle (18,4.65);

\draw[flowarr] (3.15,4.26) -- node[above=2pt,font=\sffamily\scriptsize,black!60]
    {from situational awareness to maneuver coordination} (14.85,4.26);

\fill[lay,fill=RoyalBlue!8,draw=RoyalBlue!55] (0.40,1.72) rectangle (5.90,3.92);
\fill[lay,fill=TealBlue!10,draw=TealBlue!60] (6.25,1.72) rectangle (11.75,3.92);
\fill[lay,fill=Orange!9,draw=Orange!65]      (12.10,1.72) rectangle (17.60,3.92);
\draw[lay,draw=RoyalBlue!55] (0.40,1.72) rectangle (5.90,3.92);
\draw[lay,draw=TealBlue!60]  (6.25,1.72) rectangle (11.75,3.92);
\draw[lay,draw=Orange!65]    (12.10,1.72) rectangle (17.60,3.92);

\node[font=\sffamily\footnotesize\bfseries,align=center,text width=5.1cm] at (3.15,3.60)
    {Awareness and Event Messages};
\node[font=\sffamily\scriptsize,align=center,text width=5.1cm] at (3.15,2.94)
    {BSM~\cite{saeJ2735V2X2020}, CAM~\cite{etsiEN3026372CAM2019},\\DENM~\cite{etsiEN3026373DENM2019}};
\node[font=\sffamily\scriptsize,text=black!60,align=center,text width=5.1cm] at (3.15,2.15)
    {ego state, traffic events, and\\hazard notifications};

\node[font=\sffamily\footnotesize\bfseries,align=center,text width=5.1cm] at (9.00,3.60)
    {Collective Perception Messages};
\node[font=\sffamily\scriptsize,align=center,text width=5.1cm] at (9.00,2.94)
    {CPM via ETSI CPS~\cite{etsiTS103324CollectivePerception2023}};
\node[font=\sffamily\scriptsize,text=black!60,align=center,text width=5.1cm] at (9.00,2.15)
    {perceived objects, sensor information,\\and perception regions};

\node[font=\sffamily\footnotesize\bfseries,align=center,text width=5.1cm] at (14.85,3.60)
    {Maneuver and Intent Messages};
\node[font=\sffamily\scriptsize,align=center,text width=5.1cm] at (14.85,2.94)
    {MCM~\cite{oliveiraManeuverCoordinationAnalysis2024},\\status and intent~\cite{wang2022multi}};
\node[font=\sffamily\scriptsize,text=black!60,align=center,text width=5.1cm] at (14.85,2.15)
    {future trajectories, coordination requests,\\and maneuver commitments};

\fill[black!5,rounded corners=4pt] (0.40,0.25) rectangle (17.60,1.25);
\node[anchor=west,font=\sffamily\footnotesize\bfseries,black!62] at (0.70,0.94)
    {Transport substrate};
\node[anchor=west,font=\sffamily\scriptsize,black!58] at (0.70,0.53)
    {DSRC with IEEE 802.11p and WAVE~\cite{ieee80211p2010, ieee1609wave2019}};
\node[anchor=west,font=\sffamily\scriptsize,black!58] at (7.60,0.53)
    {C-V2X and NR-V2X~\cite{threegppTS23285V2X2019, threegppTS23287V2X2022, khan2023advancingcv2x}};
\node[anchor=west,font=\sffamily\scriptsize,black!58] at (13.60,0.53)
    {V2V, V2I, and V2P links~\cite{ansari2021joint}};
\draw[dashed,draw=black!35,line width=0.6pt] (0.46,1.25) -- (17.54,1.25);

\end{tikzpicture}%
}
\vspace{-2em}

\caption{Explicit V2X message classes in MAEAD. Content progresses from awareness and event notification to collective perception and maneuver or intent exchange. A common V2X transport substrate carries these message classes over V2V, V2I, and V2P links.}
\vspace{-1em}
\label{fig:message_hierarchy}
\end{figure}
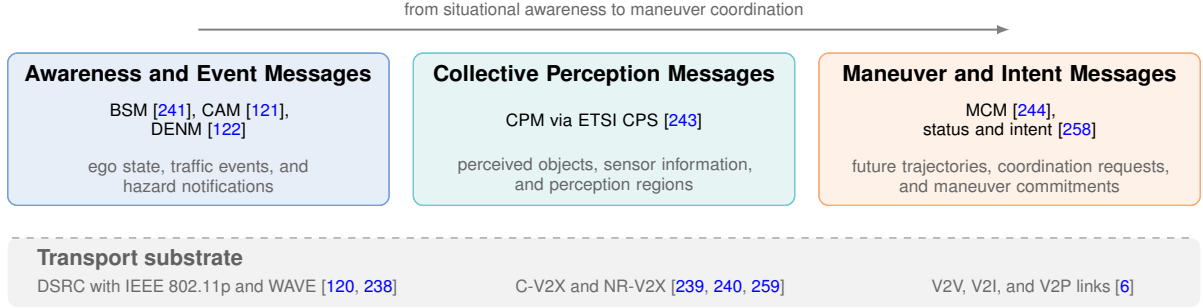

\subsection{Explicit V2X Communication and Data Exchange}

\subsubsection{V2X Protocols and Message Classes}
Explicit communication is the transport and message substrate for cooperative driving, enabling agents to exchange information over wireless channels~\cite{balkusSurveyCollaborativeMachine2022}. The two dominant families are Dedicated Short-Range Communications (DSRC), based on IEEE 802.11p/WAVE~\cite{ieee80211p2010, ieee1609wave2019}, and Cellular V2X (C-V2X)~\cite{threegppTS23285V2X2019, threegppTS23287V2X2022}, which extends cellular infrastructure toward vehicle-to-vehicle (V2V), vehicle-to-infrastructure (V2I), and vehicle-to-pedestrian (V2P) exchange~\cite{ansari2021joint, noor-a-rahim6GVehicletoEverythingV2X2022}. Their role in MAEAD is organized here through three message classes. Fig.~\ref{fig:message_hierarchy} relates these classes to the coordination content they expose while keeping the transport substrate separate.

The first class consists of awareness and event messages. Basic Safety Messages (BSMs)~\cite{jung2020v2x} carry ego-vehicle state such as position, speed, and heading. European cooperative intelligent transport systems (C-ITS) stacks use Cooperative Awareness Messages (CAMs)~\cite{etsiEN3026372CAM2019} and Decentralized Environmental Notification Messages (DENMs)~\cite{etsiEN3026373DENM2019} for analogous status and hazard information. These messages improve situational awareness, although message availability alone is insufficient for the state alignment criterion used in this survey.

The second class is message-level collective perception. The ETSI Collective Perception Service (CPS)~\cite{etsiTS103324CollectivePerception2023} defines Collective Perception Messages (CPMs) for sharing perceived objects, sensor information, and perception regions. CPMs move explicit V2X closer to cross-agent state alignment because they expose another station's perceived objects in addition to its pose. MISO-V~\cite{liuMISOMisbehaviorDetection2021} shows that CPM-based perception also requires trust and consistency checks because received objects may be faulty or malicious.

The third class is maneuver and intent exchange. Status-and-intent sharing~\cite{wang2022multi} and Maneuver Coordination Messages (MCMs)~\cite{oliveiraManeuverCoordinationAnalysis2024} communicate future trajectories, maneuver commitments, or coordination requests. These messages make intent and plan alignment more explicit, while coordinated downstream action still depends on the planner and controller that consume the exchanged intent.

Recent advances in C-V2X have been driven by the evolution from 4G-V2X to 5G New Radio V2X (NR-V2X) through 3GPP standardization~\cite{khan2023advancingcv2x}. The 5G NR-V2X line supports advanced use cases such as cooperative maneuvers, sensor data sharing, and remote driving. For MAEAD, the key implication is that communication standards provide the transport substrate, while agent-level reasoning and SWM maintenance remain higher-layer system-design problems.

\subsubsection{Channel Modeling and Resource Allocation}
Effective explicit V2X exchange depends on channel modeling and resource allocation because useful CPMs and MCMs can be larger and more timing-sensitive than baseline awareness messages. The vehicular channel experiences high mobility, frequent topology changes, and non-line-of-sight conditions, so channel models must account for rapidly changing link characteristics.

Resource allocation is especially important in dense traffic, where channel congestion can delay or drop safety-relevant messages. Distributed Congestion Control (DCC) and Mode 4 in C-V2X allow vehicles to select and manage resources without centralized scheduling~\cite{ansari2021joint}. Graph neural network (GNN) resource allocation~\cite{gnnv2x2024} models nearby vehicles as a dynamic interaction graph, while communication-critical scenario mining~\cite{glaser2023we} shows that practical bottlenecks often arise in rare topology states where visibility, demand, and contention change abruptly. Delay-aware scheduling and packet-loss-aware perception therefore remain tightly coupled problems~\cite{liLearningVehicletoVehicleCooperative2023}.

\subsubsection{Communication-Computing Co-Design}
The deployment alternatives described in Section~\ref{sec:deployment-computing} couple wireless scheduling with task placement. A collaborative pipeline may encode data onboard, transmit intermediate representations, execute fusion at a vehicle or edge node, and return an updated shared state before planning. EdgeCooper~\cite{luoEdgeCooperNetworkAwareCooperative2023} jointly selects complementary LiDAR regions and schedules multi-hop V2X delivery for edge fusion. Harbor~\cite{zhuHarborCollaborativeVehicular2024} combines V2I access with opportunistic V2V relaying and partitions the workload between vehicles and an edge node.

Task utility provides a second coupling dimension~\cite{fangPrioritizedInformationBottleneck2025, fangTaskOrientedCommunicationsVisual2025}. Q-CPTO~\cite{zakiQualityAwareTask2024} uses region-of-interest coverage and value of information to select perception tasks under latency constraints. ITORA~\cite{dongTaskOffloadingResource2025} jointly optimizes sensing-data selection, vehicle participation, task offloading, bandwidth, computing resources, and energy. These systems show that radio optimization alone can leave queueing and inference as dominant delays, while aggressive offloading can increase handover and availability risk. Communication-computing co-design therefore requires joint control of message utility, model partitioning, server selection, compute load, and end-to-end deadlines. The resulting policy must preserve a local fallback when remote outputs become stale or computation becomes unavailable.

\subsubsection{Limitations of Explicit Communication}
Explicit communication is mature and deployable, but its operational contribution depends on message type and downstream tasking. BSMs, CAMs, and DENMs mainly provide awareness. CPMs can support partial state alignment when received messages for objects are fused, validated, and time-aligned. MCMs and status-intent messages can support partial intent alignment when planners consume the future maneuver information. Coordinated downstream action requires planner and controller integration beyond these message families. Dense traffic, packet loss, pose error, stale messages, and malicious packets still motivate the implicit, semantic, and language-based paradigms discussed below.

\subsection{Implicit Communication (Behavior-Driven)}

\subsubsection{Trajectory Game Theory and Intent Inference}
Implicit communication occurs when an agent reads another agent's motion as a behavioral signal. Game-theoretic interaction models~\cite{fisac2019hierarchical} formalize this setting by treating trajectories as evidence about hidden intent. A slight deceleration, for example, can signal yielding intent. Recent game-theoretic driving studies use this logic to select actions that balance self-interest with clear intention signaling~\cite{huang2024game, yuan2024game}.

Trajectory-prediction methods clarify what can be inferred from behavior before explicit messages are exchanged. EqMotion~\cite{eqmotion2023} uses equivariant prediction to keep motion forecasts consistent across coordinate frames. HiVT~\cite{zhou2022hivt} uses hierarchical vector transformers for interaction encoding. Wayformer~\cite{nayakanti2023wayformer} uses factorized attention, and MTR++~\cite{shi2024mtr++} uses guided intention querying for multimodal motion prediction. DenseTNT~\cite{gu2021densetnt} models goal candidates, THOMAS~\cite{gilles2022thomas} learns multi-agent trajectory sampling, and FJMP~\cite{rowe2023fjmp} models joint futures that expose likely intent. These methods provide observer-side intent evidence, while coordinated control requires prediction sharing, shared-state construction, or planner-level use.

\subsubsection{Emergent Communication and Learned Signaling}
Emergent communication describes learned signaling channels that arise during multi-agent training. CommNet~\cite{sukhbaatarLearningMultiagentCommunication2016} learns continuous messages jointly with agent policies, and DIAL~\cite{foersterLearningCommunicateDeep2016} jointly trains message generation and interpretation by allowing gradients to propagate through continuous inter-agent messages during centralized training. In autonomous-driving simulation, MA-MeSN/MA-BoN~\cite{bhallaDeepMultiAgent2020} provide an early example of communication-aware MARL. Learned message channels differ from reward-induced road-rule emergence, where yielding or alternating passage can arise from interaction rewards without an explicit learned message channel~\cite{pal2020emergent, bhattacharyya2019simulating}. For MAEAD, emergent communication is relevant when learned signals expose intent or affect downstream action. Its evidence remains limited by weak interpretability, protocol instability, and safety auditing difficulty.

\subsubsection{Non-Radio Signaling and Visual Interpretation}
Beyond trajectory analysis, agents can use physical signals such as turn indicators, brake lights, pedestrian pose, gaze, and crossing motion to infer intent. Pedestrian-intention datasets such as JAAD~\cite{rasouliAreTheyGoing2017} and PIE~\cite{rasouliPIELargeScale2019} show that crossing behavior requires combining motion history, visual context, and appearance cues. IntentNet~\cite{casasIntentNetLearning2018} extends intent prediction to raw sensor data by jointly estimating high-level behaviors and future trajectories. These cues provide implicit intent evidence, while shared state construction and planner-level use remain separate requirements.

\subsection{Semantic \& Intent Communication}

\subsubsection{Semantic Communication: Transmitting Meaning Over Bits}
Semantic communication prioritizes message meaning over raw bits~\cite{sheng2024semantic}. This is implemented through shared knowledge between the transmitter and receiver to encode and decode information more efficiently. For example, a full LiDAR point cloud can be replaced by the semantic concept ``pedestrian crossing the street.'' This approach, often leveraging techniques such as deep joint source-channel coding (DJSCC)~\cite{bourtsoulatzeDeepJointSource2019}, can reduce communication overhead while preserving essential information~\cite{ye2025survey, lyu2024semantic, gimenez2024semantic}.

Recent work has advanced semantic communication for V2V collaborative perception along three aspects. At the feature level, importance-based semantic extraction uses learned saliency maps to prioritize informative regions for transmission~\cite{sheng2024semantic}. At the resource level, Confidence-V2X~\cite{confidencev2x2025} uses local confidence as a signal for what should be refined, transmitted, and fused. At the reasoning level, V2X-UniPool~\cite{v2xunipool2025} transforms multimodal V2X data into a time-indexed language knowledge pool and uses retrieval-augmented reasoning for planning. These methods provide partial alignment mechanisms, but end-to-end closed-loop validation remains limited.

\subsubsection{LLM-based Interaction and Negotiation}
Large language models (LLMs) are becoming relevant to semantic communication because they can expose intentions, constraints, and proposed maneuvers in a human-recognizable form~\cite{cui2024talking}. Their role in this section is a communication-interface role: agents can summarize local context, state intended maneuvers, critique proposed actions, and exchange negotiation traces that downstream planners can check. CoopReflect~\cite{coopreflect2026} directly targets natural-language V2V messages through multi-agent learning and debriefing. CoLMDriver~\cite{liu2025colmdriver} uses critic-guided negotiation to exchange and revise candidate actions before waypoint grounding. AgentsCoMerge~\cite{huAgentsCoMergeLargeLanguage2024} uses language dialogue for cooperative ramp merging, and V2X-UniPool~\cite{v2xunipool2025} grounds language reasoning in a time-indexed V2X knowledge pool. CoMAL~\cite{yao2024comal} and Agent-Driver~\cite{maoLanguageAgentAutonomous2023} show how memory, role assignment, and reasoning traces can structure high-level maneuver discussion. Together, these methods make intent more inspectable. System-level deployment still requires shared state construction, timing alignment, controller grounding, and safety checks.

\subsubsection{Vision-Language Grounding for Shared Scene Understanding}
Vision language models (VLMs) connect language-level coordination with physical scene evidence. DriveGPT4~\cite{xuDriveGPT4InterpretableEndtoend2023} and DriveLM~\cite{simaDriveLMDrivingGraph2023} link visual inputs, scene graphs, or bird's-eye-view representations with textual reasoning about traffic context. In MAEAD, this grounding can help agents convert shared perception into inspectable semantic context. Their evidence primarily supports scene grounding. Cross-agent consistency still depends on V2X fusion, timestamp alignment, calibration checks, and agreement tests across viewpoints. Verifiable coordination and control remain separate requirements.

\subsubsection{Plan-Level Communication}
Plan-level communication shifts the exchanged content from perceptual evidence toward compact representations of intended future actions. By sharing intended trajectories or plan summaries, agents can expose goals and constraints with lower bandwidth than raw perception streams. This approach can make future states more predictable to receivers and can degrade more gracefully under packet loss than pixel-level sharing.

CMP~\cite{wang2025cmp} studies cooperative motion prediction with multi-agent communication, where the exchanged representation is tied to future behavior. Intent-aware planning via distributed MARL~\cite{wu2023intent} shows a complementary mechanism in which inferred or exchanged intent affects policy selection. In this section, these papers motivate plan-level communication as compact intent summaries. Their full planning role is discussed in Section~\ref{sec:decision}.

\subsection{Communication Paradigms and Operational Criteria}

The preceding subsections show a progression from standardized messages to inferred behavior, compressed meaning, language negotiation, and plan summaries. Table~\ref{tab:cognition_comm_methods} lists representative standards and methods that anchor this progression. Complete autonomous-driving-system claims require additional perception, planning, and control evidence.

In summary, this section moves the survey from transport-level exchange toward richer intent communication. Explicit V2X messages provide structured information that onboard systems can directly process, while shared-state alignment requires additional validation beyond message availability. Implicit behavior inference estimates hidden intent, semantic communication compresses meaning, and language-based or plan-level communication exposes proposed actions. The remaining gap is converting negotiated or inferred intent into safe coordinated motion under real-time constraints.

\section{Collaborative Decision-Making \& Planning}
\label{sec:decision}

The transition from single-agent autonomy to multi-agent embodied autonomous driving changes the decision-making and planning problem substantially. While early approaches relied on V2X-based data exchange, recent methods increasingly model intent, role assignment, coupled feasibility, and coordinated downstream action. This section reviews collaborative decision-making from game-theoretic frameworks and multi-agent reinforcement learning (MARL) to recent generative and foundation-model-based planning methods.

\subsection{Game-Theoretic and Optimization-Based Planning}

Section~\ref{sec:cognition} treated trajectory-level game-theoretic models as implicit communication, where observed motion provides evidence about hidden intent. This subsection shifts the focus from observer-side inference to decision synthesis. Game-theoretic planners specify utility, belief, role, or response models, then search for actions that remain compatible with the predicted responses of other road users. Table~\ref{tab:game_theoretic_methods} lists representative methods and related cooperative-optimization bridges, including inverse reinforcement learning (IRL), iterative linear quadratic regulator (iLQR), and alternating direction method of multipliers (ADMM).

\begin{table}[t]
\centering
\caption{Representative Game-Theoretic and Optimization-Linked Planning Methods for MAEAD.}
\vspace{-0.5em}
\footnotesize

\label{tab:game_theoretic_methods}
\begin{tabular}{lll}
\toprule
\textbf{Method} & \textbf{Venue} & \textbf{Primary Mechanism} \\
\midrule
\multicolumn{3}{@{}l}{\emph{Equilibrium and Hierarchical Games}} \\
HGT planning~\cite{fisac2019hierarchical} & 2019 ICRA & Stackelberg response model \\
MT-GTF~\cite{yan2023multi} & 2023 TIV & Game-theoretic planning \\
N/S selection~\cite{huang2024game} & 2024 IFAC-PapersOnLine & Equilibrium model selection \\
GTDM~\cite{yuan2024game} & 2024 Book Ch. & Nash, Stackelberg, coalition games \\
\midrule
\multicolumn{3}{@{}l}{\emph{Scalable and Learned Game Models}} \\
SIS-CM~\cite{wang2022multi} & 2022 TIV & Delay-aware conflict management \\
LEAD~\cite{huang2025lead} & 2025 TITS & IRL-enhanced game model \\
\midrule
\multicolumn{3}{@{}l}{\emph{Incentives and Cooperative Optimization}} \\
FT-Intersection~\cite{pei2022fault} & 2022 TIV & Fault-tolerant cooperation \\
FL-Coop~\cite{cai2022multi} & 2022 TVT & Formation-control cooperation \\
CoopForm~\cite{li2022cooperative} & 2022 TITS & Formation-level incentives \\
LF-Nudging~\cite{yanumula2023optimal} & 2023 TIV & Nudging-based planning \\
iLQR/ADMM~\cite{huang2023decentralized} & 2023 TITS & Decentralized trajectory optimization \\
HC-Parallel~\cite{huang2024parallel} & 2024 ICRA & Safety-constrained cooperation \\
C-ADMM~\cite{liu2024improved} & 2024 TIV & Limited-communication planning \\
\midrule
\multicolumn{3}{@{}l}{\emph{Computation-Aware Cooperative Planning}} \\
HMPC~\cite{sunCooperativeMergingMixed2025} & 2025 TITS & Edge-assisted merge control \\
DCIMPC~\cite{chuV2XAssistedDistributed2026} & 2026 TMC & Distributed trajectory control \\
\bottomrule
\end{tabular}
\vspace{-1em}
\end{table}

\subsubsection{Equilibrium Models for Interactive Driving}

Nash equilibrium (NE) fits interactive scenarios where agents choose mutually responsive actions, such as lane merging, unprotected turns, and roundabout negotiation. The Nash/Stackelberg comparison~\cite{huang2024game} highlights that the proper equilibrium model depends on whether the interaction is closer to simultaneous negotiation or sequential commitment. Classical NE remains limited due to strong rationality assumptions and search difficulty in continuous, high-dimensional driving spaces.

When the interaction has a clearer order of commitment, Stackelberg games~\cite{fisac2019hierarchical, yuan2024game} provide a more natural leader-follower model for mixed-autonomy traffic. Hierarchical game-theoretic planning~\cite{fisac2019hierarchical} uses this structure when an autonomous vehicle (AV) commits to a maneuver and another road user responds. The central modeling burden is the follower model: the planner must capture behavioral intent, implicit utility, and response delay beyond kinematics alone. Coalition game-based models add another option for mixed-motive settings by allowing vehicles to form temporary groups and allocate utility across platoons or local traffic coalitions~\cite{yuan2024game}. Mixed-traffic game frameworks~\cite{yan2023multi} make this taxonomy concrete by linking equilibrium structure to scenario topology and agent type.

\subsubsection{Scalable and Learned Game Models}

Large traffic populations create a scalability problem for equilibrium computation because explicit strategic reasoning over every participant becomes intractable. Mean-field game (MFG) theory addresses this issue by approximating the population through an aggregate traffic distribution. The representative agent optimizes against this distribution, which makes large-scale density regulation and signal-free intersection analysis more tractable~\cite{yuan2024game}. In the MAEAD framing, MFG-style methods are most useful for population-level coordination, while explicit intent, role, or plan variables remain necessary for interpretable negotiation.

Learned game models address a different weakness: manually designed utility functions may miss context-dependent human behavior. Data-driven game-theoretic approaches infer rewards or response parameters from driving data. LEAD~\cite{huang2025lead} is a recent example in dynamic merging, combining a non-cooperative game model with maximum-entropy inverse reinforcement learning (IRL) and online parameter adaptation. We treat reported behavior matching and safety results as scenario-specific evidence, with general deployment guarantees left unclaimed.

\subsubsection{Incentives and Cooperative Planning Bridges}

Social dilemmas arise when individually rational driving choices reduce collective welfare, for example when every vehicle protects ego progress in a merge or intersection. The social dilemma of autonomous vehicles~\cite{bonnefon2016social} is useful here as motivation for welfare conflict and public acceptance. Operational cooperative-driving mechanisms still require reward, constraint, or protocol design that can be evaluated in traffic scenarios. Cooperative formation~\cite{li2022cooperative} and related studies on fault-tolerant intersections~\cite{pei2022fault}, flexible-lane cooperation~\cite{cai2022multi}, lane-free nudging~\cite{yanumula2023optimal}, and delay-aware status-and-intent sharing~\cite{wang2022multi} show how incentives and constraints can move from ethical framing to executable coordination.

For structured connected and autonomous vehicle (CAV) settings, distributed iLQR/ADMM~\cite{huang2023decentralized}, improved consensus ADMM~\cite{liu2024improved}, and hard-constrained parallel optimization~\cite{huang2024parallel} provide a bridge from strategic interaction models to numerically tractable collaborative planning. Their value is strongest when the system needs decentralized execution under limited communication, explicit feasibility constraints, and safety-aware trajectory updates. Recent computation-aware planners extend this bridge across edge and vehicle processors. HMPC~\cite{sunCooperativeMergingMixed2025} places merge-sequence optimization at the edge and trajectory planning on vehicles. DCIMPC~\cite{chuV2XAssistedDistributed2026} decomposes collaborative trajectory optimization and model predictive control for parallel execution across CAVs. These methods expose computation placement through explicit planning and control interfaces, although their evidence remains scenario-specific. Together, game-theoretic and optimization-based methods connect interaction assumptions, feasibility constraints, and computation placement with executable coordination. Scalability and transfer remain shared challenges, while game-theoretic models face additional sensitivity to utility specification and bounded-rationality assumptions. MARL relaxes some manually specified assumptions by learning policies through interaction, which motivates the next subsection.

\subsection{Multi-Agent Reinforcement Learning (MARL)}

Multi-agent reinforcement learning (MARL) moves collaborative driving from manually designed strategic models to interaction-based policy learning. This makes it a natural complement to game-theoretic planning: agents can learn coupled policies from rewards, observations, and repeated interactions, with less dependence on a fully specified utility model for every road user~\cite{dinnewethMultiagentReinforcementLearning2022}. Its relevance to MAEAD depends on whether learned coordination exposes aligned belief states, exchanged messages, intent variables, or plan-level coordination, with task reward improvement treated as insufficient evidence by itself. Table~\ref{tab:marl_comparison} lists representative MARL and cooperative policy-learning methods. Centralized training with decentralized execution (CTDE) denotes the common training pattern that uses global information during learning while preserving decentralized policies at execution time.

\begin{table}[t]
\centering
\caption{Representative MARL and Cooperative Policy-Learning Methods for MAEAD.}
\vspace{-0.5em}
\footnotesize

\label{tab:marl_comparison}
\begin{tabular}{lll}
\toprule
\textbf{Method} & \textbf{Venue} & \textbf{Primary Mechanism} \\
\midrule
\multicolumn{3}{@{}l}{\emph{General MARL and Driving Foundations}} \\
Safe MARL~\cite{shalev-shwartzSafeMultiAgentReinforcement2016} & 2016 arXiv & Safety-constrained policy \\
MADDPG~\cite{lowe2020multiagentactorcriticmixedcooperativecompetitive} & 2017 arXiv & Centralized critic \\
AD-MARL~\cite{bhallaDeepMultiAgent2020} & 2020 Advances in AI & Emergent communication \\
MAPPO~\cite{yu2021mappo} & 2022 NeurIPS & Policy-gradient CTDE \\
\midrule
\multicolumn{3}{@{}l}{\emph{Driving-Specific Cooperative Control}} \\
LC-MADRL~\cite{zhang2022multi} & 2022 TITS & Right-of-way collaboration \\
CAV-ORM~\cite{9316925} & 2022 TITS & Lane-selection policy \\
SAC planning~\cite{tang2022highway} & 2022 TVT & Soft actor-critic planning \\
CARLA-RL~\cite{gutierrez2022reinforcement} & 2022 Sensors & Intersection policy learning \\
OR-MARL~\cite{chenDeepMultiagentReinforcement2022} & 2023 TITS & Scalable merging policy \\
IA-DMARL~\cite{wu2023intent} & 2023 CoRL & Intent-aware planning \\
SAPO~\cite{dai2023socially} & 2023 CoRL & Social attention policy \\
CPPO~\cite{peng2023curriculum} & 2023 ITSC & Staged intersection learning \\
AutoDRIVE-MARL~\cite{AutoDRIVE-MARL-2023} & 2023 arXiv & Cooperative-competitive driving \\
IA-DM~\cite{chen2023interaction} & 2023 IEEE TTE & Interaction-aware policy \\
Eco-ORM~\cite{liuEcoFriendlyOnRampMerging2023} & 2023 TVT & Energy-aware merging \\
RS-DRL~\cite{xiao2024decision} & 2024 TVT & Intersection task policy \\
MARL-PPO~\cite{kolat2024cooperative} & 2024 MDPI & Highway platooning \\
AGP-MARL~\cite{liu2024cooperative} & 2024 TITS & Intersection cooperation \\
CE-MARL~\cite{hua2024communication} & 2025 TVT & Bandwidth-aware platooning \\
MAMQPPO~\cite{mamqppo2025} & 2025 WEVJ & Platoon merge control \\
RSU-HRL~\cite{marlsmart2025} & 2025 arXiv & Infrastructure-anchored policy \\
MM-MARL~\cite{chen2025mixed} & 2025 JAS & Social motivation learning \\
\midrule
\multicolumn{3}{@{}l}{\emph{Belief, Communication, and Safety Extensions}} \\
ST-Safe MARL~\cite{zhang2023spatial} & 2023 ICRA & Spatio-temporal safety \\
STL-MARL~\cite{wang2023multi} & 2023 arXiv & Logic-guided rewards \\
Diverse-MVP~\cite{wen2023bringing} & 2023 AAMAS & Interpretable policy diversity \\
MAPPO-PIS~\cite{guo2024mappo} & 2024 arXiv & Prior intent sharing \\
PDM~\cite{xia2024parameterized} & 2024 ICDE & Multimodal decision parameters \\
SMPE~\cite{smpe2025} & 2025 ICML & Belief-state learning \\
\bottomrule
\end{tabular}%
\vspace{-1em}
\end{table}

\subsubsection{CTDE and Communication-Aware MARL}

The core training challenge is non-stationarity: from one agent's perspective, other learning agents continuously change the environment. CTDE addresses this by using global information during training while preserving decentralized policies for execution. MADDPG~\cite{lowe2020multiagentactorcriticmixedcooperativecompetitive} extends actor-critic learning with a centralized critic over joint actions and observations. MAPPO~\cite{yu2021mappo} provides a CTDE baseline in the style of proximal policy optimization (PPO)~\cite{schulman2017proximalpolicyoptimizationalgorithms}, which remains common in cooperative control.

Driving-specific MARL adapts the CTDE framework through topology encoding, communication control, and reward structure~\cite{chenDeepMultiagentReinforcement2022, zhang2022multi, zhang2023spatial, wang2023multi}. Graph RL for mixed-autonomy traffic~\cite{liu2022graph} organizes nearby vehicles as an interaction graph. Interaction-aware decision-making~\cite{britoLearningInteractionawareGuidance2021} uses interaction context for autonomous-vehicle decisions. Communication-aware MARL~\cite{dingRobustMultiAgentCommunication2024, yuRobustCommunicativeMultiAgent2024} learns robust message representations or down-weights unreliable messages under noisy or adversarial communication. These mechanisms strengthen downstream coordination, while explicit shared-state variables, communicated intent, role assignment, or plan messages are still needed for auditable multi-agent reasoning.

\subsubsection{Driving Scenarios and Belief-State Extensions}

MARL evidence in autonomous driving clusters around a few recurring scenarios. For unsignalized intersections, Curriculum PPO~\cite{peng2023curriculum} uses staged learning, while attention and hierarchical game priors~\cite{liu2024cooperative} combine learned policies with structured interaction cues. RSU Hybrid RL~\cite{marlsmart2025} uses infrastructure as coordination anchors, with offline pre-training followed by MAPPO fine-tuning for smart intersections. For lane changes and highway control, lane-change MARL~\cite{zhang2022multi} incorporates right-of-way collaboration, and highway soft actor-critic planning~\cite{tang2022highway} shows how learned policies can support tactical driving decisions.

Merging and platooning form another application area. Cooperative MARL-PPO~\cite{kolat2024cooperative} uses PPO-style learning for highway platooning. MAMQPPO~\cite{mamqppo2025} studies autonomous vehicle merging into a platoon with a partially decoupled reward function and a highway simulation setup. Its reported efficiency and energy gains should be interpreted within that simulation protocol, with cross-platooning generalization left unclaimed. Mixed-motive MARL~\cite{chen2025mixed} adds a social-learning objective for traffic involving competing goals.

Safety and belief-state extensions are especially relevant to MAEAD because vehicles observe different occlusions, poses, and traffic histories. Safe MARL for CAVs~\cite{zhang2023spatial} adds spatio-temporal safety structure. Signal temporal logic (STL)-guided MARL~\cite{wang2023multi} uses formal logic specifications to shape multi-agent behavior. MAPPO-PIS~\cite{guo2024mappo} adds prior intent sharing for cooperative CAV decision-making. SMPE~\cite{smpe2025} studies cooperative MARL through belief-state modeling and adversarial exploration. This makes it relevant to MAEAD as a transferable mechanism, because partial observability is a core reason that agents need aligned state estimates before coordinated action.

\subsubsection{Limitations for MAEAD: Credit, Observability, and Transfer}

MARL exposes three recurring limitations for MAEAD. The first is credit assignment: cooperative driving rewards often depend on a sequence of joint actions, so the learner must estimate how much each vehicle contributes to safety, progress, or smoothness. Centralized critics and PPO-style surrogates provide common tools for this problem, and they help explain why CTDE baselines built on MADDPG~\cite{lowe2020multiagentactorcriticmixedcooperativecompetitive} and MAPPO~\cite{yu2021mappo} remain common.

The second limitation is partial observability. Vehicles observe different occlusions, poses, and local traffic histories, and communication may be delayed or missing. Memory, attention, selective communication, and belief-state estimation can reduce uncertainty, but benchmark reward improvement alone does not show explicit state alignment or intent agreement. The third limitation is transfer across scenarios, agent populations, communication conditions, and physical deployment. Policies trained in a simulator or a narrow scenario may encode coordination conventions tied to reward shaping, scripted non-ego behavior, fixed agent populations, or idealized communication. Changes in these conditions can invalidate the learned conventions, causing inconsistent intent interpretation, unstable joint actions, or degraded safety. Reward-induced coordination therefore provides scenario-bounded evidence. Emergent road-rule work~\cite{pal2020emergent} and reward-augmented imitation studies~\cite{bhattacharyya2019simulating} demonstrate yielding, alternating passage, and socially patterned motion within their evaluated simulations, while transfer to unseen traffic configurations and physical deployment remains open.

\subsection{Generative and Language-Grounded Planning}

Generative and foundation-model-based planning methods extend the decision layer from selecting one action to reasoning over multiple plausible futures, candidate plans, and high-level constraints. Their relevance depends on whether predicted intent, explicit plans, or generated futures can be grounded in coordinated downstream actions. Table~\ref{tab:planning_generative_language} lists representative generative and language-grounded planning methods and core mechanism. Model predictive control (MPC) denotes the control formulation referenced by LanguageMPC in this section.

\begin{table}[t]
\centering
\caption{Representative Generative and Language-Grounded Planning Methods for MAEAD.}
\vspace{-0.5em}
\footnotesize
\label{tab:planning_generative_language}
\resizebox{0.7\textwidth}{!}{%

\begin{tabular}{lll}
\toprule
\textbf{Method} & \textbf{Venue} & \textbf{Primary Mechanism} \\
\midrule
\multicolumn{3}{@{}l}{\emph{Generative Trajectory and World-Model Planning}} \\
MotionDiffuser~\cite{jiang2023motiondiffuser} & 2023 CVPR & Diffusion multimodal trajectories \\
CTG++~\cite{zhong2023ctg} & 2023 CoRL & Guided-diffusion scenario synthesis \\
GAIA-1~\cite{gaia2023} & 2023 arXiv & Generative driving world model \\
DriveDreamer~\cite{drivedreamer2024} & 2024 ECCV & Real-world-driven world model \\
Diffusion Planner~\cite{diffusionplanner2025} & 2025 ICLR & Guided closed-loop diffusion planning \\
DriveGen~\cite{zhang2025drivegeninfinitediversetraffic} & 2025 IROS & Diverse traffic generation \\
CogDrive~\cite{huang2025cogdrive} & 2025 arXiv & Prediction-coupled trajectory optimization \\
GAIA-2~\cite{gaia2_2025} & 2025 arXiv & Controllable multi-view generation \\
MUVO~\cite{muvo2025} & 2025 IV & Multimodal occupancy world model \\
COMBO~\cite{combo2025} & 2025 ICLR & Compositional multi-agent world model \\
Epona~\cite{epona2025} & 2025 ICCV & Autoregressive diffusion world model \\
UniDriveDreamer~\cite{unidrivedreamer2026} & 2026 arXiv & Camera-LiDAR future generation \\
HERMES++~\cite{hermespp2026} & 2026 arXiv & 3D understanding with future geometry \\
WorldDrive~\cite{worlddrive2026} & 2026 arXiv & Unified vision-motion planning \\
ShareVerse~\cite{shareverse2026} & 2026 arXiv & Multi-agent-consistent video generation \\
SCORP~\cite{scorp2026} & 2026 arXiv & Scene-consistent cooperative denoising \\
\midrule
\multicolumn{3}{@{}l}{\emph{Language-Grounded Planning Interfaces}} \\
LanguageMPC~\cite{languagempc2023} & 2023 arXiv & Language-to-MPC control \\
Agent-Driver~\cite{maoLanguageAgentAutonomous2023} & 2024 CoLM & Structured reasoning traces \\
DriveGPT4~\cite{xuDriveGPT4InterpretableEndtoend2023} & 2024 RA-L & Interpretable control generation \\
DiLu~\cite{wenDiLuKnowledgeDrivenApproach2023} & 2024 ICLR & Memory and reflection \\
LMDrive~\cite{shaoLMDriveClosedLoopEndtoEnd2023} & 2024 CVPR & Instruction-guided closed-loop driving \\
DriveLM~\cite{simaDriveLMDrivingGraph2023} & 2024 ECCV & Graph VQA planning proxy \\
DriveVLM~\cite{tianDriveVLMConvergenceAutonomous2024} & 2024 CoRL & Chain-of-thought scene planning \\
AsyncDriver~\cite{chenAsynchronousLargeLanguage2024} & 2024 ECCV & Asynchronous LLM guidance \\
RDA-Driver~\cite{rdadriver2024} & 2024 ECCV & Reasoning-decision alignment \\
AgentsCoDriver~\cite{huAgentsCoDriverLargeLanguage2024} & 2024 arXiv & Social-reasoning LLM planner \\
KoMA~\cite{koma2024} & 2024 TIV & Knowledge-driven planning \\
CoMAL~\cite{yao2024comal} & 2025 SDM & Memory and role assignment \\
V2X-VLM~\cite{youV2XVLMEndtoEndV2X2024} & 2025 TR-C & V2X VLM trajectory planning \\
OmniDrive~\cite{wangOmniDriveHolisticLLMAgent2024} & 2025 CVPR & 3D reasoning and planning \\
AgentsCoMerge~\cite{huAgentsCoMergeLargeLanguage2024} & 2025 TMC & Ramp-merging dialogue \\
CoLMDriver~\cite{liu2025colmdriver} & 2025 ICCV & LLM negotiation with grounding \\
V2X-UniPool~\cite{v2xunipool2025} & 2025 arXiv & Language reasoning over V2X evidence \\
DriveAgent~\cite{driveagent2025} & 2025 RA-L & Structured LLM planner \\
LangCoop~\cite{langcoop2025} & 2025 CVPRW & Language-packaged collaboration \\
CoopReflect~\cite{coopreflect2026} & 2026 AAMAS & Learning and debriefing \\
PlanAgent~\cite{zhengPlanAgentMultimodalLarge2024} & 2026 TCDS & Multimodal LLM code planner \\
CAPE~\cite{cape2026} & 2026 arXiv & Checked path-editing programs \\
\midrule
\multicolumn{3}{@{}l}{\emph{VLA Models and Shared Predictive Latents}} \\
EMMA~\cite{hwangEMMAEndtoEndMultimodal2024} & 2024 TMLR & End-to-end multimodal planning \\
Senna~\cite{jiangSennaBridgingLargeVisionLanguage2024} & 2024 arXiv & VLM-to-driving bridge \\
VLM-AD~\cite{xuVLMADEndtoEndAutonomous2024} & 2025 CoRL & VLM-supervised driving \\
LCDrive~\cite{lcdrive2025} & 2025 arXiv & Latent chain-of-thought world modeling \\
AutoVLA~\cite{zhouAutoVLAVisionLanguageAction2025} & 2025 NeurIPS & Adaptive VLA reasoning \\
DriveMoE~\cite{yangDriveMoEMixtureofExperts2025} & 2026 CVPR & VLA mixture-of-experts \\
DriveWorld-VLA~\cite{driveworldvla2026} & 2026 arXiv & Latent world-state planning \\
Uni-World VLA~\cite{uniworldvla2026} & 2026 arXiv & Interleaved prediction and planning \\
VLA-World~\cite{vlaworld2026} & 2026 arXiv & Future-scene trajectory refinement \\
CoWorld-VLA~\cite{coworldvla2026} & 2026 arXiv & Multi-expert world model \\
\bottomrule
\end{tabular}
}
\vspace{-1em}
\end{table}

\subsubsection{Generative Trajectory and SWM Planning}

Diffusion and generative trajectory models are useful because multi-agent planning is inherently multimodal. MotionDiffuser~\cite{jiang2023motiondiffuser} provides a concrete example of diffusion-based multimodal trajectory generation in multi-agent scenes. CTG++~\cite{zhong2023ctg} uses guided diffusion for constraint-conditioned scenario synthesis. Diffusion Planner~\cite{diffusionplanner2025} moves diffusion closer to closed-loop planning through flexible guidance, while DriveGen~\cite{zhang2025drivegeninfinitediversetraffic} extends this data-engine role by synthesizing diverse traffic scenarios for autonomous-driving evaluation. CogDrive~\cite{huang2025cogdrive} couples multimodal prediction with safety-aware trajectory optimization, which makes the prediction-to-planning link more explicit. Their planning evidence remains the strongest when generated trajectories are evaluated inside closed-loop cooperative control.

World models add a predictive state-transition layer that can be queried before action execution. GAIA-1~\cite{gaia2023} and DriveDreamer~\cite{drivedreamer2024} show how video, text, action, and structured traffic constraints can support driving-video generation and scenario testing. GAIA-2~\cite{gaia2_2025} adds controllable multi-view generation conditioned on ego dynamics, agent configurations, environment factors, and road semantics. Epona~\cite{epona2025}, UniDriveDreamer~\cite{unidrivedreamer2026}, and HERMES++~\cite{hermespp2026} extend this direction toward autoregressive diffusion, camera-LiDAR future generation, and 3D future geometry. MUVO~\cite{muvo2025} anchors multimodal occupancy-oriented world modeling, while COMBO~\cite{combo2025} is adjacent multi-agent evidence because it makes partial observation, interaction dynamics, and intent inference explicit planning objects.

For MAEAD, the planning value of world models is counterfactual reasoning. A planner can ask how another agent may react to a proposed maneuver, whether a joint plan creates unsafe future states, and which observations are most useful before action. WorldDrive~\cite{worlddrive2026} directly links scene generation with planning through unified vision-motion representations. ShareVerse~\cite{shareverse2026} targets multi-agent-consistent video generation for shared world modeling, and SCORP~\cite{scorp2026} couples scene-consistent multi-agent denoising with online reinforcement post-training for cooperative driving. These methods support the SWM argument, with deployment relevance depending on behavioral realism, sensor realism, and closed-loop communication assumptions. Section~\ref{sec:e2e} summarizes the E2E evidence boundary for motion sequence modeling and shared predictive latents.

\subsubsection{Language-Grounded Planning Interfaces}
\label{sec:fm-limitations}

Large language models (LLMs) are being explored as high-level reasoning interfaces for autonomous driving. Their main value in MAEAD is that they can express traffic rules, intended maneuvers, role assignments, critiques, and negotiation traces in a form that other agents or safety layers can inspect. Their outputs still require grounding in perception, timing, vehicle dynamics, and controller-level constraints before they can support safety-critical action.

For LLM planners, the useful organization separates four mechanisms. Language negotiation makes intended actions explicit and allows agents to revise proposals after feedback. CoLMDriver~\cite{liu2025colmdriver} supports this mechanism through InterDrive/CARLA evaluation with critic-guided negotiation and waypoint grounding. AgentsCoMerge~\cite{huAgentsCoMergeLargeLanguage2024}, LangCoop~\cite{langcoop2025}, and CoopReflect~\cite{coopreflect2026} extend language communication toward ramp merging, compact language packages, and learned message debriefing. CoMAL~\cite{yao2024comal} uses memory and role assignment. LanguageMPC~\cite{languagempc2023} connects language output to MPC-style control. Agent-Driver~\cite{maoLanguageAgentAutonomous2023} uses structured reasoning traces, and CAPE~\cite{cape2026} uses checked path-editing programs. V2X-UniPool~\cite{v2xunipool2025} connects language reasoning to real-time V2X evidence.

These mechanisms make intent and planning traces more visible, but visibility should not be confused with certified safety. DiLu~\cite{wenDiLuKnowledgeDrivenApproach2023}, LMDrive~\cite{shaoLMDriveClosedLoopEndtoEnd2023}, DriveGPT4~\cite{xuDriveGPT4InterpretableEndtoend2023}, DriveVLM~\cite{tianDriveVLMConvergenceAutonomous2024}, PlanAgent~\cite{zhengPlanAgentMultimodalLarge2024}, AsyncDriver~\cite{chenAsynchronousLargeLanguage2024}, RDA-Driver~\cite{rdadriver2024}, and OmniDrive~\cite{wangOmniDriveHolisticLLMAgent2024} show different ways to connect language or vision-language reasoning to closed-loop control, asynchronous guidance, reasoning-decision alignment, or 3D planning. CoLMDriver~\cite{liu2025colmdriver} and DriveAgent~\cite{driveagent2025} are examples of structured LLM planners. CoMAL~\cite{yao2024comal} adds role memory. AgentsCoDriver~\cite{huAgentsCoDriverLargeLanguage2024} focuses on social reasoning, and KoMA~\cite{koma2024} uses knowledge-driven planning. These systems are mainly validated in benchmark or simulation settings. Multi-step reasoning pipelines can compound inference delay, and latency comparisons require documented or reproduced hardware settings. LLMs can also generate plausible scene descriptions or negotiation proposals that conflict with geometry, map constraints, or agent identity. Foundation models are therefore most likely to enter MAEAD planning as advisory components operating alongside conventional safety filters and certified controllers.

\subsubsection{VLA Models and Shared Predictive Latents}

Vision language action (VLA) models are relevant to this planning section only when they expose a mechanism for choosing actions from predictive scene representations. EMMA~\cite{hwangEMMAEndtoEndMultimodal2024}, Senna~\cite{jiangSennaBridgingLargeVisionLanguage2024}, VLM-AD~\cite{xuVLMADEndtoEndAutonomous2024}, AutoVLA~\cite{zhouAutoVLAVisionLanguageAction2025}, and DriveMoE~\cite{yangDriveMoEMixtureofExperts2025} represent the broader VLA moves toward action prediction from language-supervised or multimodal latent states. DriveWorld-VLA~\cite{driveworldvla2026} uses latent world states as decision variables for a VLA planner, and Uni-World VLA~\cite{uniworldvla2026} interleaves future-frame prediction with trajectory planning. VLA-World~\cite{vlaworld2026} pairs future-scene imagination with reflective trajectory refinement, CoWorld-VLA~\cite{coworldvla2026} conditions action generation on semantic interaction, geometry, dynamics, and expert tokens, and LCDrive~\cite{lcdrive2025} uses action-aligned latent reasoning tokens grounded in future rollouts.

These systems are mainly ego-centric planning and world-model prediction. Their role here is to clarify a mechanism direction for MAEAD: intent modeling and action selection may eventually be learned inside a shared predictive latent space. Section~\ref{sec:e2e} discusses their end-to-end evidence boundary in more detail.

\subsection{Agreement and Safety Gaps for Deployment}

The planning methods above show how agents can infer intent, negotiate proposals, and optimize coupled actions. Deployment requires a narrower test: whether the proposed maneuver remains compatible with other agents' plans, local vehicle limits, communication timing, and a verifiable safety envelope. We use agreement to mean this operational compatibility, distinct from formal consensus convergence.

\subsubsection{Plan Agreement and Coordinated Execution}

Plan agreement requires observable mechanisms for conflict detection, role or priority assignment, feasibility checking, and fallback when agreement fails. The distributed optimization bridges reviewed above provide examples because they expose coupling constraints and decentralized trajectory updates. SMART~\cite{huang2025smartdense} adds a dense-traffic example by combining centralized interaction-mode search with distributed convex trajectory refinement. Language-grounded planners can expose intended maneuvers and critiques, as illustrated by LanguageMPC~\cite{languagempc2023} and CoLMDriver~\cite{liu2025colmdriver}, although the agreement trace still needs controller-level feasibility checks before execution.

\subsubsection{Safety Envelopes for Coordinated Motion}

Agreement over a plan becomes a safety claim only when an independent envelope checks the executed motion. For MAEAD, this envelope must account for uncertain prediction, stale or missing messages, malicious collaborators, actuator limits, and open-set human behavior. Safe MARL for CAVs~\cite{zhang2023spatial} and signal temporal logic-guided MARL~\cite{wang2023multi} show how learned policies can be shaped by explicit safety structure. Communication-aware MARL~\cite{dingRobustMultiAgentCommunication2024, yuRobustCommunicativeMultiAgent2024} adds robustness to unreliable messages. These mechanisms support coordinated action, while runtime monitors, reachability or barrier-function filters, fallback maneuvers, and rejected-plan logs remain necessary for safety assurance.

\subsubsection{Certification Evidence Boundary}

Certification-grade evidence is stronger than benchmark success or closed-loop simulation performance~\cite{koopmanChallengesAutonomousVehicle2016}. It requires traceable inputs, timing budgets, recorded communication failures, rejected alternatives, constraint violations, and fallback triggers. Current game-theoretic, MARL, generative, and language-grounded planners mainly provide technical capability evidence under bounded evaluation protocols. This evidence falls short of deployment certification. Across the reviewed corpus, the missing artifact for MAEAD is an auditable safety case~\cite{ul4600AutonomousProducts2020, iso21448SOTIF2022} that links shared-state records, negotiation records, planner outputs, safety-filter decisions, and execution outcomes across multi-agent scenarios.

In summary, this section shows how planning methods move from interactive intent modeling toward coordinated execution. Game-theoretic and MARL methods formalize interaction and policy coupling. Generative, language-grounded, and world-model planners make proposed actions more explicit and predictive. The unresolved gap is certification: current planners can generate or negotiate coordinated actions, yet safety under delayed communication, wrong intent estimates, and open-set human behavior remains difficult to establish. Section~\ref{sec:e2e} below examines whether end-to-end shared latent policies can close the loop and what interpretability and validation costs they introduce.

\section{End-to-End Multi-Agent Driving}
\label{sec:e2e}

The evolution of autonomous driving systems has trended toward greater integration, culminating in the end-to-end (E2E) paradigm. E2E systems simplify the traditional modular pipeline by mapping sensor inputs, intermediate representations, or fused latent states directly to planning and control outputs. ALVINN~\cite{pomerleau1989alvinn} established the feasibility of learning steering behavior directly from sensor inputs without separately engineering perception and control modules. Modern E2E driving has since expanded to perception-rich and multimodal settings~\cite{chenEndendAutonomousDriving2024}. Extending this idea to multi-agent embodied autonomous driving (MAEAD) remains an open problem because the policy must reason over shared state, interacting intentions, and coordinated downstream action.

\subsection{From Single-Agent E2E to Collaborative Perception-Control}

Single-agent E2E driving provides the technical baseline for this section. TransFuser~\cite{chittaTransFuserImitationTransformerBased2022} uses transformer-based sensor fusion for imitation driving. ST-P3~\cite{huSTP3EndtoEndVisionBased2022} jointly models perception, prediction, and planning in an interpretable vision-based stack. UniAD~\cite{huPlanningorientedAutonomousDriving2023} organizes full-stack driving around planning-oriented task interfaces, and GenAD~\cite{zhengGenADGenerativeEndtoEnd2024} casts E2E driving as generative future modeling. These methods show how perception-to-planning coupling can be learned inside one vehicle. Their evidence remains ego-centric, so MAEAD must extend the latent state from ego planning to cross-agent state alignment, intent and plan alignment, and coordinated downstream action.

The transition from single-vehicle E2E to multi-agent E2E changes the control loop from an ego-centric setting to a collective one. A multi-agent E2E system may process observations from several vehicles or infrastructure nodes and produce coordinated actions for participating agents. We use shared perception-to-control coupling to describe architectures in which cross-agent observations are converted into action-relevant latent state before planning or control. This design can reduce some interface overhead in modular V2X stacks, while creating harder questions about communication, attribution, safety checking, and interpretability.

UniV2X~\cite{yuEndtoEndAutonomousDriving2024} is a representative recent system in this direction. It extends cooperation beyond object detection by connecting perception, online mapping, occupancy prediction, and planning within a unified vehicle-infrastructure architecture. The method uses sparse-dense hybrid V2X transmission and fusion over DAIR-V2X, embedding cross-agent state alignment and downstream planning effects into one model while leaving explicit intent negotiation outside the main design. Coopernaut~\cite{coopernaut2022} is an early E2E cooperative driving system in which V2V latent sharing feeds directly into control outputs. CoDriving~\cite{10979246} is the methodological component of a full-stack collaborative autonomous driving study, using driving-request-aware communication to share sparse perceptual features for E2E driving. V2Xverse~\cite{10979246} is the accompanying comprehensive simulation platform for collaborative autonomous driving. Together, these systems shift E2E cooperation from feature sharing alone toward perception-control coupling, where the shared latent states are expected to affect downstream behavior under communication constraints.

MDrive~\cite{mdrive2026} provides recent closed-loop benchmark evidence grounded in pre-crash typologies and real-world V2X datasets. It reports that perception sharing can fail to improve planning while negotiation can degrade performance in dense traffic. This result is directly relevant to MAEAD because it tests whether perception gains propagate to downstream decisions under interaction conditions.

\subsection{Predictive Latents and Motion Sequence Interfaces}

A related frontier comes from VLA and world-model E2E systems. EMMA~\cite{hwangEMMAEndtoEndMultimodal2024}, Senna~\cite{jiangSennaBridgingLargeVisionLanguage2024}, VLM-AD~\cite{xuVLMADEndtoEndAutonomous2024}, and AutoVLA~\cite{zhouAutoVLAVisionLanguageAction2025} absorb more perception, reasoning, and action generation into language-supervised or multimodal models. V2X-VLM~\cite{youV2XVLMEndtoEndV2X2024} and UniMM-V2X~\cite{songUniMMV2X2026} are direct multi-agent E2E driving systems: V2X-VLM couples vehicle and infrastructure views with textual scene descriptions for trajectory planning, while UniMM-V2X uses shared queries and multi-level fusion to connect perception, prediction, and planning cooperation. DriveWorld-VLA~\cite{driveworldvla2026}, Uni-World VLA~\cite{uniworldvla2026}, VLA-World~\cite{vlaworld2026}, CoWorld-VLA~\cite{coworldvla2026}, and HEAT~\cite{heat2026} show that E2E driving is moving toward closed predictive action loops where imagined scene evolution and action choice are optimized together. Most of these methods remain ego-centric E2E planning or world-model prediction. MAEAD can inherit this direction only when the latent state is extended to cross-agent state alignment and intent and plan alignment.

Motion sequence modeling is a representation mechanism inside this frontier. It encodes future motion, intent-like behavior, and candidate plans as ordered tokens or latent sequences. This framing keeps motion sequence modeling at the representation level within E2E systems. SMART~\cite{wu2024smart} uses next-token prediction over vectorized maps and trajectories, while GenAD~\cite{zhengGenADGenerativeEndtoEnd2024} uses instance-centric tokens and a generative latent space for prediction and planning. For MAEAD, this mechanism makes future behavior modelable, and its cooperative value depends on whether generated motion tokens are consumed by closed-loop planners or controllers.

Table~\ref{tab:e2e_comparison} summarizes the paper lineage used in this section. Single-agent rows provide mechanism lineage, cooperative rows provide direct MAEAD evidence, and frontier rows provide mechanism context for predictive action loops.

\begin{table}[t]
\centering
\caption{Representative End-to-End Driving Methods and Frontier Directions for MAEAD.}
\vspace{-0.5em}
\footnotesize
\label{tab:e2e_comparison}

\begin{tabular}{lll}
\toprule
\textbf{Method} & \textbf{Venue} & \textbf{Primary Mechanism} \\
\midrule
\multicolumn{3}{@{}l}{\emph{Single-Agent E2E Mechanism Lineage}} \\
ALVINN~\cite{pomerleau1989alvinn} & 1989 NeurIPS & Neural sensor-to-steering policy \\
TransFuser~\cite{chittaTransFuserImitationTransformerBased2022} & 2022 TPAMI & Transformer sensor fusion \\
ST-P3~\cite{huSTP3EndtoEndVisionBased2022} & 2022 ECCV & Joint perception-prediction-planning \\
UniAD~\cite{huPlanningorientedAutonomousDriving2023} & 2023 CVPR & Planning-oriented full stack \\
SMART~\cite{wu2024smart} & 2024 NeurIPS & Trajectory tokenization \\
GenAD~\cite{zhengGenADGenerativeEndtoEnd2024} & 2024 ECCV & Generative E2E planning \\
\midrule
\multicolumn{3}{@{}l}{\emph{Cooperative and V2X E2E Evidence}} \\
Coopernaut~\cite{coopernaut2022} & 2022 CVPR & V2V latent sharing \\
V2X-VLM~\cite{youV2XVLMEndtoEndV2X2024} & 2025 TR-C & V2X VLM trajectory planning \\
UniV2X~\cite{yuEndtoEndAutonomousDriving2024} & 2025 AAAI & Sparse-dense V2X E2E \\
CoDriving~\cite{10979246} & 2025 TPAMI & Driving-request V2X E2E \\
UniMM-V2X~\cite{songUniMMV2X2026} & 2026 AAAI & Multi-level V2X E2E fusion \\
MDrive~\cite{mdrive2026} & 2026 arXiv & Closed-loop cooperative benchmark \\
\midrule
\multicolumn{3}{@{}l}{\emph{Frontier VLA and World-Model Directions}} \\
EMMA~\cite{hwangEMMAEndtoEndMultimodal2024} & 2024 TMLR & Multimodal E2E planning \\
Senna~\cite{jiangSennaBridgingLargeVisionLanguage2024} & 2024 arXiv & VLM-to-driving bridge \\
VLM-AD~\cite{xuVLMADEndtoEndAutonomous2024} & 2025 CoRL & VLM-supervised driving \\
AutoVLA~\cite{zhouAutoVLAVisionLanguageAction2025} & 2025 NeurIPS & Adaptive VLA reasoning \\
DriveMoE~\cite{yangDriveMoEMixtureofExperts2025} & 2026 CVPR & VLA mixture-of-experts \\
DriveWorld-VLA~\cite{driveworldvla2026} & 2026 arXiv & Latent world-state planning \\
Uni-World VLA~\cite{uniworldvla2026} & 2026 arXiv & Interleaved prediction and planning \\
VLA-World~\cite{vlaworld2026} & 2026 arXiv & Future-scene trajectory refinement \\
CoWorld-VLA~\cite{coworldvla2026} & 2026 arXiv & Multi-expert world reasoning \\
HEAT~\cite{heat2026} & 2026 arXiv & Trajectory-guided world modeling \\
GameAD~\cite{gamead2026} & 2026 arXiv & Risk-prioritized game planning \\
\bottomrule
\end{tabular}
\vspace{0.3em}

\begin{minipage}{0.65\textwidth}
\footnotesize \textbf{Note:} Single-agent rows provide mechanism lineage, cooperative rows provide direct MAEAD evidence, and frontier rows provide mechanism context for predictive action loops.
\end{minipage}
\vspace{-1em}
\end{table}

\begin{table}[t]
\centering
\caption{Functional Comparison of Traditional Simulators and Cooperative-Driving Simulation Extensions for MAEAD.}
\vspace{-0.5em}
\footnotesize
\label{tab:traditional_simulators}
\resizebox{\textwidth}{!}{%
\begin{tabular}{llll}
\toprule
\textbf{Platform} & \textbf{Source} & \textbf{Simulation Scope} & \textbf{Cooperation Support} \\
\midrule
\multicolumn{4}{@{}l}{\emph{Behavior-Level Traffic and Policy Simulation}} \\
MATSim~\cite{horniMATSim2016} & 2016 Ubiquity Press & Agent-based transport simulation & Large-scale activity-based travel dynamics \\
CityFlow~\cite{zhangCityFlowMultiAgent2019} & 2019 WWW & City-scale microscopic traffic simulation & Multi-agent traffic-signal control environment \\
SUMO~\cite{sumoZenodo2025} & 2025 Zenodo & Microscopic traffic-flow simulation & Large-scale route and signal dynamics \\
HighwayEnv~\cite{leurentHighwayEnv2018} & 2018 GitHub & Tactical driving decision tasks & Configurable multi-agent settings \\
SMARTS~\cite{zhouSMARTSScalableMultiAgent2020} & 2020 CoRL & Interactive traffic simulation & Multi-agent RL scenarios \\
Flow~\cite{wuFlowModularLearning2022} & 2022 T-RO & Mixed-autonomy traffic learning framework & RL control under partial AV adoption \\
MetaDrive~\cite{liMetaDrive2023} & 2023 TPAMI & Compositional RL driving scenarios & Single-agent and multi-agent RL tasks \\
AutoDRIVE-MARL~\cite{AutoDRIVE-MARL-2023} & 2023 arXiv & Autonomous-vehicle learning ecosystem & Cooperative and competitive MARL tasks \\
\midrule
\multicolumn{4}{@{}l}{\emph{Sensor and Physical Simulation}} \\
AirSim~\cite{shahAirSim2017} & 2017 arXiv & Unreal-based visual and physical simulation & Extensible vehicles and sensors \\
CARLA~\cite{dosovitskiyCARLAOpenUrban2017} & 2017 CoRL & 3D urban sensor simulation & Scripted and extensible traffic agents \\
\midrule
\multicolumn{4}{@{}l}{\emph{Scenario Replay and Planning Benchmarks}} \\
CommonRoad~\cite{althoffCommonRoad2017} & 2017 IV & Composable motion-planning benchmarks & Dynamic obstacles and vehicle models \\
Waymax~\cite{gulinoWaymax2023} & 2023 NeurIPS & Data-driven multi-agent simulation & Learned and rule-based sim agents \\
ScenarioNet~\cite{liScenarioNetOpenSourcePlatform2023} & 2023 NeurIPS & Reusable traffic scenario database & Scenario replay from real logs \\
LimSim++~\cite{fuLimSimClosedLoopPlatform2024} & 2024 IV & Closed-loop language-model driving & Multi-scenario interactive agents \\
\midrule
\multicolumn{4}{@{}l}{\emph{Cooperative Driving and V2X Evaluation}} \\
OpenCDA~\cite{xuOpenCDAOpenCooperative2021} & 2021 ITSC & CARLA-based CDA co-simulation & V2X and cooperative-driving pipelines \\
OpenCDA-ROS~\cite{zheng2023opencda} & 2023 TIV & Simulation-to-robot integration & ROS-enabled cooperative stacks \\
V2Xverse~\cite{10979246} & 2025 TPAMI & Full-stack V2X E2E simulation & Closed-loop V2X collaboration \\
M3CAD~\cite{zhuM3CADTowardsGenericCooperative2025} & 2026 ICRA & Cooperative multi-task benchmark & Perception, mapping, forecasting, and planning \\
\bottomrule
\end{tabular}
}
\vspace{0.3em}

\begin{minipage}{0.95\textwidth} 
\footnotesize \textbf{Note:} Group labels indicate the dominant validation role. Several resources also support secondary roles across adjacent groups.
\end{minipage}
\vspace{-1em}
\end{table}

\subsection{Emergent Coordination and Evidence Boundaries}

E2E multi-agent driving is often valued for emergent coordination: coordinated actions that arise from the training objective and interaction data without a manually designed negotiation protocol. Automatic alternating passage at unsignalized intersections is a classic example~\cite{pal2020emergent}. E2E or MARL-trained systems can learn recurring coordination patterns from dense interactive simulation, while those patterns become useful evidence only when they are reproducible, interpretable, and connected to closed-loop safety checks.

The latency argument for E2E systems should be stated cautiously. Collapsing multiple modules into a single forward pass can reduce some interface delay, but total response time still depends on sensing, communication, model size, hardware, and safety monitors. Latency benefits therefore require measurement under a specified deployment stack.

Multi-agent E2E systems face three evidence gaps. Scalability remains limited relative to dense urban traffic, and both communication and attention costs can grow quickly as the number of agents increases. Interpretability is difficult because collapsing perception, prediction, and planning into a single model weakens the module boundaries that engineers use to diagnose failures. Real-world deployment remains unresolved because evidence is dominated by simulation and benchmark studies, leaving the transfer of shared latent representations to physical multi-vehicle systems uncertain.

Within this evidence boundary, generative AI, world models, and motion sequence modeling are best regarded as tools for stress-testing and augmenting E2E coordination. They become relevant when future motion tokens or latent rollouts are linked to prediction, planning, or closed-loop action. MotionDiffuser~\cite{jiang2023motiondiffuser} provides diffusion-based multimodal trajectory generation for multi-agent scenes, and CTG++~\cite{zhong2023ctg} uses guided diffusion to synthesize constraint-conditioned traffic scenarios. Related learned-policy evidence for cooperative lane changes, platoon-like formations, and coordinated merging comes from lane-change MARL~\cite{zhang2022multi}, cooperative MARL-PPO~\cite{kolat2024cooperative}, and MAMQPPO~\cite{mamqppo2025}. These examples should be regarded as research directions, with deployment readiness still open.

In summary, E2E multi-agent driving provides a concrete way to study whether shared latent representations can support collaborative behavior without a fully manually designed negotiation pipeline. It offers a plausible route toward tighter multi-agent coordination, and it raises harder questions about interpretability, controllability, validation, and certification.

Current multi-agent E2E systems provide shared BEV or latent features and downstream planning outputs, but explicit negotiation is usually missing. This suggests that hybrid architectures may be needed to combine shared representation learning, interpretable intent exchange, and verified control under one deployment-facing stack.

\section{Embodied Simulation \& Data Engine}
\label{sec:simulation}

The development and validation of MAEAD systems require scalable simulation and data-generation infrastructure. Real-world testing is indispensable, but cost, safety risk, and rare-event coverage make simulation necessary for iterating collaborative strategies. As methods claim broader shared-state, intent-modeling, and coordinated-control capabilities, simulators must represent perception, interaction, communication, and downstream control with enough fidelity to expose failure modes before deployment.

\subsection{Traditional Simulators}

Traditional platforms remain useful baselines. SUMO~\cite{sumoZenodo2025} supports efficient traffic-flow simulation, while CARLA~\cite{dosovitskiyCARLAOpenUrban2017} provides an Unreal-Engine-based 3D driving simulator with flexible sensor APIs. CARLA has supported many single-agent and early multi-agent studies, but advanced MAEAD stresses three limitations: the perceptual reality gap of synthetic assets, simplified non-ego behavior, and the difficulty of manually generating diverse corner cases.

Dedicated cooperative-driving ecosystems extend this base. SMARTS~\cite{zhouSMARTSScalableMultiAgent2020} targets scalable multi-agent reinforcement learning with interactive traffic scenarios. OpenCDA~\cite{xuOpenCDAOpenCooperative2021} and OpenCDA-ROS~\cite{zheng2023opencda} provide cooperative-driving pipelines and robot-integration support. AutoDRIVE~\cite{AutoDRIVE-MARL-2023} exposes cooperative and competitive MARL tasks at the vehicle-control layer. LimSim++~\cite{fuLimSimClosedLoopPlatform2024} targets closed-loop evaluation for multimodal LLM-based driving. ScenarioNet~\cite{liScenarioNetOpenSourcePlatform2023} standardizes reusable scenarios mined from real-world datasets, and M3CAD~\cite{zhuM3CADTowardsGenericCooperative2025} broadens cooperative evaluation across perception, mapping, forecasting, occupancy, and planning. The separation between simulators, scenario libraries, and cooperative benchmarks is useful when identifying which MAEAD capability a resource can actually test.

Table~\ref{tab:traditional_simulators} compares representative simulator and simulation-extension papers by dominant validation role. The grouping follows what a resource is mainly used to evaluate: behavior-level traffic and policy dynamics, sensor embodiment, scenario replay and planning benchmarks, or cooperative V2X evaluation. Rows can span multiple roles, so the table prioritizes the role most directly used by MAEAD validation.

\subsection{Neural and Generative Simulation for MAEAD}

\subsubsection{Scene Reconstruction and Log-Based Simulation}

Neural rendering shifts simulation from manually authored virtual worlds toward data-driven reconstructions of real scenes. Neural radiance fields (NeRF)~\cite{mildenhall2020nerf} and 3D Gaussian splatting (3DGS)~\cite{kerbl2023gaussian} reconstruct a 3D scene from images by learning either a continuous volumetric function or a set of view-dependent Gaussians. These methods can generate realistic novel views, which is valuable for perception models and VLMs that are sensitive to visual-domain mismatch.

The multi-agent challenge is making reconstructed scenes dynamic. Recent systems decompose scenes into static background and dynamic actors, enabling insertion, removal, or animation of vehicles and pedestrians~\cite{pumarola2021dnerf}. UniSim \cite{yang2023unisim} constructs a neural closed-loop simulator from real-world logs. StreetGaussians extends 3DGS to dynamic urban scenes with editable tracked vehicles~\cite{yan2024streetgaussians}. FRUC~\cite{taoFRUCFeedforward2026} extends feedforward 3D Gaussian splatting to collaborative multi-vehicle reconstruction from uncalibrated views. It models motion-aware occlusion and fuses cross-agent features as constrained residuals to inpaint ego blind spots. At the behavior layer, coordinated-policy particle simulators provide a complementary path by learning interaction dynamics directly for closed-loop social simulation~\cite{peng2021learning}.

\subsubsection{Controllable Scenario and World Generation}

Scene reconstruction improves realism for observed worlds, while generative simulation creates counterfactual and rare cases for stress testing. The generation stack has two separable functions. LLMs and VLMs specify, retrieve, or translate scenario constraints, while diffusion and world models generate sensor streams, geometry, or agent evolution. Recent work indicates a shift from camera-only video generation toward multimodal, geometry-aware, and controllable simulation. DriveGen~\cite{zhang2025drivegeninfinitediversetraffic} synthesizes traffic scenarios under user-defined conditions. DriveDreamer-2~\cite{zhao2024drivedreamer2world} uses video diffusion conditioned on HD maps and 3D layouts to generate multi-view driving videos with controllable traffic composition. GAIA-2~\cite{gaia2_2025} conditions multi-view driving-video generation on ego dynamics, agent configurations, environment factors, and road semantics. Diffusion models are one mechanism inside this controllable-generation family. UniDriveDreamer~\cite{unidrivedreamer2026} directly generates future camera and LiDAR observations in a single-stage framework, and HERMES++~\cite{hermespp2026} connects 3D scene understanding with future geometry prediction. MotionDiffuser~\cite{jiang2023motiondiffuser} generates diverse and socially aware trajectories for multiple agents, while CTG++~\cite{zhong2023ctg} uses guided diffusion conditioned on user-specified constraints to generate safety-critical traffic scenarios. X-World~\cite{xworld2026} adds action-conditioned multi-camera rollouts with optional controls over traffic agents, road elements, and appearance, and Waymo's World Model~\cite{waymoworldmodel2026} describes a multi-sensor world model for rare-event simulation and language-driven scene control. AutoWorld~\cite{autoworld2026} adds a complementary scaling direction: it learns a world model from unlabeled LiDAR occupancy representations and uses it to drive multi-agent traffic simulation, suggesting that future simulators may scale through self-supervised sensor data beyond hand-labeled scenario libraries. ScenarioNet~\cite{liScenarioNetOpenSourcePlatform2023} provides a scenario database that aggregates traffic scenarios from nuScenes, Waymo, and nuPlan into a unified format, giving generative methods reusable seeds and evaluation contexts. This trend is important for MAEAD because shared-world-model validation needs semantic consistency, physically plausible sensor evolution, and controllable multi-agent behavior.

\emph{Language-driven scenario generation.} Language interfaces serve as the authoring layer for the generation stack. They let engineers specify complex events in text, such as a distracted cyclist, an illegal turn, or adverse weather, then translate that description into simulator assets or executable scripts. LCTGen~\cite{tanLCTGenLanguageConditioned2023} maps text queries to road-context selection, actor initialization, and vehicle dynamics for language-conditioned traffic generation. ScenicNL~\cite{elmaaroufiScenicNLGenerating2024} turns crash-report text into probabilistic Scenic programs, making uncertainty explicit through a domain-specific representation. ChatSim~\cite{wei2024editable} edits photorealistic driving scenes through an LLM-guided neural rendering pipeline. ChatScene~\cite{zhangChatSceneKnowledgeEnabledSafetyCritical2024} retrieves scenario-code fragments to translate textual risk descriptions into CARLA safety-critical cases. Text2Scenario~\cite{caiText2ScenarioTextDrivenScenario2026} parses user descriptions into scenario components and links them to executable domain-specific representations. TARGET~\cite{dengTARGETTrafficRuleBased2025} extracts traffic-rule knowledge through an LLM and validates domain-specific representations before synthesizing executable tests. Learning from Risk~\cite{wang2025learningrisk} uses prior-knowledge-guided LLM optimization to search for rare safety-critical interactions. Simple placement and environment edits are more mature than language-specified multi-agent interaction dynamics, so generated cases still require checks for physical plausibility, behavioral realism, and downstream safety relevance.

Together, these approaches shift the MAEAD data engine from passive collection toward on-demand scenario generation, but generated scenarios still require physical, behavioral, and safety validation before being used as evidence of deployment readiness.

Table~\ref{tab:neural_generative_sim_methods} summarizes representative methods for reconstruction, language-guided authoring, and controllable generation.

\begin{table}[t]
\centering
\caption{Representative Neural Reconstruction and Generative Simulation Methods for MAEAD.}
\vspace{-0.5em}
\footnotesize
\label{tab:neural_generative_sim_methods}

\begin{tabular}{lll}
\toprule
\textbf{Method} & \textbf{Venue} & \textbf{Primary Mechanism} \\
\midrule
\multicolumn{3}{@{}l}{\emph{Scene Reconstruction and Log-Based Simulation}} \\
NeRF~\cite{mildenhall2020nerf} & 2020 ECCV & Neural radiance-field rendering \\
D-NeRF~\cite{pumarola2021dnerf} & 2021 CVPR & Dynamic neural scene rendering \\
DriveGAN~\cite{kim2021drivegan} & 2021 CVPR & Controllable neural driving simulation \\
UniSim~\cite{yang2023unisim} & 2023 CVPR & Neural closed-loop sensor simulation \\
3DGS~\cite{kerbl2023gaussian} & 2023 TOG & Real-time Gaussian splatting \\
StreetGaussians~\cite{yan2024streetgaussians} & 2024 ECCV & Editable dynamic urban reconstruction \\
FRUC~\cite{taoFRUCFeedforward2026} & 2026 arXiv & Collaborative feedforward 3DGS \\
\midrule
\multicolumn{3}{@{}l}{\emph{Language-Guided Scenario Authoring}} \\
LCTGen~\cite{tanLCTGenLanguageConditioned2023} & 2023 CoRL & Text-conditioned traffic generation \\
ScenicNL~\cite{elmaaroufiScenicNLGenerating2024} & 2024 COLM & Crash-report-to-Scenic translation \\
ChatSim~\cite{wei2024editable} & 2024 CVPR & LLM-guided photorealistic editing \\
ChatScene~\cite{zhangChatSceneKnowledgeEnabledSafetyCritical2024} & 2024 CVPR & Text-to-CARLA safety cases \\
TARGET~\cite{dengTARGETTrafficRuleBased2025} & 2025 TSE & Validated rule-to-test synthesis \\
Learning from Risk~\cite{wang2025learningrisk} & 2025 arXiv & LLM-guided rare-risk search \\
\midrule
\multicolumn{3}{@{}l}{\emph{Controllable Scenario and World Generation}} \\
MotionDiffuser~\cite{jiang2023motiondiffuser} & 2023 CVPR & Diffusion multi-agent trajectories \\
CTG++~\cite{zhong2023ctg} & 2023 CoRL & Guided diffusion traffic synthesis \\
GAIA-1~\cite{gaia2023} & 2023 arXiv & Generative driving world model \\
Drive-WM~\cite{wangDrivingFutureMultiview2023} & 2024 CVPR & Multiview futures for planning \\
DriveDreamer~\cite{drivedreamer2024} & 2024 ECCV & Real-world-driven video world model \\
DriveDreamer-2~\cite{zhao2024drivedreamer2world} & 2025 AAAI & Layout-conditioned video generation \\
MUVO~\cite{muvo2025} & 2025 IV & Multimodal geometric world model \\
DriveGen~\cite{zhang2025drivegeninfinitediversetraffic} & 2025 IROS & Diverse traffic scenario generation \\
GAIA-2~\cite{gaia2_2025} & 2025 arXiv & Controllable multi-view world model \\
MiLA~\cite{mila2025} & 2025 arXiv & Long-term multi-view video generation \\
Epona~\cite{epona2025} & 2025 ICCV & Autoregressive diffusion world model \\
DrivingGen~\cite{drivinggen2026} & 2026 ICLR & Generative world-model benchmark \\
UniDriveDreamer~\cite{unidrivedreamer2026} & 2026 arXiv & Camera-LiDAR future generation \\
HERMES++~\cite{hermespp2026} & 2026 arXiv & 3D understanding and future geometry \\
X-World~\cite{xworld2026} & 2026 arXiv & Action-conditioned multi-camera rollout \\
WorldDrive~\cite{worlddrive2026} & 2026 arXiv & Unified vision-motion world model \\
AutoWorld~\cite{autoworld2026} & 2026 arXiv & Self-supervised traffic world model \\
ShareVerse~\cite{shareverse2026} & 2026 arXiv & Multi-agent consistent video generation \\
V2XCrafter~\cite{taoV2XCrafterGenerate2026} & 2026 arXiv & Cross-agent driving-scene generation \\
Waymo's World Model~\cite{waymoworldmodel2026} & 2026 Waymo Blog & Controllable multi-sensor simulation \\
\bottomrule
\end{tabular}%
\vspace{-1em}
\end{table}

\subsection{Evaluation Benchmarks and Datasets}
\label{sec:benchmarks}

Beyond simulation platforms, MAEAD also depends on evaluation resources that expose perception, forecasting, planning, and coordination behavior under comparable protocols. This subsection classifies these resources by their dominant evaluation roles: planning and interaction benchmarks, collaborative perception and V2X datasets, and closed-loop cooperative evaluation suites. The goal is to clarify which resources can test shared state, intent and plan alignment, or coordinated downstream action.

\subsubsection{Planning, Interaction, and Driving Logs}

Driving logs and planning benchmarks expose how learned or rule-based agents perceive and act under interactive traffic constraints. nuScenes~\cite{caesarNuScenesMultimodalDataset2020} provides multimodal urban driving logs for perception, tracking, and prediction backbones that later appear inside planning stacks. nuPlan~\cite{nuplan} provides large-scale real-world driving logs and a lightweight closed-loop simulator with reactive agents, making it useful for long-horizon planning evaluation. NAVSIM~\cite{Dauner2024NEURIPS} derives sensor-based planning evaluation from nuPlan and scores driving behavior through collision avoidance, drivable-area compliance, driving direction, comfort, and progress.

Waymo Open Motion Dataset (WOMD)~\cite{ettinger2021large} is better viewed as an interaction and motion-forecasting dataset than as a planning benchmark. It provides large-scale multi-agent trajectories for testing intent prediction, interaction modeling, and counterfactual reasoning modules that can later support planning and control evaluation.

\subsubsection{Collaborative Perception and V2X Datasets}

Collaborative perception datasets provide the main empirical basis for evaluating cross-agent state alignment. OPV2V~\cite{xuOPV2VOpenBenchmark2022} establishes a simulated V2V benchmark with early, late, and intermediate fusion pipelines. DAIR-V2X~\cite{yuDAIRV2XLargeScaleDataset2022} moves this setting to real vehicle-infrastructure sensing and exposes temporal asynchrony and transmission cost. V2X-Sim~\cite{V2Xsim2022} adds a multi-agent collaborative perception benchmark with perception and planning relevance. V2V4Real~\cite{xuV2V4RealRealWorldLargeScale2023} provides synchronized real-world V2V LiDAR, camera, and pose data for cooperative detection, tracking, and domain transfer.

Recent datasets broaden coverage from pairwise cooperation to sequential, roadside, intersection, and accident-oriented settings. V2X-Seq~\cite{v2x-seq} extends DAIR-V2X toward sequential perception and forecasting. V2X-Real~\cite{xiang2024v2x} covers large-scale real-world V2X collaborative perception. RCooper~\cite{hao2024rcooper} targets roadside collaborative perception. HoloVIC~\cite{ma2024holoviclargescaledatasetbenchmark} focuses on holographic intersection and vehicle-infrastructure cooperation. DeepAccident~\cite{Wang_2023_DeepAccident} adds accident-oriented motion and prediction labels for safety-oriented evaluation.

\subsubsection{Closed-Loop Cooperative Evaluation Suites}

Closed-loop cooperative evaluation is needed when the question moves from perception quality to downstream behavior. SMARTS~\cite{zhouSMARTSScalableMultiAgent2020} and AutoDRIVE-MARL~\cite{AutoDRIVE-MARL-2023} supply controllable multi-agent policy-learning tasks. OpenCDA~\cite{xuOpenCDAOpenCooperative2021} and OpenCDA-ROS~\cite{zheng2023opencda} support cooperative-driving stack integration across simulation and robot-oriented execution. V2Xverse~\cite{10979246} provides a full-stack V2X simulation platform for collaborative autonomous driving and supports end-to-end perception-control evaluation through CoDriving. MDrive~\cite{mdrive2026} links scenario generation, Real2Sim conversion, and human-in-the-loop simulation with interactive V2X evaluation. Its failure analyses are useful because they separate perception-sharing gains from downstream planning gains. WHALES~\cite{chen2024whalesmultiagentschedulingdataset} extends the resource landscape toward communication-aware agent scheduling and scalable collaborative perception. ScenarioNet~\cite{liScenarioNetOpenSourcePlatform2023} provides reusable traffic scenarios that can seed planning, simulation, and generative evaluation pipelines.

Table~\ref{tab:datasets} lists representative benchmark and dataset papers based on evaluation roles.

\begin{table}[t]
\centering
\caption{Representative Benchmark and Dataset Papers for MAEAD Evaluation.}
\vspace{-0.5em}
\footnotesize

\label{tab:datasets}
\begin{tabular}{lll}
\toprule
\textbf{Method} & \textbf{Venue} & \textbf{Primary Mechanism} \\
\midrule
\multicolumn{3}{@{}l}{\emph{Planning, Interaction, and Driving Logs}} \\
nuScenes~\cite{caesarNuScenesMultimodalDataset2020} & 2020 CVPR & Multimodal urban logs \\
WOMD~\cite{ettinger2021large} & 2021 ICCV & Interaction forecasting trajectories \\
nuPlan~\cite{nuplan} & 2021 CVPRW & Closed-loop reactive planning \\
NAVSIM~\cite{Dauner2024NEURIPS} & 2024 NeurIPS & Sensor-based planning score \\
\midrule
\multicolumn{3}{@{}l}{\emph{Collaborative Perception and V2X Datasets}} \\
OPV2V~\cite{xuOPV2VOpenBenchmark2022} & 2022 ICRA & Simulated V2V fusion \\
DAIR-V2X~\cite{yuDAIRV2XLargeScaleDataset2022} & 2022 CVPR & Real V2I detection \\
V2X-Sim~\cite{V2Xsim2022} & 2022 RA-L & V2X perception labels \\
V2XSet~\cite{xuV2XViTVehicletoEverythingCooperative2022a} & 2022 ECCV & Noisy V2X perception \\
V2X-Seq~\cite{v2x-seq} & 2023 CVPR & Sequential V2I forecasting \\
V2V4Real~\cite{xuV2V4RealRealWorldLargeScale2023} & 2023 CVPR & Real V2V tracking \\
DeepAccident~\cite{Wang_2023_DeepAccident} & 2024 AAAI & Accident motion prediction \\
RCooper~\cite{hao2024rcooper} & 2024 CVPR & Roadside collaborative perception \\
HoloVIC~\cite{ma2024holoviclargescaledatasetbenchmark} & 2024 CVPR & Intersection V2I evaluation \\
V2X-Real~\cite{xiang2024v2x} & 2024 ECCV & Large-scale real V2X \\
\midrule
\multicolumn{3}{@{}l}{\emph{Closed-Loop and Scenario Evaluation Suites}} \\
SMARTS~\cite{zhouSMARTSScalableMultiAgent2020} & 2020 CoRL & Multi-agent traffic RL \\
OpenCDA~\cite{xuOpenCDAOpenCooperative2021} & 2021 ITSC & Cooperative-driving co-simulation \\
AutoDRIVE-MARL~\cite{AutoDRIVE-MARL-2023} & 2023 arXiv & Cooperative AV control \\
OpenCDA-ROS~\cite{zheng2023opencda} & 2023 TIV & ROS cooperative stacks \\
ScenarioNet~\cite{liScenarioNetOpenSourcePlatform2023} & 2023 NeurIPS & Reusable scenario replay \\
LimSim++~\cite{fuLimSimClosedLoopPlatform2024} & 2024 IV & Closed-loop language agents \\
WHALES~\cite{chen2024whalesmultiagentschedulingdataset} & 2025 IROS & Communication-aware agent scheduling \\
V2Xverse~\cite{10979246} & 2025 TPAMI & Full-stack V2X simulation \\
M3CAD~\cite{zhuM3CADTowardsGenericCooperative2025} & 2026 ICRA & Cooperative multi-task benchmark \\
MDrive~\cite{mdrive2026} & 2026 arXiv & Closed-loop V2X tests \\
\bottomrule
\end{tabular}%
\vspace{-1em}
\end{table}

In summary, the simulation and data engine determines which MAEAD claims can be tested before deployment. Traditional simulators support controlled closed-loop studies, neural reconstruction improves perceptual realism, generative scenario engines broaden rare-event coverage, and multi-agent datasets determine whether perception, planning, and coordination claims can be compared under shared protocols.

\section{Future Directions}
\label{sec:future}
\begingroup

Multi-agent embodied autonomous driving (MAEAD) is moving from decentralized information exchange toward more explicit forms of shared state alignments and coordinated actions. Early work established the value of collaborative perception and V2X-supported decision-making through systems such as V2X-ViT~\cite{xuV2XViTVehicletoEverythingCooperative2022a} and CoBEVT~\cite{xuCoBEVTCooperativeBird2023}. The next stage is broader and harder. A credible MAEAD research program must scale to dense traffic, reason about unknown agents, resist manipulation of shared beliefs, and retain legibility for human road users. Foundation models, including large language models (LLMs), vision-language models (VLMs), and shared world models (SWMs), may support these goals, but their roles in safety-critical deployment remain unsettled. We therefore frame the research agenda on four major issues that are likely to determine whether the SWM-driven organizing view can move beyond simulation.

\subsection{Scalability, Communication and Computing Budgets, and Scaling Behavior}

Scaling MAEAD extends single-agent autonomy into a sophisticated multi-agent regime. In a multi-agent system, more sensors and agents can interact and increase observability, but they also increase communication load, inference cost, coordination ambiguity, and safety risk. A central research question is whether MAEAD has measurable scaling behavior similar to the empirical regularities observed in LLMs and VLMs~\cite{kaplan2020scaling}. The relevant variables include model size, data scale, compute, agent population size, interaction density, communication topology, and the diversity of cooperative and non-cooperative behaviors.

Traditional MARL exposes the difficulty of this regime. Joint action spaces grow rapidly, policies become non-stationary during training, and centralized critics become expensive in dense scenes. Generative world models offer one possible path, with datasets such as V2X-Sim~\cite{V2Xsim2022}, V2V4Real~\cite{xuV2V4RealRealWorldLargeScale2023}, and WHALES~\cite{chen2024whalesmultiagentschedulingdataset},  providing training and evaluation data for interaction modeling and communication-aware agent selection. Current public evidence remains benchmark-bound, and methods such as MAPPO~\cite{yu2021mappo} remain constrained by sample efficiency and transferability. Two recent works sharpen this scaling agenda. AutoWorld~\cite{autoworld2026} explores self-supervised world-model simulation from unlabeled LiDAR occupancy data, while NOMAD~\cite{nomad2026} uses self-play MARL in target-city map simulators to adapt driving policies without human demonstrations from the new city. These works leave deployment reliability unresolved and shift scaling from labeled trajectory accumulation toward unlabeled sensor logs, map priors, and closed-loop self-play. A related trend is graph world modeling, which represents entities as nodes and interactions as edges to improve structured prediction and planning~\cite{graphworldmodels2026}. This direction is especially relevant to MAEAD because shared world models are naturally relational: vehicles, pedestrians, lanes, signals, and communication links all change the meaning of an agent's local observation.

A practical scaling agenda should therefore quantify when additional agents help and when they harm. Future work should report performance as a function of agent count, communication budget, packet loss, scene density, and end-to-end latency, together with a latency breakdown across encoding, transmission, queueing, inference, and control. Evaluations should also disclose computation placement, accelerator and memory budgets, energy use, server load, edge coverage, and handover assumptions. Sparse communication is especially important. Agents should learn \textit{who} to query, \textit{what} to share, and \textit{when} to remain silent, with broad all-to-all feature broadcast avoided. Attention-based routing, learned relevance filters, and event-triggered semantic messages are promising mechanisms, provided that their inference overhead fits within control-loop latency budgets. The long-term target is to identify regime changes that separate basic collision avoidance from strategic cooperation, and to define safety-oriented scaling curves beyond perception accuracy.

Future computing studies should treat the timeliness of shared-state updates as a first-class systems concern because an accurate regional state may become unsafe once it becomes stale. Evaluations should report state age, deadline-miss rate, stale-update rejection, and recovery time after an edge interruption. The ETSI continuous edge computing profile~\cite{etsiContinuousEdgeComputing2025} also motivates state and service migration across roadside edge zones as vehicles move. Such migration must preserve temporal context, uncertainty, and provenance without blocking local control. Model partitioning offers a complementary adaptation mechanism. MEOCI~\cite{liMEOCIModelPartitioning2026} jointly selects vehicle-edge split points and early exits under inference-accuracy constraints. MAEAD extends this problem because a split representation must remain useful for cross-agent fusion and downstream planning. When migration, communication, or remote inference misses its deadline, a minimal onboard perception, planning, and safety path should continue operation.

\subsection{Open-Set Coordination}

Most MAEAD studies assume a closed cooperative set where agents share protocols, objectives, and communication interfaces. Public roads violate this assumption. A deployed system must interact with legacy vehicles, aggressive human drivers, pedestrians, cyclists, construction workers, and non-standard robots. In this open-set regime, shared-state alignment and intent modeling cannot depend on reliable digital messages. Agents must infer the capabilities, intent, and likely response of unknown actors from observation and motion alone.

Generative behavior models are a natural candidate for this problem. They can synthesize plausible futures for observed entities and help planners choose actions that remain robust under multiple hypotheses. LLMs and VLMs may contribute high-level semantic interpretation, for example by grounding visual cues such as tailgating, hesitant lane positioning, or pedestrian attention. However, a pipeline that connects VLM scene understanding, LLM social reasoning, and real-time trajectory generation has not yet been demonstrated end to end in a driving system. The compounded latency, calibration error, and failure modes of such a pipeline remain open concerns, as discussed in Section~\ref{sec:fm-limitations}.

Open-set coordination also requires online adaptation. A useful world model should update its belief about an unfamiliar actor after only a few observations, while preserving conservative safety margins. In-context learning is one possible mechanism for rapid adaptation, but its reliability in safety-critical traffic remains untested. More immediate progress may come from hybrid systems that combine learned prediction with rule-based safety envelopes, uncertainty-aware planning, and fallback maneuvers. The research target is fast enough behavioral categorization to support safe, socially acceptable action under uncertainty, even when actor identity remains imperfect.

\subsection{Trustworthiness \& Security}

As MAEAD moves toward shared-world-model alignment and coordinated action, the attack surface expands from sensors and messages to shared beliefs and collective decisions. A malicious or faulty agent can report false obstacles, hide its intent, poison world-model training data, or shape the shared state so that other vehicles converge to an unsafe plan. This creates consensus-level attacks, where the immediate objective of manipulation becomes the collective belief used for downstream planning.

Byzantine-resilient evidence aggregation and consistency checking~\cite{dengInvestigationByzantineThreats2021} are therefore core requirements that extend beyond communication add-ons. Recent collaborative-perception defenses provide practical anchors: ROBOSAC~\cite{liUsAdversariallyRobust2023} uses sampling-based consensus, CP-Guard~\cite{huCPGuardMaliciousAgent2024} filters suspicious collaborators, GCP~\cite{taoGCPGuardedCollaborative2025} adds spatio-temporal malicious-agent detection, and CP-UniGuard~\cite{huCPUniGuardUnified2026} targets unified malicious-agent defense. These mechanisms must be extended to continuous, real-time driving, where state estimates are uncertain, communication is lossy, and delayed rejection of a malicious message can still cause unsafe operations. Shared generative world models add further risk. Data poisoning or biased fine-tuning can make the model systematically underestimate braking distance, overtrust a specific sensor modality, or generate unsafe joint futures in rare cases.

Recent safety-oriented multimodal foundation-model driving systems point to pragmatic mitigation directions. PKRD-CoT~\cite{luoPKRDCoTUnifiedChainofthought2024} decomposes perception, knowledge grounding, reasoning, and decision-making, while SafeAuto~\cite{zhangSafeAutoKnowledgeEnhancedSafe2025} injects external knowledge and safety constraints into a multimodal foundation-model pipeline. These designs can improve interpretability and commonsense grounding, but current evidence is still benchmark- or simulation-based. Certification-grade evidence for real-time collective control remains unavailable.

Recent risk-attribution work asks whether evidence maps in six-view E2E planners can be converted into planning-risk signals. For MAEAD, this suggests an audit path: expose the planned trajectory together with the views, agents, and received messages that dominate the decision~\cite{planningriskattr2026}.

A key future direction is verifiable shared-world-model alignments. Each collective decision should be accompanied by provenance, uncertainty estimates, and consistency checks that allow peer agents or infrastructure to audit why a coordinated maneuver was selected. This requires logging raw messages, semantic claims, predicted trajectories, rejected alternatives, and safety constraints. Explainable AI may help summarize these traces for human inspection, but the safety mechanism itself should not depend on unconstrained natural-language reasoning. MAEAD systems need machine-checkable validation layers that can reject inconsistent beliefs and unsafe plans before execution.

\subsection{Embodied Social Compliance}

Collision avoidance is a necessary but incomplete measure of driving intelligence. MAEAD must also fit into the social fabric of road use. Human drivers rely on implicit norms such as yielding conventions, lane-change etiquette, assertiveness levels, eye contact, vehicle posture, and context-dependent interpretations of gestures. These norms vary across cities and cultures. A system that is physically safe but socially awkward can still degrade traffic flow, trigger confusion, or provoke risky human responses.

This motivates a socially aware extension of  SWM that estimates physical outcomes and social consequences. Such a model would help answer whether a proposed maneuver is likely to be interpreted as cooperative, aggressive, hesitant, or ambiguous. It must combine kinematic prediction with social cue interpretation. For example, a VLM-equipped agent may need to infer whether a pedestrian is about to cross from head pose and gait, or whether a flashing headlight is a yielding signal or an assertion of priority in the local driving culture.

LLMs and VLMs are potentially useful because they encode broad social and contextual knowledge from pre-training corpora. In simulation, LLM-driven agents such as AgentsCoDriver~\cite{huAgentsCoDriverLargeLanguage2024} show that contextual prompts can modulate planning behavior. The open question is whether prompt-based cultural adaptation transfers to real-world driving behavior, especially when rare events and local norms are underrepresented in training data.

Three research problems are especially important: cultural generalization, implicit cue interpretation, and socially aware motion generation. Cultural generalization asks whether a model can adapt from one driving culture to another with limited data. Implicit cue interpretation asks whether VLMs can decode subtle motion, posture, and timing cues that humans use for negotiation. Socially aware motion generation asks whether planners such as MotionDiffuser~\cite{jiang2023motiondiffuser} can optimize for trajectories that are feasible, safe, predictable, and low friction for surrounding road users. Progress on these problems will determine whether MAEAD becomes merely a high-capacity technical system or a socially compatible participant in mixed traffic.

Overall, the next stage of MAEAD requires scalable shared-state construction under communication limits, robust behavior inference for mixed traffic, auditable belief formation, and socially legible coordination. These directions turn the SWM view from a survey lens into a testable research agenda for deployment-facing systems.
\endgroup

\section{Conclusion}

This survey has examined multi-agent embodied autonomous driving (MAEAD) as a system-level transition from connected sensing to shared predictive world modeling. Across collaborative perception, communication, joint decision-making, end-to-end learning, and simulation infrastructure, the literature shows clear progress in exchanging and fusing evidence across agents. It also shows that deployment-ready cooperation requires more than better perception scores.

\subsection{Key Findings}

We distill four cross-cutting findings:

\begin{enumerate}
    \item \emph{Shared-state construction is the most mature layer.} Intermediate-fusion methods provide concrete mechanisms for cross-agent representation alignment. Explicit plan negotiation and coordinated downstream action remain less mature, and reported evidence for LLM-based coordination systems is still largely simulation-based.
    \item \emph{Foundation models are useful advisory components.} LLMs and VLMs introduce useful capabilities for semantic reasoning and negotiation, yet inference overhead, hallucination risk, and the absence of formal safety guarantees prevent their use as sole decision-makers in safety-critical MAEAD systems.
    \item \emph{The simulation gap is a primary bottleneck.} Neural rendering and generative scenario synthesis can improve perceptual realism, but the behavioral gap remains difficult because human road users react in nuanced and sometimes inconsistent ways.
    \item \emph{Hybrid architectures offer a scalable path.} Combining rule-based safety constraints, MARL adaptation, and foundation-model-based reasoning may provide a multi-layer architecture that can grow with fleet size while preserving interpretability and safety checks.
\end{enumerate}

\subsection{Forward Outlook}

If realized in a reliable and deployable form, MAEAD could improve transportation safety, efficiency, and traffic flow by enabling vehicles to perceive, reason, and act in a coordinated way. The path forward requires progress on certifiable safety for foundation-model-in-the-loop systems, intent-driven communication protocols that move beyond raw data exchange, real-world multi-agent testbeds that complement simulation, and socio-technical frameworks for culturally adaptive driving behavior.

In closing, this survey argues that the next stage of autonomous-driving research will depend on stronger isolated agents, verifiable shared states, explicit intent modeling, and safe coordinated action. Foundation models and embodied multi-agent systems may support this direction, but major questions remain regarding safety, interpretability, and deployment. The research call is clear: the field needs verifiable shared-world-model alignment and safe coordinated action under real-time, lossy, and open-set traffic conditions.

{
  \footnotesize
  \bibliographystyle{unsrtnat}
  \bibliography{ref}

@inproceedings{kim2021drivegan,
  title={{D}rive{GAN}: {T}owards a {C}ontrollable {H}igh-{Q}uality {N}eural {S}imulation},
  author={Kim, Seung Wook and Philion, Jonah and Torralba, Antonio and Fidler, Sanja},
  booktitle={Proceedings of the IEEE/CVF Conference on Computer Vision and Pattern Recognition},
  pages={5820--5829},
  year={2021}
}

@article{zhaoSafetyFieldVehicle2025,
  title={{S}afety {F}ield-{B}ased {V}ehicle-{I}nfrastructure {C}ooperative {P}erception for {A}utonomous {D}riving {U}sing 3{D} {P}oint {C}louds},
  author={Zhao, Cong and Ding, Delong and Lei, Cailin and Wang, Shiyu and Ji, Yuxiong and Du, Yuchuan},
  journal={IEEE Transactions on Intelligent Transportation Systems},
  volume={26},
  number={4},
  pages={4676--4691},
  year={2025},
  doi={10.1109/TITS.2025.3546980}
}

@inproceedings{qiuAVRAugmentedVehicular2018,
  title={{A}{V}{R}: {A}ugmented {V}ehicular {R}eality},
  author={Qiu, Hang and Ahmad, Fawad and Bai, Fan and Gruteser, Marco and Govindan, Ramesh},
  booktitle={Proceedings of the 16th Annual International Conference on Mobile Systems, Applications, and Services (MobiSys)},
  pages={81--95},
  year={2018},
  doi={10.1145/3210240.3210319}
}

@inproceedings{qiuAutoCastScalableInfrastructureless2022,
  title={{A}uto{C}ast: {S}calable {I}nfrastructure-{L}ess {C}ooperative {P}erception for {D}istributed {C}ollaborative {D}riving},
  author={Qiu, Hang and Huang, Pohan and Asavisanu, Namo and Liu, Xiaochen and Psounis, Konstantinos and Govindan, Ramesh},
  booktitle={Proceedings of the 20th Annual International Conference on Mobile Systems, Applications and Services (MobiSys)},
  pages={128--141},
  year={2022},
  doi={10.1145/3498361.3538925}
}

@inproceedings{luRobustCollaborative3D2023,
  title={{R}obust {C}ollaborative 3{D} {O}bject {D}etection in {P}resence of {P}ose {E}rrors},
  author={Lu, Yifan and Li, Quanhao and Liu, Baoan and Dianati, Mehrdad and Feng, Chen and Chen, Siheng and Wang, Yanfeng},
  booktitle={2023 IEEE International Conference on Robotics and Automation (ICRA)},
  year={2023},
  doi={10.1109/ICRA48891.2023.10160546}
}

@inproceedings{yangHow2commCommunicationEfficient2023,
  title={{H}ow2comm: {C}ommunication-{E}fficient and {C}ollaboration-{P}ragmatic {M}ulti-{A}gent {P}erception},
  author={Yang, Dingkang and Yang, Kun and Wang, Yuzheng and Liu, Jing and Xu, Zhi and Yin, Rongbin and Zhai, Peng and Zhang, Lihua},
  booktitle={Advances in Neural Information Processing Systems (NeurIPS)},
  year={2023}
}

@inproceedings{yuFlowBasedFeatureFusion2023,
  title={{F}low-{B}ased {F}eature {F}usion for {V}ehicle-{I}nfrastructure {C}ooperative 3{D} {O}bject {D}etection},
  author={Yu, Haibao and Tang, Yingjuan and Xie, Enze and Mao, Jilei and Luo, Ping and Nie, Zaiqing},
  booktitle={Advances in Neural Information Processing Systems (NeurIPS)},
  year={2023}
}

@inproceedings{yangSpatioTemporalDomainAwareness2023,
  title={{S}patio-{T}emporal {D}omain {A}wareness for {M}ulti-{A}gent {C}ollaborative {P}erception},
  author={Yang, Kun and Yang, Dingkang and Zhang, Jingyu and Li, Mingcheng and Liu, Yang and Liu, Jing and Wang, Hanqi and Sun, Peng and Song, Liang},
  booktitle={2023 IEEE/CVF International Conference on Computer Vision (ICCV)},
  year={2023}
}

@inproceedings{gaoSTAMPScalableTask2025,
  title={{S}{T}{A}{M}{P}: {S}calable {T}ask- and {M}odel-{A}gnostic {C}ollaborative {P}erception},
  author={Gao, Xiangbo and Xu, Runsheng and Li, Jiachen and Wang, Ziran and Fan, Zhiwen and Tu, Zhengzhong},
  booktitle={International Conference on Learning Representations (ICLR)},
  year={2025}
}

@article{ji2024coop,
  title={{T}oward {A}utonomous {V}ehicles: {A} {S}urvey on {C}ooperative {V}ehicle-{I}nfrastructure {S}ystem},
  author={Ji, Yangjie and Zhou, Zewei and Yang, Ziru and Huang, Yanjun and Zhang, Yuanjian and Zhang, Wanting and Xiong, Lu and Yu, Zhuoping},
  journal={iScience},
  volume={27},
  number={5},
  pages={109751},
  year={2024},
  month={may},
  doi={10.1016/j.isci.2024.109751},
}

@article{zhao2024sim,
  title={{A} {S}urvey of {A}utonomous {D}riving {F}rameworks and {S}imulators},
  author={Zhao, Hui and Meng, Min and Li, Xiuxian and Xu, Jia and Li, Li and Galland, Stephane},
  journal={Advanced Engineering Informatics},
  volume={62},
  pages={102850},
  year={2024},
  month={oct},
  doi={10.1016/j.aei.2024.102850},
}

@article{balador2022platoon,
  title={{A} {S}urvey on {V}ehicular {C}ommunication for {C}ooperative {T}ruck {P}latooning {A}pplication},
  author={Balador, Ali and Bazzi, Alessandro and Hernandez-Jayo, Unai and de la Iglesia, Idoia and Ahmadvand, Hossein},
  journal={Vehicular Communications},
  volume={35},
  pages={100460},
  year={2022},
  month={jun},
  doi={10.1016/j.vehcom.2022.100460},
}

@article{hu2024collab,
  title={{C}ollaborative {P}erception for {C}onnected and {A}utonomous {D}riving: {C}hallenges, {P}ossible {S}olutions and {O}pportunities},
  author={Hu, Senkang and Fang, Zhengru and Deng, Yiqin and Chen, Xianhao and Fang, Yuguang},
  journal={IEEE Wireless Communications},
  volume={32},
  number={5},
  pages={228--234},
  year={2025},
  doi={10.1109/MWC.002.2400348}
}

@article{jung2020v2x,
  title={{V}2{X}-{C}ommunication-{A}ided {A}utonomous {D}riving},
  author={Jung, Chanyoung and Lee, Daegyu and Lee, Seungwook and Shim, David Hyunchul},
  journal={Sensors},
  year={2020}
}

@article{ansari2021joint,
  title={{J}oint {U}se of {D}{S}{R}{C} and {C}-{V}2{X} for {V}2{X} {C}ommunications in the 5.9 {G}{H}z {I}{T}{S} {B}and},
  author={Ansari, Keyvan},
  journal={IET Intelligent Transport Systems},
  volume={15},
  number={2},
  pages={213--224},
  year={2021},
  doi={10.1049/itr2.12015}
}

@techreport{ieee80211p2010,
  title={{{I}{E}{E}{E}} {S}tandard for {I}nformation {T}echnology, {T}elecommunications and {I}nformation {E}xchange {B}etween {S}ystems, {L}ocal and {M}etropolitan {A}rea {N}etworks, {S}pecific {R}equirements, {P}art 11: {W}ireless {{L}{A}{N}} {M}edium {A}ccess {C}ontrol ({{M}{A}{C}}) and {P}hysical {L}ayer ({{P}{H}{Y}}) {S}pecifications, {A}mendment 6: {W}ireless {A}ccess in {V}ehicular {E}nvironments},
  author={{IEEE}},
  institution={IEEE Standards Association},
  type={Standard},
  number={IEEE Std 802.11p-2010},
  year={2010},
  doi={10.1109/IEEESTD.2010.5514475},
}

@techreport{ieee1609wave2019,
  title={{{I}{E}{E}{E}} {G}uide for {W}ireless {A}ccess in {V}ehicular {E}nvironments ({{W}{A}{V}{E}}) {A}rchitecture},
  author={{IEEE}},
  institution={IEEE Standards Association},
  type={Standard},
  number={IEEE Std 1609.0-2019},
  year={2019},
}

@techreport{etsiEN3026372CAM2019,
  title={{I}ntelligent {T}ransport {S}ystems ({I}{T}{S}); {V}ehicular {C}ommunications; {B}asic {S}et of {A}pplications; {P}art 2: {S}pecification of {C}ooperative {A}wareness {B}asic {S}ervice},
  author={{ETSI}},
  institution={European Telecommunications Standards Institute},
  type={European Standard},
  number={ETSI EN 302 637-2 V1.4.1},
  year={2019},
  month={apr},
}

@techreport{etsiEN3026373DENM2019,
  title={{I}ntelligent {T}ransport {S}ystems ({I}{T}{S}); {V}ehicular {C}ommunications; {B}asic {S}et of {A}pplications; {P}art 3: {S}pecifications of {D}ecentralized {E}nvironmental {N}otification {B}asic {S}ervice},
  author={{ETSI}},
  institution={European Telecommunications Standards Institute},
  type={European Standard},
  number={ETSI EN 302 637-3 V1.3.1},
  year={2019},
  month={apr},
}

@techreport{threegppTS23285V2X2019,
  title={{A}rchitecture {E}nhancements for {V}2{X} {S}ervices},
  author={{3GPP}},
  institution={3rd Generation Partnership Project},
  type={Technical Specification},
  number={3GPP TS 23.285 Release 15},
  year={2019},
}

@techreport{threegppTS23287V2X2022,
  title={{A}rchitecture {E}nhancements for 5{G} {S}ystem (5{G}{S}) to {S}upport {V}ehicle-to-{E}verything ({V}2{X}) {S}ervices},
  author={{3GPP}},
  institution={3rd Generation Partnership Project},
  type={Technical Specification},
  number={3GPP TS 23.287 Release 16},
  year={2022},
  month={jul},
}

@techreport{saeJ2735V2X2020,
  title={{V}2{X} {C}ommunications {M}essage {S}et {D}ictionary},
  author={{SAE International}},
  institution={SAE International},
  type={Technical Standard},
  number={SAE J2735\_202007},
  year={2020},
  month={jul},
  doi={10.4271/J2735_202007},
}

@techreport{etsiTS103324CollectivePerception2023,
  title={{I}ntelligent {T}ransport {S}ystems ({I}{T}{S}); {V}ehicular {C}ommunications; {B}asic {S}et of {A}pplications; {C}ollective {P}erception {S}ervice; {R}elease 2},
  author={{ETSI}},
  institution={European Telecommunications Standards Institute},
  type={Technical Specification},
  number={ETSI TS 103 324 V2.1.1},
  year={2023},
  month={jun},
}

@article{oliveiraManeuverCoordinationAnalysis2024,
  title={{A} {M}aneuver {C}oordination {A}nalysis {U}sing {A}rtery {V}2{X} {S}imulation {F}ramework},
  author={Oliveira, Jo{\~a}o and Vieira, Emanuel and Almeida, Jo{\~a}o and Ferreira, Joaquim and Bartolomeu, Paulo C.},
  journal={Electronics},
  volume={13},
  number={23},
  pages={4813},
  year={2024},
  doi={10.3390/electronics13234813}
}

@article{sheng2024semantic,
  title={{S}emantic {C}ommunication for {C}ooperative {P}erception {B}ased on {I}mportance {M}ap},
  author={Sheng, Yucheng and Hao, Yongxing and Ye, Hao and Liang, Le and Jin, Shi and Li, Geoffrey Ye},
  journal={ScienceDirect},
  year={2024}
}

@article{ye2025survey,
  title={{A} {S}urvey on {S}emantic {C}ommunications in {I}nternet of {V}ehicles},
  author={Ye, Sha and Wu, Qiong and Fan, Pingyi and Fan, Qiang},
  journal={Entropy},
  year={2025}
}

@article{lyu2024semantic,
  title={{S}emantic {V}ehicle-to-{E}verything ({V}2{X}) {C}ommunications {T}owards 6{G}},
  author={Lyu, Tengfei and Noor-A-Rahim, Md. and O'Driscoll, Aisling and Pesch, Dirk},
  journal={arXiv preprint},
  year={2024},
  eprint={2407.17186},
  archivePrefix={arXiv},
  doi={10.48550/arXiv.2407.17186},
}

@article{gimenez2024semantic,
  title={{S}emantic {V}2{X} {C}ommunications for {I}mage {T}ransmission},
  author={Gimenez-Guzman, Jose Manuel and Leyva-Mayorga, Israel and Popovski, Petar},
  journal={IEEE Transactions on Vehicular Technology},
  year={2024}
}

@article{cui2024talking,
  title={{T}alking {V}ehicles: {C}ooperative {D}riving via {N}atural {L}anguage},
  author={Cui, Jiaxun and Tang, Chen and Holtz, Jarrett and Nguyen, Janice and Allievi, Alessandro G. and Qiu, Hang and Stone, Peter},
  journal={OpenReview},
  year={2024}
}

@inproceedings{foersterLearningCommunicateDeep2016,
  title={{L}earning to {C}ommunicate with {D}eep {M}ulti-{A}gent {R}einforcement {L}earning},
  author={Foerster, Jakob and Assael, Ioannis Alexandros and de Freitas, Nando and Whiteson, Shimon},
  booktitle={Advances in Neural Information Processing Systems},
  volume={29},
  year={2016},
}

@inproceedings{sukhbaatarLearningMultiagentCommunication2016,
  title={{L}earning {M}ultiagent {C}ommunication with {B}ackpropagation},
  author={Sukhbaatar, Sainbayar and Szlam, Arthur and Fergus, Rob},
  booktitle={Advances in Neural Information Processing Systems},
  volume={29},
  year={2016},
}

@inproceedings{rasouliAreTheyGoing2017,
  title={{A}re {T}hey {G}oing to {C}ross? {A} {B}enchmark {D}ataset and {B}aseline for {P}edestrian {C}rosswalk {B}ehavior},
  author={Rasouli, Amir and Kotseruba, Iuliia and Tsotsos, John K.},
  booktitle={Proceedings of the IEEE International Conference on Computer Vision Workshops},
  pages={206--213},
  year={2017}
}

@inproceedings{rasouliPIELargeScale2019,
  title={{{P}{I}{E}}: {A} {L}arge-{S}cale {D}ataset and {M}odels for {P}edestrian {I}ntention {E}stimation and {T}rajectory {P}rediction},
  author={Rasouli, Amir and Kotseruba, Iuliia and Kunic, Toni and Tsotsos, John K.},
  booktitle={Proceedings of the IEEE/CVF International Conference on Computer Vision},
  pages={6262--6271},
  year={2019}
}

@inproceedings{casasIntentNetLearning2018,
  title={{I}ntent{N}et: {L}earning to {P}redict {I}ntention from {R}aw {S}ensor {D}ata},
  author={Casas, Sergio and Luo, Wenjie and Urtasun, Raquel},
  booktitle={Proceedings of The 2nd Conference on Robot Learning},
  pages={947--956},
  year={2018},
  volume={87},
  series={Proceedings of Machine Learning Research},
  publisher={PMLR},
}

@inproceedings{fisac2019hierarchical,
  title={{H}ierarchical {G}ame-{T}heoretic {P}lanning for {A}utonomous {V}ehicles},
  author={Fisac, Jaime F. and Bronstein, Elias and Stefansson, Emmanuel and Sadigh, Dorsa and Sastry, S. Shankar and Dragan, Anca D.},
  booktitle={2019 International Conference on Robotics and Automation (ICRA)},
  pages={9590--9596},
  year={2019},
  doi={10.1109/ICRA.2019.8794007}
}

@article{huang2024game,
  title={{N}ash or {S}tackelberg? {A} {C}omparative {S}tudy for {G}ame-{T}heoretic {A}utonomous {V}ehicle {D}ecision-{M}aking},
  author={Bateman, Brady and Xin, Ming and Tseng, H. Eric and Liu, Mushuang},
  journal={IFAC-PapersOnLine},
  volume={58},
  number={28},
  pages={504--509},
  year={2024},
  doi={10.1016/j.ifacol.2025.01.096}
}

@incollection{yuan2024game,
  title={{G}ame-{T}heoretic {D}ecision-{M}aking for {A}utonomous {D}riving {V}ehicles},
  author={Yuan, Mingfeng and Shan, Jinjun},
  booktitle={Autonomous Vehicles and Systems},
  pages={269--300},
  year={2024},
  publisher={River Publishers}
}

@article{chen2025mixed,
  title={{M}ixed {M}otivation {D}riven {S}ocial {M}ulti-{A}gent {R}einforcement {L}earning for {A}utonomous {D}riving},
  author={Chen, Long and Deng, Peng and Li, Lingxi and Hu, Xuemin},
  journal={IEEE/CAA Journal of Automatica Sinica},
  volume={12},
  number={6},
  pages={1272--1282},
  year={2025},
  doi={10.1109/JAS.2025.125201},
}

@article{kolat2024cooperative,
  title={{C}ooperative {M}{A}{R}{L}-{P}{P}{O} {A}pproach for {A}utomated {H}ighway {P}latooning},
  author={Ko{\l}at, Robert and Podrou{\v{z}}kov{\'a}, Petra and Dobe{\v{s}}, Petr and Ji{\v{r}}{\'i} P{\v{r}}ibil},
  journal={MDPI},
  year={2024}
}

@article{hua2024communication,
  title={{C}ommunication-{E}fficient {M}{A}{R}{L} for {P}latoon {S}tability and {E}nergy-{E}fficiency {C}o-{O}ptimization in {C}ooperative {A}daptive {C}ruise {C}ontrol of {C}{A}{V}s},
  author={Hua, Min and Chen, Dong and Jiang, Kun and Zhang, Fanggang and Wang, Jinhai and Wang, Bo and Zhou, Quan and Xu, Hongming},
  journal={IEEE Transactions on Vehicular Technology},
  volume={74},
  number={4},
  pages={6076--6087},
  year={2025},
  doi={10.1109/TVT.2024.3511091},
}

@inproceedings{pal2020emergent,
  title={{E}mergent {R}oad {R}ules {I}n {M}ulti-{A}gent {D}riving {E}nvironments},
  author={Pal, Avik and Philion, Jonah and Liao, Yuan-Hong and Fidler, Sanja},
  booktitle={International Conference on Learning Representations (ICLR)},
  year={2021},
  doi={10.48550/arXiv.2011.10753},
}

@article{bonnefon2016social,
  title={{T}he {S}ocial {D}ilemma of {A}utonomous {V}ehicles},
  author={Bonnefon, Jean-Fran{\c{c}}ois and Shariff, Azim and Rahwan, Iyad},
  journal={Science},
  volume={352},
  number={6293},
  pages={1573--1576},
  year={2016}
}

@inproceedings{bhattacharyya2019simulating,
  title={{S}imulating {E}mergent {P}roperties of {H}uman {D}riving {B}ehavior {U}sing {M}ulti-{A}gent {R}eward {A}ugmented {I}mitation {L}earning},
  author={Bhattacharyya, Raunak P and Phillips, Derek J and Liu, Changliu and Gupta, Jayesh K and Driggs-Campbell, Katherine and Kochenderfer, Mykel J},
  booktitle={IEEE International Conference on Robotics and Automation (ICRA)},
  year={2019}
}

@article{marlsmart2025,
  title={{M}ulti-{A}gent {R}einforcement {L}earning-{B}ased {C}ooperative {A}utonomous {D}riving in {S}mart {I}ntersections},
  author={Yu, Taoyuan and Wang, Kui and Li, Zongdian and Yu, Tao and Sakaguchi, Kei},
  journal={arXiv preprint},
  year={2025},
  eprint={2505.04231},
  archivePrefix={arXiv},
  doi={10.48550/arXiv.2505.04231},
}

@inproceedings{smpe2025,
  title={{E}nhancing {C}ooperative {M}ulti-{A}gent {R}einforcement {L}earning with {S}tate {M}odelling and {A}dversarial {E}xploration},
  author={Kontogiannis, Andreas and Papathanasiou, Konstantinos and Shen, Yi and Stamou, Giorgos and Zavlanos, Michael M. and Vouros, George},
  booktitle={Proceedings of the 42nd International Conference on Machine Learning},
  series={Proceedings of Machine Learning Research},
  volume={267},
  pages={31437--31466},
  year={2025},
  publisher={PMLR},
}

@article{mamqppo2025,
  title={{M}ulti-{A}gent {D}eep {R}einforcement {L}earning {C}ooperative {C}ontrol {M}odel for {A}utonomous {V}ehicle {M}erging into {P}latoon in {H}ighway},
  author={Chen, Jiajia and Zhu, Bingqing and Zhang, Mengyu and Ling, Xiang and Ruan, Xiaobo and Deng, Yifan and Guo, Ning},
  journal={World Electric Vehicle Journal},
  volume={16},
  number={4},
  pages={225},
  year={2025},
  doi={10.3390/wevj16040225},
}

@inproceedings{liu2025colmdriver,
  title={{C}o{L}{M}{D}river: {L}{L}{M}-{B}ased {N}egotiation {B}enefits {C}ooperative {A}utonomous {D}riving},
  author={Liu, Changxing and Liu, Genjia and Wang, Zijun and Yang, Jinchang and Chen, Siheng},
  booktitle={Proceedings of the IEEE/CVF International Conference on Computer Vision (ICCV)},
  pages={25951--25960},
  month={October},
  year={2025},
}

@article{confidencev2x2025,
  title={{C}onfidence-{V}2{X}: {C}onfidence-{D}riven {S}parse {C}ommunication for {E}fficient {V}2{X} {C}ooperative {P}erception},
  author={Tan, Xiaojun and Wang, Rui and Wang, Jinping and Wang, Shuai and Wang, Xu and Wu, Dongsheng},
  journal={Advanced Engineering Informatics},
  volume={69},
  pages={103914},
  year={2026},
  doi={10.1016/j.aei.2025.103914},
}

@article{gaia2_2025,
  title={{G}{A}{I}{A}-2: {A} {C}ontrollable {M}ulti-{V}iew {G}enerative {W}orld {M}odel for {A}utonomous {D}riving},
  author={Russell, Lloyd and Hu, Anthony and Bertoni, Lorenzo and Fedoseev, George and Shotton, Jamie and Arani, Elahe and Corrado, Gianluca},
  journal={arXiv preprint},
  year={2025},
  eprint={2503.20523},
  archivePrefix={arXiv},
  doi={10.48550/arXiv.2503.20523},
}

@article{v2xunipool2025,
  title={{V}2{X}-{U}ni{P}ool: {U}nifying {M}ultimodal {P}erception and {K}nowledge {R}easoning for {A}utonomous {D}riving},
  author={Luo, Xuewen and Yang, Fengze and Ding, Fan and Gao, Xiangbo and Xing, Shuo and Zhou, Yang and Tu, Zhengzhong and Liu, Chenxi},
  journal={arXiv preprint},
  year={2025},
  eprint={2506.02580},
  archivePrefix={arXiv},
  doi={10.48550/arXiv.2506.02580},
}

@article{hyperv2x2026,
  title={{H}yper-{V}2{X}: {H}ypernetworks for {E}stimating {E}pistemic and {A}leatoric {U}ncertainty in {C}ooperative {B}{ird'S}-{E}ye-{V}iew {S}emantic {S}egmentation},
  author={Jagtap, Abhishek Dinkar and Sadashivaiah, Sanath Tiptur and Festag, Andreas},
  journal={arXiv preprint},
  year={2026},
  eprint={2605.21309},
  archivePrefix={arXiv},
  doi={10.48550/arXiv.2605.21309},
}

@article{unidrivedreamer2026,
  title={{U}ni{D}rive{D}reamer: {A} {S}ingle-{S}tage {M}ultimodal {W}orld {M}odel for {A}utonomous {D}riving},
  author={Zhao, Guosheng and Wang, Yaozeng and Wang, Xiaofeng and Zhu, Zheng and Yu, Tingdong and Huang, Guan and Zai, Yongchen and Jiao, Ji and Xue, Changliang and Wang, Xiaole and Yang, Zhen and Zhu, Futang and Wang, Xingang},
  journal={arXiv preprint},
  year={2026},
  eprint={2602.02002},
  archivePrefix={arXiv},
  doi={10.48550/arXiv.2602.02002},
}

@article{driveworldvla2026,
  title={{D}rive{W}orld-{V}{L}{A}: {U}nified {L}atent-{S}pace {W}orld {M}odeling with {V}ision-{L}anguage-{A}ction for {A}utonomous {D}riving},
  author={Jia, Feiyang and Liu, Lin and Song, Ziying and Jia, Caiyan and Ye, Hangjun and Hao, Xiaoshuai and Chen, Long},
  journal={arXiv preprint},
  year={2026},
  eprint={2602.06521},
  archivePrefix={arXiv},
  doi={10.48550/arXiv.2602.06521},
}

@article{uniworldvla2026,
  title={{U}ni-{W}orld {V}{L}{A}: {I}nterleaved {W}orld {M}odeling and {P}lanning for {A}utonomous {D}riving},
  author={Liu, Qiqi and Xu, Huan and Li, Jingyu and Sun, Bin and Hao, Zhihui and She, Dangen and Zhu, Xiatian and Zhang, Li},
  journal={arXiv preprint},
  year={2026},
  eprint={2603.27287},
  archivePrefix={arXiv},
  doi={10.48550/arXiv.2603.27287},
}

@article{hermespp2026,
  title={{H}{E}{R}{M}{E}{S}++: {T}oward a {U}nified {D}riving {W}orld {M}odel for 3{D} {S}cene {U}nderstanding and {G}eneration},
  author={Zhou, Xin and Liang, Dingkang and Chen, Xiwu and Tan, Feiyang and Zhang, Dingyuan and Zhao, Hengshuang and Bai, Xiang},
  journal={arXiv preprint},
  year={2026},
  eprint={2604.28196},
  archivePrefix={arXiv},
  doi={10.48550/arXiv.2604.28196},
}

@article{graphworldmodels2026,
  title={{G}raph {W}orld {M}odels: {C}oncepts, {T}axonomy, and {F}uture {D}irections},
  author={Liu, Jiawei and Yang, Senqiao and Wang, Mingjun and Wang, Yu and Yu, Bei},
  journal={arXiv preprint},
  year={2026},
  eprint={2604.27895},
  archivePrefix={arXiv},
  doi={10.48550/arXiv.2604.27895},
}

@article{afformer2026,
  title={{A}{F}{F}ormer: {A}daptive {F}eature {F}usion {T}ransformer for {V}2{X} {C}ooperative {P}erception {U}nder {C}hannel {I}mpairments},
  author={Zhou, Xi and Huang, Tao and Han, Qing-Long and Abbas, Rana and Rahimi Azghadi, Mostafa},
  journal={arXiv preprint},
  year={2026},
  eprint={2605.01888},
  archivePrefix={arXiv},
  doi={10.48550/arXiv.2605.01888},
}

@inproceedings{coopreflect2026,
  title={{C}oop{R}eflect: {T}owards {N}atural {L}anguage {C}ommunication for {C}ooperative {A}utonomous {D}riving via {M}ulti-{A}gent {L}earning},
  author={Cui, Jiaxun and Tang, Chen and Holtz, Jarrett and Nguyen, Janice and Allievi, Alessandro G. and Qiu, Hang and Stone, Peter},
  booktitle={Proceedings of the 25th International Conference on Autonomous Agents and Multiagent Systems (AAMAS)},
  pages={744--752},
  year={2026},
  month={May},
  address={Paphos, Cyprus},
  doi={10.65109/MOAV6406},
}

@article{scorp2026,
  title={{S}{C}{O}{R}{P}: {S}cene-{C}onsistent {M}ulti-{A}gent {D}iffusion {P}lanning with {S}table {O}nline {R}einforcement {P}ost-{T}raining for {C}ooperative {D}riving},
  author={Bai, Haojie and Li, Aimin and Yao, Ruoyu and Zhao, Xiongwei and Zhang, Tingting and Zhang, Xing and Gao, Lin and Ma, Jun},
  journal={arXiv preprint},
  year={2026},
  eprint={2604.11734},
  archivePrefix={arXiv},
  doi={10.48550/arXiv.2604.11734},
}

@article{mdrive2026,
  title={{M}{D}rive: {B}enchmarking {C}losed-{L}oop {C}ooperative {D}riving for {E}nd-to-{E}nd {M}ulti-{A}gent {S}ystems},
  author={Coscoy, Marco and Zhou, Zewei and Zhao, Seth Z. and Wei, Henry and Magtoto, Angela and Liu, Johnson and Song, Rui and Zimmer, Walter and Huang, Zhiyu and Tang, Chen and Zhou, Bolei and Ma, Jiaqi},
  journal={arXiv preprint},
  year={2026},
  eprint={2605.10904},
  archivePrefix={arXiv},
  doi={10.48550/arXiv.2605.10904},
}

@article{vlaworld2026,
  title={{L}earning {V}ision-{L}anguage-{A}ction {W}orld {M}odels for {A}utonomous {D}riving},
  author={Wang, Guoqing and Tang, Pin and Ren, Xiangxuan and Zhao, Guodongfang and Feng, Bailan and Ma, Chao},
  journal={arXiv preprint},
  year={2026},
  eprint={2604.09059},
  archivePrefix={arXiv},
  doi={10.48550/arXiv.2604.09059},
}

@article{coworldvla2026,
  title={{C}o{W}orld-{V}{L}{A}: {T}hinking in a {M}ulti-{E}xpert {W}orld {M}odel for {A}utonomous {D}riving},
  author={Huang, Minqing and Xiang, Yujiao and Liang, Zihan and Huang, Jiajie and Wang, Jingqi and Xu, Zhi and Tan, Feiyang and Zhou, Hangning and Yang, Mu and Che, Gong},
  journal={arXiv preprint},
  year={2026},
  eprint={2605.10426},
  archivePrefix={arXiv},
  doi={10.48550/arXiv.2605.10426},
}

@article{xworld2026,
  title={{X}-{W}orld: {C}ontrollable {E}go-{C}entric {M}ulti-{C}amera {W}orld {M}odels for {S}calable {E}nd-to-{E}nd {D}riving},
  author={Zheng, Chaoda and Li, Sean and Deng, Jinhao and Wang, Zhennan and Chen, Shijia and Xiao, Liqiang and Chi, Ziheng and Lin, Hongbin and Chen, Kangjie and Wang, Boyang and Zhang, Yu and Liu, Xianming},
  journal={arXiv preprint},
  year={2026},
  eprint={2603.19979},
  archivePrefix={arXiv},
  doi={10.48550/arXiv.2603.19979},
}

@article{waymoworldmodel2026,
  title={{T}he {W}aymo {W}orld {M}odel: {A} {N}ew {F}rontier for {A}utonomous {D}riving {S}imulation},
  author={Jiang, Chiyu Max and Masotto, Xander and Sun, Bo},
  howpublished={Waymo Blog},
  year={2026},
  month={February},
}

@article{heat2026,
  title={{H}{E}{A}{T}: {H}eterogeneous {E}nd-to-{E}nd {A}utonomous {D}riving via {T}rajectory-{G}uided {W}orld {M}odels},
  author={Cho, Hoonhee and Lee, Giwon and Kang, Jae-Young and Yang, Hyemin and Park, Heejun and Yoon, Kuk-Jin},
  journal={arXiv preprint},
  year={2026},
  eprint={2605.19631},
  archivePrefix={arXiv},
  doi={10.48550/arXiv.2605.19631},
}

@article{lcdrive2025,
  title={{L}atent {C}hain-of-{T}hought {W}orld {M}odeling for {E}nd-to-{E}nd {D}riving},
  author={Tan, Shuhan and Chitta, Kashyap and Chen, Yuxiao and Tian, Ran and You, Yurong and Wang, Yan and Luo, Wenjie and Cao, Yulong and Krahenbuhl, Philipp and Pavone, Marco and Ivanovic, Boris},
  journal={arXiv preprint},
  year={2025},
  eprint={2512.10226},
  archivePrefix={arXiv},
  doi={10.48550/arXiv.2512.10226},
}

@article{driveagent2025,
  title={{D}rive{A}gent: {M}ulti-{A}gent {S}tructured {R}easoning with {L}{L}{M} and {M}ultimodal {S}ensor {F}usion for {A}utonomous {D}riving},
  author={Hou, Xinmeng and Wang, Wuqi and Yang, Long and Lin, Hao and Feng, Jinglun and Min, Haigen and Zhao, Xiangmo},
  journal={IEEE Robotics and Automation Letters},
  year={2025},
}

@article{languagempc2023,
  title={{L}anguage{M}{P}{C}: {L}arge {L}anguage {M}odels as {D}ecision {M}akers for {A}utonomous {D}riving},
  author={Sha, Hao and Mu, Yao and Jiang, Yuxuan and Chen, Li and Xu, Chenfeng and Luo, Ping and Li, Shengbo Eben and Tomizuka, Masayoshi and Zhan, Wei and Ding, Mingyu},
  journal={arXiv preprint},
  year={2023},
  eprint={2310.03026},
  archivePrefix={arXiv},
  doi={10.48550/arXiv.2310.03026},
}

@article{gaia2023,
  title={{G}{A}{I}{A}-1: {A} {G}enerative {W}orld {M}odel for {A}utonomous {D}riving},
  author={Hu, Anthony and Russell, Lloyd and Yeo, Hudson and Murez, Zak and Fedoseev, George and Kendall, Alex and Shotton, Jamie and Corrado, Gianluca},
  journal={arXiv preprint},
  year={2023},
  eprint={2309.17080},
  archivePrefix={arXiv},
  doi={10.48550/arXiv.2309.17080},
}

@inproceedings{drivedreamer2024,
  title={{D}rive{D}reamer: {T}owards {R}eal-{W}orld-{D}riven {W}orld {M}odels for {A}utonomous {D}riving},
  author={Wang, Xiaofeng and Zhu, Zheng and Huang, Guan and Chen, Xinze and Zhu, Jiagang and Lu, Jiwen},
  booktitle={European Conference on Computer Vision (ECCV)},
  year={2024},
  doi={10.48550/arXiv.2309.09777},
}

@inproceedings{muvo2025,
  title={{M}{U}{V}{O}: {A} {M}ultimodal {G}enerative {W}orld {M}odel for {A}utonomous {D}riving with {G}eometric {R}epresentations},
  author={Bogdoll, Daniel and Yang, Yitian and Joseph, Tim and Yazgan, Melih and Z{\"o}llner, J. Marius},
  booktitle={IEEE Intelligent Vehicles Symposium (IV)},
  year={2025},
  doi={10.1109/IV64158.2025.11097718},
}

@inproceedings{combo2025,
  title={{C}{O}{M}{B}{O}: {C}ompositional {W}orld {M}odels for {E}mbodied {M}ulti-{A}gent {C}ooperation},
  author={Zhang, Hongxin and Wang, Zeyuan and Lyu, Qiushi and Zhang, Zheyuan and Chen, Sunli and Shu, Tianmin and Dariush, Behzad and Lee, Kwonjoon and Du, Yilun and Gan, Chuang},
  booktitle={International Conference on Learning Representations (ICLR)},
  year={2025},
  doi={10.48550/arXiv.2404.10775},
}

@article{khan2023advancingcv2x,
  title={{A}dvancing {C}-{V}2{X} for {L}evel 5 {A}utonomous {D}riving from the {P}erspective of 3{G}{P}{P} {S}tandards},
  author={Khan, Muhammad Jalal and Khan, Manzoor Ahmed and Malik, Sumbal and Kulkarni, Parag and Alkaabi, Najla and Ullah, Obaid and El-Sayed, Hesham and Ahmed, Amir and Turaev, Sherzod},
  journal={Sensors},
  volume={23},
  number={4},
  pages={2261},
  year={2023},
  doi={10.3390/s23042261},
}

@article{gnnv2x2024,
  title={{G}raph {N}eural {N}etworks and {D}eep {R}einforcement {L}earning {B}ased {R}esource {A}llocation for {V}2{X} {C}ommunications},
  author={Ji, Maoxin and Wu, Qiong and Fan, Pingyi and Cheng, Nan and Chen, Wen and Wang, Jiangzhou and Letaief, Khaled B.},
  journal={IEEE Internet of Things Journal},
  year={2024},
}

@inproceedings{eqmotion2023,
  title={{E}q{M}otion: {E}quivariant {M}ulti-{A}gent {M}otion {P}rediction with {I}nvariant {I}nteraction {R}easoning},
  author={Xu, Chenxin and Tan, Robby T. and Tan, Yuhong and Chen, Siheng and Wang, Yu Guang and Wang, Xinchao and Wang, Yanfeng},
  booktitle={Proceedings of the IEEE/CVF Conference on Computer Vision and Pattern Recognition (CVPR)},
  year={2023}
}

@article{huang2025cogdrive,
  title={{C}og{D}rive: {C}ognition-{D}riven {M}ultimodal {P}rediction-{P}lanning {F}usion for {S}afe {A}utonomy},
  author={Huang, Heye and Yang, Yibin and Fan, Mingfeng and Wang, Haoran and Zhao, Xiaocong and Wang, Jianqiang},
  journal={arXiv preprint},
  year={2025},
  doi={10.48550/arXiv.2512.02777}
}

@article{huang2025smartdense,
  title={{S}{M}{A}{R}{T}: {S}calable {M}ulti-{A}gent {R}easoning and {T}rajectory {P}lanning in {D}ense {E}nvironments},
  author={Huang, Heye and Yang, Yibin and Chen, Wang and Chen, Tiantian and Li, Xiaopeng and Chen, Sikai},
  journal={arXiv preprint},
  year={2025},
  doi={10.48550/arXiv.2509.15737}
}

@article{wang2025learningrisk,
  title={{L}earning from {R}isk: {L}{L}{M}-{G}uided {G}eneration of {S}afety-{C}ritical {S}cenarios with {P}rior {K}nowledge},
  author={Wang, Yuhang and Huang, Heye and Xu, Zhenhua and Sun, Kailai and Guo, Baoshen and Zhao, Jinhua},
  journal={arXiv preprint},
  year={2025},
  doi={10.48550/arXiv.2511.20726}
}

@article{huang2025lead,
  title={{L}{E}{A}{D}: {L}earning-{E}nhanced {A}daptive {D}ecision-{M}aking for {A}utonomous {D}riving in {D}ynamic {E}nvironments},
  author={Huang, Heye and Liu, Jinxin and Zhang, Bin and Zhao, Shaokang and Li, Baojia and Wang, Jianqiang},
  journal={IEEE Transactions on Intelligent Transportation Systems},
  volume={26},
  number={5},
  pages={6142--6156},
  year={2025}
}

@inproceedings{abdelfattahAdversarialAttacksCameraLiDAR2021,
  title = {{A}dversarial {{{A}ttacks}} on {{{C}amera-{L}i{D}{A}{R} {M}odels}} for {{3{D} {C}ar {D}etection}}},
  booktitle = {2021 {{IEEE}}/{{RSJ International Conference}} on {{Intelligent Robots}} and {{Systems}} ({{IROS}})},
  author = {Abdelfattah, Mazen and Yuan, Kaiwen and Wang, Z. Jane and Ward, Rabab},
  year = {2021},
  month = sep,
  pages = {2189--2194},
  publisher = {IEEE},
  address = {Prague, Czech Republic},
  doi = {10.1109/IROS51168.2021.9636638},
  urldate = {2023-11-26},
  isbn = {978-1-6654-1714-3},
  langid = {english}
}

@article{abdelnabiLLMDeliberationEvaluatingLLMs2023,
  title = {{{{L}{L}{M}-{D}eliberation}}: {{{E}valuating {L}{L}{M}s}} with {{{I}nteractive {M}ulti-{A}gent {N}egotiation {G}ames}}},
  shorttitle = {{{LLM-Deliberation}}},
  author = {Abdelnabi, Sahar and Gomaa, Amr and Sivaprasad, Sarath and Sch{\"o}nherr, Lea and Fritz, Mario},
  year = {2023},
  month = sep,
  eprint = {2309.17234},
  primaryclass = {cs},
  publisher = {arXiv},
  urldate = {2024-04-11},
  archiveprefix = {arXiv},
  journal = {arXiv preprint},
}

@article{albertiIDDALargeScaleMultiDomain2020,
  title = {{{{I}{D}{D}{A}}}: {{{A} {L}arge-{S}cale {M}ulti-{D}omain {D}ataset}} for {{{A}utonomous {D}riving}}},
  shorttitle = {{{IDDA}}},
  author = {Alberti, Emanuele and Tavera, Antonio and Masone, Carlo and Caputo, Barbara},
  year = {2020},
  month = oct,
  journal = {IEEE Robotics and Automation Letters},
  volume = {5},
  number = {4},
  pages = {5526--5533},
  issn = {2377-3766},
  doi = {10.1109/LRA.2020.3009075}
}

@inproceedings{arrudaCrossDomainCarDetection2019,
  title = {{C}ross-{{{D}omain {C}ar {D}etection {U}sing {U}nsupervised {I}mage-to-{I}mage {T}ranslation}}: {{{F}rom {D}ay}} to {{{N}ight}}},
  shorttitle = {Cross-{{Domain Car Detection Using Unsupervised Image-to-Image Translation}}},
  booktitle = {2019 {{International Joint Conference}} on {{Neural Networks}} ({{IJCNN}})},
  author = {Arruda, Vinicius F. and Paix{\~a}o, Thiago M. and Berriel, Rodrigo F. and De Souza, Alberto F. and Badue, Claudine and Sebe, Nicu and {Oliveira-Santos}, Thiago},
  year = {2019},
  month = jul,
  pages = {1--8},
  doi = {10.1109/IJCNN.2019.8852008},
  urldate = {2023-10-13}
}

@article{azfarDeepLearningBasedComputer2024,
  title = {{D}eep {{{L}earning-{B}ased {C}omputer {V}ision {M}ethods}} for {{{C}omplex {T}raffic {E}nvironments {P}erception}}: {{{A}~{R}eview}}},
  author = {Azfar, Talha and Li, Jinlong and Yu, Hongkai and Cheu, Ruey L. and Lv, Yisheng and Ke, Ruimin},
  year = {2024},
  month = jan,
  journal = {Data Science for Transportation},
  volume = {6},
  number = {1},
  pages = {1},
  issn = {2948-1368},
  doi = {10.1007/s42421-023-00086-7}
}

@inproceedings{baiInfrastructureBasedObjectDetection2022,
  title = {{I}nfrastructure-{{{B}ased {O}bject {D}etection}} and {{{T}racking}} for {{{C}ooperative {D}riving {A}utomation}}: {{{A} {S}urvey}}},
  shorttitle = {Infrastructure-{{Based Object Detection}} and {{Tracking}} for {{Cooperative Driving Automation}}},
  booktitle = {2022 {{IEEE Intelligent Vehicles Symposium}} ({{IV}})},
  author = {Bai, Zhengwei and Wu, Guoyuan and Qi, Xuewei and Liu, Yongkang and Oguchi, Kentaro and Barth, Matthew J.},
  year = {2022},
  month = jun,
  pages = {1366--1373},
  doi = {10.1109/IV51971.2022.9827461},
  urldate = {2023-11-13}
}

@article{baiQwenVLVersatileVisionLanguage2023,
  title = {{Q}wen-{{{V}{L}}}: {{{A} {V}ersatile {V}ision-{L}anguage {M}odel}} for {{{U}nderstanding}}, {{{L}ocalization}}, {{{T}ext {R}eading}}, and {{{B}eyond}}},
  shorttitle = {Qwen-{{VL}}},
  author = {Bai, Jinze and Bai, Shuai and Yang, Shusheng and Wang, Shijie and Tan, Sinan and Wang, Peng and Lin, Junyang and Zhou, Chang and Zhou, Jingren},
  year = {2023},
  month = oct,
  eprint = {2308.12966},
  primaryclass = {cs},
  publisher = {arXiv},
  urldate = {2024-05-18},
  archiveprefix = {arXiv},
  langid = {english},
  journal = {arXiv preprint},
}

@inproceedings{baiTransFusionRobustLiDARCamera2022,
  title={{{{T}rans{F}usion}}: {{{R}obust {L}i{D}{A}{R}-{C}amera {F}usion}} for {{3{D} {O}bject {D}etection}} with {{{T}ransformers}}},
  author={Bai, Xuyang and Hu, Zeyu and Zhu, Xinge and Huang, Qingqiu and Chen, Yilun and Fu, Hongbo and Tai, Chiew-Lan},
  booktitle={Proceedings of the IEEE/CVF Conference on Computer Vision and Pattern Recognition (CVPR)},
  year={2022},
}

@article{balkusSurveyCollaborativeMachine2022,
  title = {{A} {{{S}urvey}} of {{{C}ollaborative {M}achine {L}earning {U}sing 5{G} {V}ehicular {C}ommunications}}},
  author = {Balkus, Salvador V. and Wang, Honggang and Cornet, Brian D. and Mahabal, Chinmay and Ngo, Hieu and Fang, Hua},
  year = {2022},
  journal = {IEEE Communications Surveys \& Tutorials},
  volume = {24},
  number = {2},
  pages = {1280--1303},
  issn = {1553-877X},
  doi = {10.1109/COMST.2022.3149714},
  urldate = {2023-11-13}
}

@article{bendiabAutonomousVehiclesSecurity2023,
  title = {{A}utonomous {{{V}ehicles {S}ecurity}}: {{{C}hallenges}} and {{{S}olutions {U}sing {B}lockchain}} and {{{A}rtificial {I}ntelligence}}},
  shorttitle = {Autonomous {{Vehicles Security}}},
  author = {Bendiab, Gueltoum and Hameurlaine, Amina and Germanos, Georgios and Kolokotronis, Nicholas and Shiaeles, Stavros},
  year = {2023},
  month = apr,
  journal = {IEEE Transactions on Intelligent Transportation Systems},
  volume = {24},
  number = {4},
  pages = {3614--3637},
  issn = {1558-0016},
  doi = {10.1109/TITS.2023.3236274}
}

@incollection{bhallaDeepMultiAgent2020,
  title = {{D}eep {{{M}ulti {A}gent {R}einforcement {L}earning}} for {{{A}utonomous {D}riving}}},
  booktitle = {Advances in {{Artificial Intelligence}}},
  author = {Bhalla, Sushrut and Ganapathi Subramanian, Sriram and Crowley, Mark},
  editor = {Goutte, Cyril and Zhu, Xiaodan},
  year = {2020},
  volume = {12109},
  pages = {67--78},
  publisher = {Springer International Publishing},
  address = {Cham},
  doi = {10.1007/978-3-030-47358-7_7},
  urldate = {2023-12-23},
  isbn = {978-3-030-47357-0 978-3-030-47358-7},
  langid = {english}
}

@incollection{boddupalliREDEMRealTimeDetection2020,
  title = {{{{R}{E}{D}{E}{M}}}: {{{R}eal-{T}ime {D}etection}} and {{{M}itigation}} of {{{C}ommunication {A}ttacks}} in {{{C}onnected {A}utonomous {V}ehicle {A}pplications}}},
  shorttitle = {{{REDEM}}},
  booktitle = {Internet of {{Things}}. {{A Confluence}} of {{Many Disciplines}}},
  author = {Boddupalli, Srivalli and Ray, Sandip},
  editor = {Casaca, Augusto and Katkoori, Srinivas and Ray, Sandip and Strous, Leon},
  year = {2020},
  volume = {574},
  pages = {105--122},
  publisher = {Springer International Publishing},
  address = {Cham},
  doi = {10.1007/978-3-030-43605-6_7},
  urldate = {2024-08-07},
  isbn = {978-3-030-43604-9 978-3-030-43605-6},
  langid = {english}
}

@inproceedings{boddupalliReplaceRealtimeSecurity2021,
  title = {{R}eplace: {{{R}eal-time {S}ecurity {A}ssurance}} in {{{V}ehicular {P}latoons {A}gainst {V}2{V} {A}ttacks}}},
  shorttitle = {Replace},
  booktitle = {2021 {{IEEE International Intelligent Transportation Systems Conference}} ({{ITSC}})},
  author = {Boddupalli, Srivalli and Hegde, Ashwini and Ray, Sandip},
  year = {2021},
  month = sep,
  pages = {1179--1185},
  doi = {10.1109/ITSC48978.2021.9564757},
  urldate = {2024-08-07}
}

@inproceedings{bogdollAnomalyDetectionAutonomous2022,
  title = {{A}nomaly {{{D}etection}} in {{{A}utonomous {D}riving}}: {{{A} {S}urvey}}},
  shorttitle = {Anomaly {{Detection}} in {{Autonomous Driving}}},
  booktitle = {2022 {{IEEE}}/{{CVF Conference}} on {{Computer Vision}} and {{Pattern Recognition Workshops}} ({{CVPRW}})},
  author = {Bogdoll, Daniel and Nitsche, Maximilian and Zollner, J. Marius},
  year = {2022},
  month = jun,
  pages = {4487--4498},
  publisher = {IEEE},
  address = {New Orleans, LA, USA},
  doi = {10.1109/CVPRW56347.2022.00495},
  urldate = {2024-09-11},
  copyright = {https://doi.org/10.15223/policy-029},
  isbn = {978-1-6654-8739-9},
  langid = {english}
}

@inproceedings{bogdollPerceptionDatasetsAnomaly2023,
  title = {{P}erception {{{D}atasets}} for {{{A}nomaly {D}etection}} in {{{A}utonomous {D}riving}}: {{{A} {S}urvey}}},
  shorttitle = {Perception {{Datasets}} for {{Anomaly Detection}} in {{Autonomous Driving}}},
  booktitle = {2023 {{IEEE Intelligent Vehicles Symposium}} ({{IV}})},
  author = {Bogdoll, Daniel and Uhlemeyer, Svenja and Kowol, Kamil and Z{\"o}llner, J. Marius},
  year = {2023},
  month = jun,
  pages = {1--8},
  issn = {2642-7214},
  doi = {10.1109/IV55152.2023.10186609},
  urldate = {2024-09-11}
}

@article{britoLearningInteractionawareGuidance2021,
  title={{L}earning {{{I}nteraction-aware {G}uidance {P}olicies}} for {{{M}otion {P}lanning}} in {{{D}ense {T}raffic {S}cenarios}}},
  author={Brito, Bruno and Agarwal, Achin and {Alonso-Mora}, Javier},
  journal={IEEE Transactions on Intelligent Transportation Systems},
  year={2022},
}

@inproceedings{brohanRT2VisionLanguageActionModels2023,
  title={{{{R}{T}-2}}: {{{V}ision-{L}anguage-{A}ction {M}odels {T}ransfer {W}eb {K}nowledge}} to {{{R}obotic {C}ontrol}}},
  author={Brohan, Anthony and Brown, Noah and Carbajal, Justice and Chebotar, Yevgen and Chen, Xi and Choromanski, Krzysztof and Ding, Tianli and Driess, Danny and Dubey, Avinava and Finn, Chelsea and Florence, Pete and Fu, Chuyuan and Arenas, Montse Gonzalez and Gopalakrishnan, Keerthana and Han, Kehang and Hausman, Karol and Herzog, Alexander and Hsu, Jasmine and Ichter, Brian and Irpan, Alex and Joshi, Nikhil and Julian, Ryan and Kalashnikov, Dmitry and Kuang, Yuheng and Leal, Isabel and Lee, Lisa and Lee, Tsang-Wei Edward and Levine, Sergey and Lu, Yao and Michalewski, Henryk and Mordatch, Igor and Pertsch, Karl and Rao, Kanishka and Reymann, Krista and Ryoo, Michael and Salazar, Grecia and Sanketi, Pannag and Sermanet, Pierre and Singh, Jaspiar and Singh, Anikait and Soricut, Radu and Tran, Huong and Vanhoucke, Vincent and Vuong, Quan and Wahid, Ayzaan and Welker, Stefan and Wohlhart, Paul and Wu, Jialin and Xia, Fei and Xiao, Ted and Xu, Peng and Xu, Sichun and Yu, Tianhe and Zitkovich, Brianna},
  booktitle={Proceedings of the Conference on Robot Learning (CoRL)},
  year={2023},
}

@inproceedings{nuplan,
  title = {{{nuPlan}: A Closed-Loop ML-Based Planning Benchmark for Autonomous Vehicles}},
  author = {Caesar, Holger and Kabzan, Juraj and Tan, Kok Seang and Fong, Whye Kit and Wolff, Eric and Lang, Alex and Fletcher, Luke and Beijbom, Oscar and Omari, Sammy},
  booktitle = {CVPR 2021 ADP3 Workshop},
  year = {2021}
}

@inproceedings{caesarNuScenesMultimodalDataset2020,
  title = {{{nu{S}cenes}}: {{{A} {M}ultimodal {D}ataset}} for {{{A}utonomous {D}riving}}},
  shorttitle = {{{nuScenes}}},
  booktitle = {2020 {{IEEE}}/{{CVF Conference}} on {{Computer Vision}} and {{Pattern Recognition}} ({{CVPR}})},
  author = {Caesar, Holger and Bankiti, Varun and Lang, Alex H. and Vora, Sourabh and Liong, Venice Erin and Xu, Qiang and Krishnan, Anush and Pan, Yu and Baldan, Giancarlo and Beijbom, Oscar},
  year = {2020},
  month = jun,
  pages = {11618--11628},
  publisher = {IEEE},
  address = {Seattle, WA, USA},
  doi = {10.1109/CVPR42600.2020.01164},
  urldate = {2023-10-24},
  isbn = {978-1-7281-7168-5},
  langid = {english}
}

@article{caillotSurveyCooperativePerception2022,
  title = {{S}urvey on {{{C}ooperative {P}erception}} in an {{{A}utomotive {C}ontext}}},
  author = {Caillot, Antoine and Ouerghi, Safa and Vasseur, Pascal and Boutteau, Remi and Dupuis, Yohan},
  year = {2022},
  month = sep,
  journal = {IEEE Transactions on Intelligent Transportation Systems},
  volume = {23},
  number = {9},
  pages = {14204--14223},
  issn = {1524-9050, 1558-0016},
  doi = {10.1109/TITS.2022.3153815},
  urldate = {2023-11-13},
  langid = {english}
}

@inproceedings{caoAdversarialSensorAttack2019,
  title = {{A}dversarial {{{S}ensor {A}ttack}} on {{{L}i{D}{A}{R}-based {P}erception}} in {{{A}utonomous {D}riving}}},
  booktitle = {Proceedings of the 2019 {{ACM SIGSAC Conference}} on {{Computer}} and {{Communications Security}}},
  author = {Cao, Yulong and Xiao, Chaowei and Cyr, Benjamin and Zhou, Yimeng and Park, Won and Rampazzi, Sara and Chen, Qi Alfred and Fu, Kevin and Mao, Z. Morley},
  year = {2019},
  month = nov,
  pages = {2267--2281},
  publisher = {ACM},
  address = {London United Kingdom},
  doi = {10.1145/3319535.3339815},
  urldate = {2023-11-28},
  isbn = {978-1-4503-6747-9},
  langid = {english}
}

@article{caoCCTRCalibratingTrajectory2024,
  title = {{{{C}{C}{T}{R}}}: {{{C}alibrating {T}rajectory {P}rediction}} for {{{U}ncertainty-{A}ware {M}otion {P}lanning}} in {{{A}utonomous {D}riving}}},
  shorttitle = {{{CCTR}}},
  author = {Cao, Chengtai and Chen, Xinhong and Wang, Jianping and Song, Qun and Tan, Rui and Li, Yung-Hui},
  year = {2024},
  month = mar,
  journal = {Proceedings of the AAAI Conference on Artificial Intelligence},
  volume = {38},
  number = {19},
  pages = {20949--20957},
  issn = {2374-3468, 2159-5399},
  doi = {10.1609/aaai.v38i19.30085},
  urldate = {2024-06-25},
  langid = {english}
}

@article{caoEmergingThreatsDeep2022,
  title = {{E}merging {{{T}hreats}} in {{{D}eep {L}earning-{B}ased {A}utonomous {D}riving}}: {{{A} {C}omprehensive {S}urvey}}},
  shorttitle = {Emerging {{Threats}} in {{Deep Learning-Based Autonomous Driving}}},
  author = {Cao, Hui and Zou, Wenlong and Wang, Yinkun and Song, Ting and Liu, Mengjun},
  year = {2022},
  month = oct,
  eprint = {2210.11237},
  primaryclass = {cs},
  publisher = {arXiv},
  doi = {10.48550/arXiv.2210.11237},
  urldate = {2023-09-28},
  archiveprefix = {arXiv},
  journal = {arXiv preprint},
}

@inproceedings{caoInvisibleBothCamera2021,
  title = {{I}nvisible for {B}oth {{{C}amera}} and {{{L}i{D}{A}{R}}}: {{{S}ecurity}} of {{{M}ulti-{S}ensor {F}usion}} {B}ased {{{P}erception}} in {{{A}utonomous {D}riving {U}nder {P}hysical-{W}orld {A}ttacks}}},
  shorttitle = {Invisible for Both {{Camera}} and {{LiDAR}}},
  booktitle = {2021 {{IEEE Symposium}} on {{Security}} and {{Privacy}} ({{SP}})},
  author = {Cao, Yulong and Wang, Ningfei and Xiao, Chaowei and Yang, Dawei and Fang, Jin and Yang, Ruigang and Chen, Qi Alfred and Liu, Mingyan and Li, Bo},
  year = {2021},
  month = may,
  pages = {176--194},
  issn = {2375-1207},
  doi = {10.1109/SP40001.2021.00076}
}

@inproceedings{chenAgentVerseFacilitatingMultiAgent2023,
  title={{{{A}gent{V}erse}}: {{{F}acilitating {M}ulti-{A}gent {C}ollaboration}} and {{{E}xploring {E}mergent {B}ehaviors}}},
  author={Chen, Weize and Su, Yusheng and Zuo, Jingwei and Yang, Cheng and Yuan, Chenfei and Chan, Chi-Min and Yu, Heyang and Lu, Yaxi and Hung, Yi-Hsin and Qian, Chen and Qin, Yujia and Cong, Xin and Xie, Ruobing and Liu, Zhiyuan and Sun, Maosong and Zhou, Jie},
  booktitle={International Conference on Learning Representations (ICLR)},
  year={2024},
}

@inproceedings{chenAsynchronousLargeLanguage2024,
  title={{A}synchronous {{{L}arge {L}anguage {M}odel {E}nhanced {P}lanner}} for {{{A}utonomous {D}riving}}},
  author={Chen, Yuan and Ding, Zi-han and Wang, Ziqin and Wang, Yan and Zhang, Lijun and Liu, Si},
  booktitle={Proceedings of the European Conference on Computer Vision (ECCV)},
  year={2024},
}

@inproceedings{chenAutomaticOvertakingTwoway2021,
  title = {{A}utomatic {{{O}vertaking}} on {{{T}wo-way {R}oads}} with {{{V}ehicle {I}nteractions {B}ased}} on {{{P}roximal {P}olicy {O}ptimization}}},
  booktitle = {2021 {{IEEE Intelligent Vehicles Symposium}} ({{IV}})},
  author = {Chen, Xiaochang and Wei, Jieqiang and Ren, Xiaoqiang and Johansson, Karl H. and Wang, Xiaofan},
  year = {2021},
  month = jul,
  pages = {1057--1064},
  publisher = {IEEE},
  address = {Nagoya, Japan},
  doi = {10.1109/IV48863.2021.9575954},
  urldate = {2023-12-21},
  isbn = {978-1-7281-5394-0},
  langid = {english}
}

@article{chenConnectedAutomatedVehicle2021,
  title = {{C}onnected and {{{A}utomated {V}ehicle {D}istributed {C}ontrol}} for {{{O}n-ramp {M}erging {S}cenario}}: {{{A} {V}irtual {R}otation {A}pproach}}},
  shorttitle = {Connected and {{Automated Vehicle Distributed Control}} for {{On-ramp Merging Scenario}}},
  author = {Chen, Tianyi and Wang, Meng and Gong, Siyuan and Zhou, Yang and Ran, Bin},
  year = {2021},
  month = dec,
  journal = {Transportation Research Part C: Emerging Technologies},
  volume = {133},
  pages = {103451},
  issn = {0968090X},
  doi = {10.1016/j.trc.2021.103451},
  urldate = {2024-03-18}
}

@article{chenCooperativeRampMerging2024,
  title = {{C}ooperative {{{R}amp {M}erging {S}trategy}} at {{{M}ulti-{L}ane {A}rea}} for {{{A}utomated {V}ehicles}}},
  author = {Chen, Ruishuang and Yang, Zaiyue},
  year = {2024},
  journal = {IEEE Transactions on Vehicular Technology},
  pages = {1--15},
  issn = {0018-9545, 1939-9359},
  doi = {10.1109/TVT.2024.3401232},
  urldate = {2024-05-22},
  copyright = {https://ieeexplore.ieee.org/Xplorehelp/downloads/license-information/IEEE.html},
  langid = {english}
}

@inproceedings{chenCooperCooperativePerception2019,
  title = {{C}ooper: {{{C}ooperative {P}erception}} for {{{C}onnected {A}utonomous {V}ehicles {B}ased}} on {{3{D} {P}oint {C}louds}}},
  shorttitle = {Cooper},
  booktitle = {2019 {{IEEE}} 39th {{International Conference}} on {{Distributed Computing Systems}} ({{ICDCS}})},
  author = {Chen, Qi and Tang, Sihai and Yang, Qing and Fu, Song},
  year = {2019},
  month = jul,
  pages = {514--524},
  publisher = {IEEE},
  address = {Dallas, TX, USA},
  doi = {10.1109/ICDCS.2019.00058},
  urldate = {2024-05-30},
  copyright = {https://ieeexplore.ieee.org/Xplorehelp/downloads/license-information/IEEE.html},
  isbn = {978-1-7281-2519-0},
  langid = {english}
}

@article{chenDeepMultiagentReinforcement2022,
  title={{D}eep {{{M}ulti-agent {R}einforcement {L}earning}} for {{{H}ighway {O}n-{R}amp {M}erging}} in {{{M}ixed {T}raffic}}},
  author={Chen, Dong and Hajidavalloo, Mohammad and Li, Zhaojian and Chen, Kaian and Wang, Yongqiang and Jiang, Longsheng and Wang, Yue},
  journal={IEEE Transactions on Intelligent Transportation Systems},
  year={2023},
}

@article{chenDeepNeuralNetwork2021,
  title = {{D}eep {{{N}eural {N}etwork {B}ased {V}ehicle}} and {{{P}edestrian {D}etection}} for {{{A}utonomous {D}riving}}: {{{A} {S}urvey}}},
  shorttitle = {Deep {{Neural Network Based Vehicle}} and {{Pedestrian Detection}} for {{Autonomous Driving}}},
  author = {Chen, Long and Lin, Shaobo and Lu, Xiankai and Cao, Dongpu and Wu, Hangbin and Guo, Chi and Liu, Chun and Wang, Fei-Yue},
  year = {2021},
  month = jun,
  journal = {IEEE Transactions on Intelligent Transportation Systems},
  volume = {22},
  number = {6},
  pages = {3234--3246},
  issn = {1524-9050, 1558-0016},
  doi = {10.1109/TITS.2020.2993926},
  urldate = {2024-03-21},
  langid = {english}
}

@inproceedings{chenDrivingLLMsFusing2023,
  title={{D}riving with {{{L}{L}{M}s}}: {{{F}using {O}bject-{L}evel {V}ector {M}odality}} for {{{E}xplainable {A}utonomous {D}riving}}},
  author={Chen, Long and Sinavski, Oleg and H{\"u}nermann, Jan and Karnsund, Alice and Willmott, Andrew James and Birch, Danny and Maund, Daniel and Shotton, Jamie},
  booktitle={Proceedings of the IEEE International Conference on Robotics and Automation (ICRA)},
  year={2024},
}

@article{chenEndendAutonomousDriving2024,
  title = {{E}nd-to-{E}nd {{{A}utonomous {D}riving}}: {{{C}hallenges}} and {{{F}rontiers}}},
  shorttitle = {End-to-End {{Autonomous Driving}}},
  author = {Chen, Li and Wu, Penghao and Chitta, Kashyap and Jaeger, Bernhard and Geiger, Andreas and Li, Hongyang},
  year = {2024},
  journal = {IEEE Transactions on Pattern Analysis and Machine Intelligence},
  pages = {1--20},
  issn = {1939-3539},
  doi = {10.1109/TPAMI.2024.3435937},
  urldate = {2024-10-28}
}

@inproceedings{chenFcooperFeatureBased2019,
  title = {{F}-{C}ooper: {F}eature {B}ased {C}ooperative {P}erception for {A}utonomous {V}ehicle {E}dge {C}omputing {S}ystem {U}sing {{3{D}}} {P}oint {C}louds},
  shorttitle = {F-Cooper},
  booktitle = {Proceedings of the 4th {{ACM}}/{{IEEE Symposium}} on {{Edge Computing}}},
  author = {Chen, Qi and Ma, Xu and Tang, Sihai and Guo, Jingda and Yang, Qing and Fu, Song},
  year = {2019},
  month = nov,
  pages = {88--100},
  publisher = {ACM},
  address = {Arlington Virginia},
  doi = {10.1145/3318216.3363300},
  urldate = {2024-05-30},
  isbn = {978-1-4503-6733-2},
  langid = {english}
}

@inproceedings{chenLearningAllVehicles2022a,
  title={{L}earning from {{{A}ll {V}ehicles}}},
  author={Chen, Dian and Kr{\"a}henb{\"u}hl, Philipp},
  booktitle={Proceedings of the IEEE/CVF Conference on Computer Vision and Pattern Recognition (CVPR)},
  year={2022},
}

@article{chenMilestonesAutonomousDriving2023,
  title = {{M}ilestones in {{{A}utonomous {D}riving}} and {{{I}ntelligent {V}ehicles}}: {{{S}urvey}} of {{{S}urveys}}},
  shorttitle = {Milestones in {{Autonomous Driving}} and {{Intelligent Vehicles}}},
  author = {Chen, Long and Li, Yuchen and Huang, Chao and Li, Bai and Xing, Yang and Tian, Daxin and Li, Li and Hu, Zhongxu and Na, Xiaoxiang and Li, Zixuan and Teng, Siyu and Lv, Chen and Wang, Jinjun and Cao, Dongpu and Zheng, Nanning and Wang, Fei-Yue},
  year = {2023},
  month = feb,
  journal = {IEEE Transactions on Intelligent Vehicles},
  volume = {8},
  number = {2},
  pages = {1046--1056},
  issn = {2379-8904},
  doi = {10.1109/TIV.2022.3223131},
  urldate = {2023-12-07}
}

@inproceedings{chibLGTrajLLMGuided2024,
  title={{{{L}{G}-{T}raj}}: {{{L}{L}{M} {G}uided {P}edestrian {T}rajectory {P}rediction}}},
  author={Chib, Pranav Singh and Singh, Pravendra},
  booktitle={Proceedings of the IEEE/CVF International Conference on Computer Vision Workshops (ICCVW)},
  year={2025},
}

@article{chittaTransFuserImitationTransformerBased2022,
  title = {{{{T}rans{F}user}}: {{{I}mitation}} with {{{T}ransformer-{B}ased {S}ensor {F}usion}} for {{{A}utonomous {D}riving}}},
  shorttitle = {{{TransFuser}}},
  author = {Chitta, Kashyap and Prakash, Aditya and Jaeger, Bernhard and Yu, Zehao and Renz, Katrin and Geiger, Andreas},
  year = {2022},
  journal = {IEEE Transactions on Pattern Analysis and Machine Intelligence},
  pages = {1--18},
  issn = {0162-8828, 2160-9292, 1939-3539},
  doi = {10.1109/TPAMI.2022.3200245},
  urldate = {2023-12-04},
  langid = {english}
}

@inproceedings{choADoPTLiDARSpoofing2023,
  title={{{{A}{D}o{P}{T}}}: {{{L}i{D}{A}{R} {S}poofing {A}ttack {D}etection {B}ased}} on {{{P}oint-{L}evel {T}emporal {C}onsistency}}},
  author={Cho, Minkyoung and Cao, Yulong and Zhou, Zixiang and Mao, Z. Morley},
  booktitle={Proceedings of the British Machine Vision Conference (BMVC)},
  year={2023},
}

@inproceedings{choudharyTalk2BEVLanguageenhancedBird2023,
  title={{{{T}alk2{B}{E}{V}}}: {{{L}anguage-enhanced {B}ird}}'{S}-{E}ye {{{V}iew {M}aps}} for {{{A}utonomous {D}riving}}},
  author={Choudhary, Tushar and Dewangan, Vikrant and Chandhok, Shivam and Priyadarshan, Shubham and Jain, Anushka and Singh, Arun K. and Srivastava, Siddharth and Jatavallabhula, Krishna Murthy and Krishna, K. Madhava},
  booktitle={Proceedings of the IEEE International Conference on Robotics and Automation (ICRA)},
  year={2024},
}

@inproceedings{congSTCrowdMultimodalDataset2022,
  title={{{{S}{T}{C}rowd}}: {{{A} {M}ultimodal {D}ataset}} for {{{P}edestrian {P}erception}} in {{{C}rowded {S}cenes}}},
  author={Cong, Peishan and Zhu, Xinge and Qiao, Feng and Ren, Yiming and Peng, Xidong and Hou, Yuenan and Xu, Lan and Yang, Ruigang and Manocha, Dinesh and Ma, Yuexin},
  booktitle={Proceedings of the IEEE/CVF Conference on Computer Vision and Pattern Recognition (CVPR)},
  year={2022},
}

@inproceedings{courseyFTAEDBenchmarkDataset2024,
  title={{{{F}{T}-{A}{E}{D}}}: {{{B}enchmark {D}ataset}} for {{{E}arly {F}reeway {T}raffic {A}nomalous {E}vent {D}etection}}},
  author={Coursey, Austin and Ji, Junyi and {Quinones-Grueiro}, Marcos and Barbour, William and Zhang, Yuhang and Derr, Tyler and Biswas, Gautam and Work, Daniel B.},
  booktitle={Advances in Neural Information Processing Systems (NeurIPS) Datasets and Benchmarks Track},
  year={2024},
}

@article{cuiDeepLearningImage2022,
  title = {{D}eep {{{L}earning}} for {{{I}mage}} and {{{P}oint {C}loud {F}usion}} in {{{A}utonomous {D}riving}}: {{{A} {R}eview}}},
  shorttitle = {Deep {{Learning}} for {{Image}} and {{Point Cloud Fusion}} in {{Autonomous Driving}}},
  author = {Cui, Yaodong and Chen, Ren and Chu, Wenbo and Chen, Long and Tian, Daxin and Li, Ying and Cao, Dongpu},
  year = {2022},
  month = feb,
  journal = {IEEE Transactions on Intelligent Transportation Systems},
  volume = {23},
  number = {2},
  pages = {722--739},
  issn = {1524-9050, 1558-0016},
  doi = {10.1109/TITS.2020.3023541},
  urldate = {2023-11-26},
  langid = {english}
}

@article{cuiDriveLLMChartingPath2023,
  title = {{{{D}rive{L}{L}{M}}}: {{{C}harting {T}he {P}ath {T}oward {F}ull {A}utonomous {D}riving}} with {{{L}arge {L}anguage {M}odels}}},
  shorttitle = {{{DriveLLM}}},
  author = {Cui, Yaodong and Huang, Shucheng and Zhong, Jiaming and Liu, Zhenan and Wang, Yutong and Sun, Chen and Li, Bai and Wang, Xiao and Khajepour, Amir},
  year = {2023},
  journal = {IEEE Transactions on Intelligent Vehicles},
  pages = {1--15},
  issn = {2379-8904, 2379-8858},
  doi = {10.1109/TIV.2023.3327715},
  urldate = {2023-12-04},
  langid = {english}
}

@inproceedings{cuiMultimodalTrajectoryPredictions2019,
  title = {{M}ultimodal {{{T}rajectory {P}redictions}} for {{{A}utonomous {D}riving}} {U}sing {{{D}eep {C}onvolutional {N}etworks}}},
  booktitle = {2019 {{International Conference}} on {{Robotics}} and {{Automation}} ({{ICRA}})},
  author = {Cui, Henggang and Radosavljevic, Vladan and Chou, Fang-Chieh and Lin, Tsung-Han and Nguyen, Thi and Huang, Tzu-Kuo and Schneider, Jeff and Djuric, Nemanja},
  year = {2019},
  month = may,
  pages = {2090--2096},
  issn = {2577-087X},
  doi = {10.1109/ICRA.2019.8793868}
}

@ARTICLE{10491134,
  author={Cui, Can and Ma, Yunsheng and Cao, Xu and Ye, Wenqian and Wang, Ziran},
  journal={IEEE Intelligent Transportation Systems Magazine},
  title={{R}eceive, {R}eason, and {R}eact: {D}rive as {Y}ou {S}ay, {W}ith {L}arge {L}anguage {M}odels in {A}utonomous {V}ehicles},
  year={2024},
  volume={16},
  number={4},
  pages={81-94},
  doi={10.1109/MITS.2024.3381793}
}

@inproceedings{cuiSurveyMultimodalLarge2023,
  title={{A} {{{S}urvey}} on {{{M}ultimodal {L}arge {L}anguage {M}odels}} for {{{A}utonomous {D}riving}}},
  author={Cui, Can and Ma, Yunsheng and Cao, Xu and Ye, Wenqian and Zhou, Yang and Liang, Kaizhao and Chen, Jintai and Lu, Juanwu and Yang, Zichong and Liao, Kuei-Da and Gao, Tianren and Li, Erlong and Tang, Kun and Cao, Zhipeng and Zhou, Tong and Liu, Ao and Yan, Xinrui and Mei, Shuqi and Cao, Jianguo and Wang, Ziran and Zheng, Chao},
  booktitle={Proceedings of the IEEE/CVF Winter Conference on Applications of Computer Vision Workshops (WACVW)},
  pages={958--979},
  year={2024},
  doi={10.1109/WACVW60836.2024.00106},
}

@inproceedings{wuMultiAgentAutonomousDrivingLLM2025,
  title={{M}ulti-{A}gent {A}utonomous {D}riving {S}ystems with {L}arge {L}anguage {M}odels: {A} {S}urvey of {R}ecent {A}dvances, {R}esources, and {F}uture {D}irections},
  author={Wu, Yaozu and Li, Dongyuan and Chen, Yankai and Jiang, Renhe and Zou, Henry Peng and Huang, Wei-Chieh and Li, Yangning and Fang, Liancheng and Wang, Zhen and Yu, Philip S.},
  booktitle={Findings of the Association for Computational Linguistics: EMNLP 2025},
  pages={12756--12773},
  year={2025},
  doi={10.18653/v1/2025.findings-emnlp.683},
}

@article{fouratiFoundationModelsAutonomous2026,
  title={{F}oundation {M}odels for {A}utonomous {D}riving: {A} {C}omprehensive {S}urvey},
  author={Fourati, Sonda and Jaafar, Wael and Baccar, Noura and Alfattani, Safwan and Langar, Rami},
  journal={Engineering Applications of Artificial Intelligence},
  volume={176},
  pages={114805},
  year={2026},
  doi={10.1016/j.engappai.2026.114805},
}

@inproceedings{dembaVehicletoVehicleCommunicationTechnology2018,
  title = {{V}ehicle-to-{{{V}ehicle {C}ommunication {T}echnology}}},
  booktitle = {2018 {{IEEE International Conference}} on {{Electro}}/{{Information Technology}} ({{EIT}})},
  author = {Demba, Albert and M{\"o}ller, Dietmar P. F.},
  year = {2018},
  month = may,
  pages = {0459--0464},
  issn = {2154-0373},
  doi = {10.1109/EIT.2018.8500189},
  urldate = {2023-11-23}
}

@inproceedings{dengInvestigationByzantineThreats2021,
  title = {{A}n {{{I}nvestigation}} of {{{B}yzantine {T}hreats}} in {{{M}ulti-{R}obot {S}ystems}}},
  booktitle = {24th {{International Symposium}} on {{Research}} in {{Attacks}}, {{Intrusions}} and {{Defenses}}},
  author = {Deng, Gelei and Zhou, Yuan and Xu, Yuan and Zhang, Tianwei and Liu, Yang},
  year = {2021},
  month = oct,
  pages = {17--32},
  publisher = {ACM},
  address = {San Sebastian Spain},
  doi = {10.1145/3471621.3471867},
  urldate = {2023-09-27},
  isbn = {978-1-4503-9058-3},
  langid = {english}
}

@inproceedings{dingDODADataorientedSimtoReal2022,
  title={{{{D}{O}{D}{A}}}: {{{D}ata-oriented {S}im-to-{R}eal {D}omain {A}daptation}} for {{3{D} {S}emantic {S}egmentation}}},
  author={Ding, Runyu and Yang, Jihan and Jiang, Li and Qi, Xiaojuan},
  booktitle={Proceedings of the European Conference on Computer Vision (ECCV)},
  year={2022},
}

@inproceedings{dingHolisticAutonomousDriving2024,
  title={{H}olistic {{{A}utonomous {D}riving {U}nderstanding}} by {{{B}ird}}'{S}-{{{E}ye-{V}iew {I}njected {M}ulti-{M}odal {L}arge {M}odels}}},
  author={Ding, Xinpeng and Han, Jinahua and Xu, Hang and Liang, Xiaodan and Zhang, Wei and Li, Xiaomeng},
  booktitle={Proceedings of the IEEE/CVF Conference on Computer Vision and Pattern Recognition (CVPR)},
  year={2024},
}

@article{dingLearningHelpEmergency2023,
  title = {{L}earning to {{{H}elp {E}mergency {V}ehicles {A}rrive {F}aster}}: {{{A} {C}ooperative {V}ehicle-{R}oad {S}cheduling {A}pproach}}},
  shorttitle = {Learning to {{Help Emergency Vehicles Arrive Faster}}},
  author = {Ding, Lige and Zhao, Dong and Wang, Zhaofeng and Wang, Guang and Tan, Chang and Fan, Lei and Ma, Huadong},
  year = {2023},
  month = oct,
  journal = {IEEE Transactions on Mobile Computing},
  volume = {22},
  number = {10},
  pages = {5949--5962},
  issn = {1536-1233, 1558-0660, 2161-9875},
  doi = {10.1109/TMC.2022.3188344},
  urldate = {2023-12-15},
  langid = {english}
}

@article{dingRobustMultiAgentCommunication2024,
  title = {{R}obust {{{M}ulti-{A}gent {C}ommunication {W}ith {G}raph {I}nformation {B}ottleneck {O}ptimization}}},
  author = {Ding, Shifei and Du, Wei and Ding, Ling and Zhang, Jian and Guo, Lili and An, Bo},
  year = {2024},
  month = may,
  journal = {IEEE Transactions on Pattern Analysis and Machine Intelligence},
  volume = {46},
  number = {5},
  pages = {3096--3107},
  issn = {1939-3539},
  doi = {10.1109/TPAMI.2023.3337534},
  urldate = {2024-07-07}
}

@inproceedings{dosovitskiyCARLAOpenUrban2017,
  title = {{{{C}{A}{R}{L}{A}}}: {{{A}n {O}pen {U}rban {D}riving {S}imulator}}},
  shorttitle = {{{CARLA}}},
  booktitle = {Proceedings of the 1st {{Annual Conference}} on {{Robot Learning}}},
  author = {Dosovitskiy, Alexey and Ros, German and Codevilla, Felipe and Lopez, Antonio and Koltun, Vladlen},
  year = {2017},
  month = oct,
  pages = {1--16},
  publisher = {PMLR},
  issn = {2640-3498},
  urldate = {2023-08-22},
  langid = {english}
}

@inproceedings{driessPaLMEEmbodiedMultimodal2023,
  title={{{{P}a{L}{M}-{E}}}: {{{A}n {E}mbodied {M}ultimodal {L}anguage {M}odel}}},
  author={Driess, Danny and Xia, Fei and Sajjadi, Mehdi S. M. and Lynch, Corey and Chowdhery, Aakanksha and Ichter, Brian and Wahid, Ayzaan and Tompson, Jonathan and Vuong, Quan and Yu, Tianhe and Huang, Wenlong and Chebotar, Yevgen and Sermanet, Pierre and Duckworth, Daniel and Levine, Sergey and Vanhoucke, Vincent and Hausman, Karol and Toussaint, Marc and Greff, Klaus and Zeng, Andy and Mordatch, Igor and Florence, Pete},
  booktitle={Proceedings of the International Conference on Machine Learning (ICML)},
  year={2023},
}

@inproceedings{duanPromptingMultiModalTokens2024,
  title={{P}rompting {{{M}ulti-{M}odal {T}okens}} to {{{E}nhance {E}nd-to-{E}nd {A}utonomous {D}riving {I}mitation {L}earning}} with {{{L}{L}{M}s}}},
  author={Duan, Yiqun and Zhang, Qiang and Xu, Renjing},
  booktitle={Proceedings of the IEEE International Conference on Robotics and Automation (ICRA)},
  year={2024},
}

@article{duFederatedLearningVehicular2020,
  title = {{F}ederated {{{L}earning}} for {{{V}ehicular {I}nternet}} of {{{T}hings}}: {{{R}ecent {A}dvances}} and {{{O}pen {I}ssues}}},
  shorttitle = {Federated {{Learning}} for {{Vehicular Internet}} of {{Things}}},
  author = {Du, Zhaoyang and Wu, Celimuge and Yoshinaga, Tsutomu and Yau, Kok-Lim Alvin and Ji, Yusheng and Li, Jie},
  year = {2020},
  journal = {IEEE Open Journal of the Computer Society},
  volume = {1},
  pages = {45--61},
  issn = {2644-1268},
  doi = {10.1109/OJCS.2020.2992630}
}

@article{duttaComprehensiveReviewRecent2024,
  title = {{A} {{{C}omprehensive {R}eview}} of {{{R}ecent {D}evelopments}} in {{{V}{A}{N}{E}{T}}} for {{{T}raffic}}, {{{S}afety}} \& {{{R}emote {M}onitoring {A}pplications}}},
  author = {Dutta, Arijit and Samaniego Campoverde, Luis Miguel and Tropea, Mauro and De Rango, Floriano},
  year = {2024},
  month = aug,
  journal = {Journal of Network and Systems Management},
  volume = {32},
  number = {4},
  pages = {73},
  issn = {1573-7705},
  doi = {10.1007/s10922-024-09853-5}
}

@article{fangPACPPriorityAwareCollaborative2024,
  title = {{{{P}{A}{C}{P}}}: {{{P}riority-{A}ware {C}ollaborative {P}erception}} for {{{C}onnected}} and {{{A}utonomous {V}ehicles}}},
  shorttitle = {{{PACP}}},
  author = {Fang, Zhengru and Hu, Senkang and An, Haonan and Zhang, Yuang and Wang, Jingjing and Cao, Hangcheng and Chen, Xianhao and Fang, Yuguang},
  year = {2024},
  month = dec,
  journal = {IEEE Transactions on Mobile Computing},
  volume = {23},
  number = {12},
  pages = {15003--15018},
  issn = {1558-0660},
  doi = {10.1109/TMC.2024.3449371},
  urldate = {2025-01-12}
}

@article{fangPrioritizedInformationBottleneck2025,
  title = {{P}rioritized {{{I}nformation {B}ottleneck {T}heoretic {F}ramework {W}ith {D}istributed {O}nline {L}earning}} for {{{E}dge {V}ideo {A}nalytics}}},
  author = {Fang, Zhengru and Hu, Senkang and Wang, Jingjing and Deng, Yiqin and Chen, Xianhao and Fang, Yuguang},
  year = {2025},
  month = jun,
  journal = {IEEE Transactions on Networking},
  volume = {33},
  number = {3},
  pages = {1203--1219},
  issn = {2998-4157},
  doi = {10.1109/TON.2025.3526148},
  urldate = {2025-01-22}
}

@article{maSense4FLVehicularCrowdsensing2026,
  title = {{{S}ense4{F}{L}}: {V}ehicular {C}rowdsensing {E}nhanced {F}ederated {L}earning for {O}bject {D}etection in {A}utonomous {D}riving},
  author = {Ma, Yanan and Hu, Senkang and Fang, Zhengru and Ji, Yun and Deng, Yiqin and Fang, Yuguang},
  journal = {IEEE Transactions on Mobile Computing},
  year = {2026},
  month = aug,
  volume = {25},
  number = {8},
  pages = {13004--13018},
  issn = {2161-9875},
  doi = {10.1109/TMC.2026.3674333},
}

@inproceedings{fangTaskOrientedCommunicationsVisual2025,
  title = {{T}ask-{O}riented {C}ommunications for {V}isual {N}avigation with {E}dge-{A}erial {C}ollaboration in {L}ow {A}ltitude {E}conomy},
  author = {Fang, Zhengru and Liu, Zhenghao and Wang, Jingjing and Hu, Senkang and Guo, Yu and Deng, Yiqin and Fang, Yuguang},
  booktitle = {GLOBECOM 2025 - 2025 IEEE Global Communications Conference},
  year = {2025},
  month = dec,
  pages = {1059--1064},
  publisher = {IEEE},
  doi = {10.1109/GLOBECOM59602.2025.11432022},
}

@article{huang2026skillconditionedgatedselfdistillationllm,
  title = {{S}kill-{C}onditioned {G}ated {S}elf-{D}istillation for {{LLM}} {R}easoning},
  author = {Huang, Jiazhen and Chen, Xiao and Luo, Xiao and Dai, Yong and Hu, Senkang and Zhao, Yuzhi},
  year = {2026},
  eprint = {2605.28791},
  archiveprefix = {arXiv},
  primaryclass = {cs.CL},
  journal = {arXiv preprint},
}

@article{hu2026selfinducedoutcomepotentialturnlevel,
  title = {{S}elf-{I}nduced {O}utcome {P}otential: {T}urn-{L}evel {C}redit {A}ssignment for {A}gents {W}ithout {V}erifiers},
  author = {Hu, Senkang and Dai, Yong and Han, Xudong and Fang, Zhengru and Zhao, Yuzhi and Kwong, Sam Tak Wu and Fang, Yuguang},
  year = {2026},
  eprint = {2605.04984},
  archiveprefix = {arXiv},
  primaryclass = {cs.LG},
  journal = {arXiv preprint},
}

@article{hu2026optimizingagenticreasoningretrieval,
  title = {{O}ptimizing {A}gentic {R}easoning with {R}etrieval via {S}ynthetic {S}emantic {I}nformation {G}ain {R}eward},
  author = {Hu, Senkang and Dai, Yong and Zhao, Yuzhi and Tao, Yihang and Guo, Yu and Fang, Zhengru and Kwong, Sam Tak Wu and Fang, Yuguang},
  year = {2026},
  eprint = {2602.00845},
  archiveprefix = {arXiv},
  primaryclass = {cs.AI},
  journal = {arXiv preprint},
}

@article{zhang2026pelicanunify10unifiedembodied,
  title = {{{P}elican-{U}nify 1.0}: {A} {U}nified {E}mbodied {I}ntelligence {M}odel for {U}nderstanding, {R}easoning, {I}magination and {A}ction},
  author = {Zhang, Yi and Chen, Yinda and Liu, Che and Ding, Zeyuan and Xu, Jin and Zou, Shilong and Liao, Junwei and Hu, Jiayu and Ren, Xiancong and Zhang, Xiaopeng and Liu, Yechi and Shi, Haoyuan and Tang, Zecong and Sun, Haosong and Cui, Renwen and Wu, Kuishu and Liu, Wenhai and Xu, Yang and Zhang, Yingji and Wang, Yidong and Hu, Senkang and Lu, Jinpeng and Chan, Nga Teng and Wu, Yechen and Liu, Zeting and Hou, Xianzhou and Dai, Yong and Tang, Jian and Ju, Xiaozhu},
  year = {2026},
  eprint = {2605.15153},
  archiveprefix = {arXiv},
  primaryclass = {cs.RO},
  journal = {arXiv preprint},
}

@inproceedings{fantauzzoFedDriveGeneralizingFederated2022,
  title = {{{{F}ed{D}rive}}: {{{G}eneralizing {F}ederated {L}earning}} to {{{S}emantic {S}egmentation}} in {{{A}utonomous {D}riving}}},
  shorttitle = {{{FedDrive}}},
  booktitle = {2022 {{IEEE}}/{{RSJ International Conference}} on {{Intelligent Robots}} and {{Systems}} ({{IROS}})},
  author = {Fantauzzo, Lidia and Fan{\`i}, Eros and Caldarola, Debora and Tavera, Antonio and Cermelli, Fabio and Ciccone, Marco and Caputo, Barbara},
  year = {2022},
  month = oct,
  pages = {11504--11511},
  issn = {2153-0866},
  doi = {10.1109/IROS47612.2022.9981098}
}

@article{fengIntelligentDrivingIntelligence2021,
  title = {{I}ntelligent {D}riving {I}ntelligence {T}est for {A}utonomous {V}ehicles with {N}aturalistic and {A}dversarial {E}nvironment},
  author = {Feng, Shuo and Yan, Xintao and Sun, Haowei and Feng, Yiheng and Liu, Henry X.},
  year = {2021},
  month = feb,
  journal = {Nature Communications},
  volume = {12},
  number = {1},
  pages = {748},
  publisher = {Nature Publishing Group},
  issn = {2041-1723},
  doi = {10.1038/s41467-021-21007-8},
  urldate = {2023-12-21},
  copyright = {2021 The Author(s)},
  langid = {english}
}

@inproceedings{fuDriveHumanRethinking2023,
  title = {{D}rive {{{L}ike}} a {{{H}uman}}: {{{R}ethinking {A}utonomous {D}riving}} with {{{L}arge {L}anguage {M}odels}}},
  shorttitle = {Drive {{Like}} a {{Human}}},
  booktitle = {{{WACV}}'24},
  author = {Fu, Daocheng and Li, Xin and Wen, Licheng and Dou, Min and Cai, Pinlong and Shi, Botian and Qiao, Yu},
  year = {2023},
  month = jul,
  publisher = {arXiv},
  urldate = {2023-12-04}
}

@inproceedings{fuLimSimClosedLoopPlatform2024,
  title = {{{{L}im{S}im}}++: {{{A} {C}losed-{L}oop {P}latform}} for {{{D}eploying {M}ultimodal {L}{L}{M}s}} in {{{A}utonomous {D}riving}}},
  shorttitle = {{{LimSim}}++},
  booktitle = {2024 {{IEEE Intelligent Vehicles Symposium}} ({{IV}})},
  author = {Fu, Daocheng and Lei, Wenjie and Wen, Licheng and Cai, Pinlong and Mao, Song and Dou, Min and Shi, Botian and Qiao, Yu},
  year = {2024},
  month = jun,
  pages = {1084--1090},
  issn = {2642-7214},
  doi = {10.1109/IV55156.2024.10588848},
  urldate = {2025-02-09}
}

@article{gaoAutonomousDrivingSecurity2022,
  title = {{A}utonomous {{{D}riving {S}ecurity}}: {{{S}tate}} of the {{{A}rt}} and {{{C}hallenges}}},
  shorttitle = {Autonomous {{Driving Security}}},
  author = {Gao, Cong and Wang, Geng and Shi, Weisong and Wang, Zhongmin and Chen, Yanping},
  year = {2022},
  month = may,
  journal = {IEEE Internet of Things Journal},
  volume = {9},
  number = {10},
  pages = {7572--7595},
  issn = {2327-4662},
  doi = {10.1109/JIOT.2021.3130054}
}

@inproceedings{geigerAreWeReady2012,
  title = {{A}re {W}e {R}eady for {A}utonomous {D}riving? {{{T}he {K}{I}{T}{T}{I}}} {V}ision {B}enchmark {S}uite},
  shorttitle = {Are We Ready for Autonomous Driving?},
  booktitle = {2012 {{IEEE Conference}} on {{Computer Vision}} and {{Pattern Recognition}}},
  author = {Geiger, Andreas and Lenz, Philip and Urtasun, Raquel},
  year = {2012},
  month = jun,
  pages = {3354--3361},
  issn = {1063-6919},
  doi = {10.1109/CVPR.2012.6248074},
  urldate = {2023-11-17}
}

@article{gholamhosseinianComprehensiveSurveyCooperative2022,
  title = {{A} {{{C}omprehensive {S}urvey}} on {{{C}ooperative {I}ntersection {M}anagement}} for {{{H}eterogeneous {C}onnected {V}ehicles}}},
  author = {Gholamhosseinian, Ashkan and Seitz, Jochen},
  year = {2022},
  journal = {IEEE Access},
  volume = {10},
  pages = {7937--7972},
  issn = {2169-3536},
  doi = {10.1109/ACCESS.2022.3142450},
  urldate = {2024-01-20},
  langid = {english}
}

@article{gongEdgeIntelligenceIntelligent2023,
  title = {{E}dge {{{I}ntelligence}} in {{{I}ntelligent {T}ransportation {S}ystems}}: {{{A} {S}urvey}}},
  shorttitle = {Edge {{Intelligence}} in {{Intelligent Transportation Systems}}},
  author = {Gong, Taiyuan and Zhu, Li and Yu, F. Richard and Tang, Tao},
  year = {2023},
  month = sep,
  journal = {IEEE Transactions on Intelligent Transportation Systems},
  volume = {24},
  number = {9},
  pages = {8919--8944},
  issn = {1524-9050, 1558-0016},
  doi = {10.1109/TITS.2023.3275741},
  urldate = {2024-03-21},
  langid = {english}
}

@article{gongSDACMultimodalSynthetic2024,
  title = {{{{S}{D}{A}{C}}}: {{{A} {M}ultimodal {S}ynthetic {D}ataset}} for {{{A}nomaly}} and {{{C}orner {C}ase {D}etection}} in {{{A}utonomous {D}riving}}},
  shorttitle = {{{SDAC}}},
  author = {Gong, Lei and Zhang, Yu and Xia, Yingqing and Zhang, Yanyong and Ji, Jianmin},
  year = {2024},
  month = mar,
  journal = {Proceedings of the AAAI Conference on Artificial Intelligence},
  volume = {38},
  number = {3},
  pages = {1914--1922},
  issn = {2374-3468, 2159-5399},
  doi = {10.1609/aaai.v38i3.27961},
  urldate = {2024-09-11},
  langid = {english}
}

@article{guanWorldModelsAutonomous2024,
  title={{W}orld {M}odels for {A}utonomous {D}riving: {A}n {I}nitial {S}urvey},
  author={Guan, Yanchen and Liao, Haicheng and Li, Zhenning and Hu, Jia and Yuan, Runze and Li, Yunjian and Zhang, Guohui and Xu, Chengzhong},
  journal={IEEE Transactions on Intelligent Vehicles},
  pages={1--17},
  year={2024},
  doi={10.1109/TIV.2024.3398357},
}

@article{guoDeepLearning3d2020,
  title={{D}eep {{{L}earning}} for 3d {{{P}oint {C}louds}}: {{{A} {S}urvey}}},
  author={Guo, Yulan and Wang, Hanyun and Hu, Qingyong and Liu, Hao and Liu, Li and Bennamoun, Mohammed},
  journal={IEEE Transactions on Pattern Analysis and Machine Intelligence},
  year={2021},
}

@inproceedings{halderPhysicsBasedRenderingImproving2019,
  title = {{P}hysics-{{{B}ased {R}endering}} for {{{I}mproving {R}obustness}} to {{{R}ain}}},
  booktitle = {2019 {{IEEE}}/{{CVF International Conference}} on {{Computer Vision}} ({{ICCV}})},
  author = {Halder, Shirsendu and Lalonde, Jean-Francois and Charette, Raoul De},
  year = {2019},
  month = oct,
  pages = {10202--10211},
  publisher = {IEEE},
  address = {Seoul, Korea (South)},
  doi = {10.1109/ICCV.2019.01030},
  urldate = {2023-10-14},
  isbn = {978-1-7281-4803-8},
  langid = {english}
}

@article{hallyburtonMultiAgentSecurityTestbed2024,
  title = {{A} {{{M}ulti-{A}gent {S}ecurity {T}estbed}} for the {{{A}nalysis}} of {{{A}ttacks}} and {{{D}efenses}} in {{{C}ollaborative {S}ensor {F}usion}}},
  author = {Hallyburton, R. Spencer and Hunt, David and Luo, Shaocheng and Pajic, Miroslav},
  year = {2024},
  month = jan,
  eprint = {2401.09387},
  primaryclass = {cs, eess},
  publisher = {arXiv},
  urldate = {2024-07-08},
  archiveprefix = {arXiv},
  langid = {english},
  journal = {arXiv preprint},
}

@article{hallyburtonPartialInformationLongitudinalCyber2023,
  title = {{P}artial-{{{I}nformation}}, {{{L}ongitudinal {C}yber {A}ttacks}} on {{{L}i{D}{A}{R}}} in {{{A}utonomous {V}ehicles}}},
  author = {Hallyburton, R. Spencer and Zhang, Qingzhao and Mao, Z. Morley and Pajic, Miroslav},
  year = {2023},
  month = apr,
  eprint = {2303.03470},
  primaryclass = {cs, eess},
  publisher = {arXiv},
  urldate = {2023-11-28},
  archiveprefix = {arXiv},
  langid = {english},
  journal = {arXiv preprint},
}

@article{hanADSLeadLifelongAnomaly2023,
  title = {{{{A}{D}{S}-{L}ead}}: {{{L}ifelong {A}nomaly {D}etection}} in {{{A}utonomous {D}riving {S}ystems}}},
  shorttitle = {{{ADS-Lead}}},
  author = {Han, Xingshuo and Zhou, Yuan and Chen, Kangjie and Qiu, Han and Qiu, Meikang and Liu, Yang and Zhang, Tianwei},
  year = {2023},
  month = jan,
  journal = {IEEE Transactions on Intelligent Transportation Systems},
  volume = {24},
  number = {1},
  pages = {1039--1051},
  issn = {1524-9050, 1558-0016},
  doi = {10.1109/TITS.2021.3122906},
  urldate = {2023-09-27},
  langid = {english}
}

@article{hanCollaborativePerceptionAutonomous2023,
  title = {{C}ollaborative {{{P}erception}} in {{{A}utonomous {D}riving}}: {{{M}ethods}}, {{{D}atasets}} and {{{C}hallenges}}},
  shorttitle = {Collaborative {{Perception}} in {{Autonomous Driving}}},
  author = {Han, Yushan and Zhang, Hui and Li, Huifang and Jin, Yi and Lang, Congyan and Li, Yidong},
  year = {2023},
  month = nov,
  journal = {IEEE Intelligent Transportation Systems Magazine},
  volume = {15},
  number = {6},
  pages = {131--151},
  issn = {1939-1390, 1941-1197},
  doi = {10.1109/MITS.2023.3298534},
  urldate = {2023-11-20}
}

@article{hangCooperativeDecisionMaking2022,
  title = {{C}ooperative {{{D}ecision {M}aking}} of {{{C}onnected {A}utomated {V}ehicles}} at {{{M}ulti-{L}ane {M}erging {Z}one}}: {{{A} {C}oalitional {G}ame {A}pproach}}},
  shorttitle = {Cooperative {{Decision Making}} of {{Connected Automated Vehicles}} at {{Multi-Lane Merging Zone}}},
  author = {Hang, Peng and Lv, Chen and Huang, Chao and Xing, Yang and Hu, Zhongxu},
  year = {2022},
  month = apr,
  journal = {IEEE Transactions on Intelligent Transportation Systems},
  volume = {23},
  number = {4},
  pages = {3829--3841},
  issn = {1524-9050, 1558-0016},
  doi = {10.1109/TITS.2021.3069463},
  urldate = {2024-01-20},
  langid = {english}
}

@article{hangDecisionMakingConnected2021,
  title = {{D}ecision {{{M}aking}} of {{{C}onnected {A}utomated {V}ehicles}} at an {{{U}nsignalized {R}oundabout {C}onsidering {P}ersonalized {D}riving {B}ehaviours}}},
  author = {Hang, Peng and Huang, Chao and Hu, Zhongxu and Xing, Yang and Lv, Chen},
  year = {2021},
  month = may,
  journal = {IEEE Transactions on Vehicular Technology},
  volume = {70},
  number = {5},
  pages = {4051--4064},
  issn = {0018-9545, 1939-9359},
  doi = {10.1109/TVT.2021.3072676},
  urldate = {2024-01-20},
  langid = {english}
}

@article{hangDecisionMakingConnected2022,
  title = {{D}ecision {{{M}aking}} for {{{C}onnected {A}utomated {V}ehicles}} at {{{U}rban {I}ntersections {C}onsidering {S}ocial}} and {{{I}ndividual {B}enefits}}},
  author = {Hang, Peng and Huang, Chao and Hu, Zhongxu and Lv, Chen},
  year = {2022},
  month = nov,
  journal = {IEEE Transactions on Intelligent Transportation Systems},
  volume = {23},
  number = {11},
  pages = {22549--22562},
  issn = {1524-9050, 1558-0016},
  doi = {10.1109/TITS.2022.3209607},
  urldate = {2024-01-20},
  langid = {english}
}

@article{hobertEnhancementsV2XCommunication2015,
  title = {{E}nhancements of {{{V}2{X}}} {C}ommunication in {S}upport of {C}ooperative {A}utonomous {D}riving},
  author = {Hobert, Laurens and Festag, Andreas and Llatser, Ignacio and Altomare, Luciano and Visintainer, Filippo and Kovacs, Andras},
  year = {2015},
  month = dec,
  journal = {IEEE Communications Magazine},
  volume = {53},
  number = {12},
  pages = {64--70},
  issn = {0163-6804},
  doi = {10.1109/MCOM.2015.7355568},
  urldate = {2023-10-13},
  langid = {english}
}

@article{houMultilevelMultimodalFeature2022,
  title = {{M}ulti-{L}evel and {M}ulti-{M}odal {F}eature {F}usion for {A}ccurate {{3{D}}} {O}bject {D}etection in {{{C}onnected}} and {{{A}utomated {V}ehicles}}},
  author = {Hou, Yiming and Rezaei, Mahdi and Romano, Richard},
  year = {2022},
  month = dec,
  eprint = {2212.07560},
  primaryclass = {cs, eess},
  publisher = {arXiv},
  doi = {10.48550/arXiv.2212.07560},
  urldate = {2023-11-26},
  archiveprefix = {arXiv},
  journal = {arXiv preprint},
}

@inproceedings{huAdaptiveCommunicationsCollaborative2023,
  title = {{A}daptive {{{C}ommunications}} in {{{C}ollaborative {P}erception}} with {{{D}omain {A}lignment}} for {{{A}utonomous {D}riving}}},
  author = {Hu, Senkang and Fang, Zhengru and An, Haonan and Xu, Guowen and Zhou, Yuan and Chen, Xianhao and Fang, Yuguang},
  booktitle = {2024 {{IEEE}} Global Communications Conference ({{GLOBECOM}})},
  year = {2024},
  pages = {746--751},
  publisher = {IEEE}
}

@article{anChannelAwareThroughputCooperative2024,
  title = {{C}hannel-{A}ware {T}hroughput {M}aximization for {C}ooperative {D}ata {F}usion in {C}onnected and {A}utonomous {V}ehicles},
  author = {An, Haonan and Fang, Zhengru and Zhang, Yuang and Hu, Senkang and Chen, Xianhao and Xu, Guowen and Fang, Yuguang},
  year = {2024},
  eprint = {2410.04320},
  archiveprefix = {arXiv},
  primaryclass = {cs},
  doi = {10.48550/arXiv.2410.04320},
  note = {arXiv:2410.04320},
  journal = {arXiv preprint}
}

@article{huAgentsCoDriverLargeLanguage2024,
  title = {{{{A}gents{C}o{D}river}}: {{{L}arge {L}anguage {M}odel {E}mpowered {C}ollaborative {D}riving}} with {{{L}ifelong {L}earning}}},
  shorttitle = {{{AgentsCoDriver}}},
  author = {Hu, Senkang and Fang, Zhengru and Fang, Zihan and Deng, Yiqin and Chen, Xianhao and Fang, Yuguang},
  year = {2024},
  month = apr,
  eprint = {2404.06345},
  primaryclass = {cs},
  publisher = {arXiv},
  urldate = {2024-05-20},
  archiveprefix = {arXiv},
  copyright = {All rights reserved},
  langid = {english},
  journal = {arXiv preprint},
}

@article{huAgentsCoMergeLargeLanguage2024,
  title = {{{{A}gents{C}o{M}erge}}: {{{L}arge {L}anguage {M}odel {E}mpowered {C}ollaborative {D}ecision {M}aking}} for {{{R}amp {M}erging}}},
  shorttitle = {{{AgentsCoMerge}}},
  author = {Hu, Senkang and Fang, Zhengru and Fang, Zihan and Deng, Yiqin and Chen, Xianhao and Fang, Yuguang and Kwong, Sam},
  journal = {IEEE Transactions on Mobile Computing},
  year = {2025},
  volume = {24},
  number = {10},
  pages = {9791--9805},
  doi = {10.1109/TMC.2025.3564163}
}

@inproceedings{huCollaborationHelpsCamera2023,
  title = {{C}ollaboration {{{H}elps {C}amera {O}vertake {L}i{D}{A}{R}}} in {{3{D} {D}etection}}},
  booktitle = {2023 {{IEEE}}/{{CVF Conference}} on {{Computer Vision}} and {{Pattern Recognition}} ({{CVPR}})},
  author = {Hu, Yue and Lu, Yifan and Xu, Runsheng and Xie, Weidi and Chen, Siheng and Wang, Yanfeng},
  year = {2023},
  month = jun,
  pages = {9243--9252},
  publisher = {IEEE},
  address = {Vancouver, BC, Canada},
  doi = {10.1109/CVPR52729.2023.00892},
  urldate = {2024-04-20},
  copyright = {https://doi.org/10.15223/policy-029},
  isbn = {979-8-3503-0129-8},
  langid = {english}
}

@inproceedings{huCommunicationEfficientCollaborativePerception2024,
  title = {{C}ommunication-{{{E}fficient {C}ollaborative {P}erception}} via {{{I}nformation {F}illing}} with {{{C}odebook}}},
  booktitle = {2024 {{IEEE}}/{{CVF Conference}} on {{Computer Vision}} and {{Pattern Recognition}} ({{CVPR}})},
  author = {Hu, Yue and Peng, Juntong and Liu, Sifei and Ge, Junhao and Liu, Si and Chen, Siheng},
  year = {2024},
  month = jun,
  pages = {15481--15490},
  publisher = {IEEE},
  address = {Seattle, WA, USA},
  doi = {10.1109/CVPR52733.2024.01466},
  urldate = {2024-10-28},
  copyright = {https://doi.org/10.15223/policy-029},
  isbn = {979-8-3503-5300-6},
  langid = {english}
}

@inproceedings{huCPGuardMaliciousAgent2024,
  title = {{{C}{P}-{G}uard}: {M}alicious {A}gent {D}etection and {D}efense in {C}ollaborative {B}{ird'S} {E}ye {V}iew {P}erception},
  author = {Hu, Senkang and Tao, Yihang and Xu, Guowen and Deng, Yiqin and Chen, Xianhao and Fang, Yuguang and Kwong, Sam},
  booktitle = {Proceedings of the AAAI Conference on Artificial Intelligence},
  volume = {39},
  number = {22},
  pages = {23203--23211},
  year = {2025},
  doi = {10.1609/aaai.v39i22.34486},
}

@inproceedings{taoDirectedCPEfficientCollaborative2025,
  title = {{D}irected-{C}{P}: {E}fficient {C}ollaborative {P}erception by {D}irected {I}nformation {F}low},
  author = {Tao, Yihang and Hu, Senkang and Fang, Zhengru and Fang, Yuguang},
  booktitle = {2025 {{IEEE}} International Conference on Robotics and Automation ({{ICRA}})},
  year = {2025},
  pages = {7004--7010},
  doi = {10.1109/ICRA55743.2025.11127818},
  publisher = {IEEE}
}

@article{taoGCPGuardedCollaborative2025,
  title = {{{G}{C}{P}}: {G}uarded {C}ollaborative {P}erception with {S}patial-{T}emporal {A}ware {M}alicious {A}gent {D}etection},
  author = {Tao, Yihang and Hu, Senkang and Hu, Yue and An, Haonan and Cao, Hangcheng and Fang, Yuguang},
  journal = {IEEE Transactions on Dependable and Secure Computing},
  year = {2026},
  pages = {1--14},
  doi = {10.1109/TDSC.2026.3693684},
}

@article{huCPGuardPlusNewParadigm2025,
  title = {{{{C}{P}-{G}uard}}+: {A} {N}ew {P}aradigm for {M}alicious {A}gent {D}etection and {D}efense in {C}ollaborative {P}erception},
  author = {Hu, Senkang and Tao, Yihang and Fang, Zihan and Xu, Guowen and Deng, Yiqin and Kwong, Sam and Fang, Yuguang},
  year = {2025},
  eprint = {2502.07807},
  archiveprefix = {arXiv},
  primaryclass = {cs},
  doi = {10.48550/arXiv.2502.07807},
  note = {arXiv:2502.07807},
  journal = {arXiv preprint}
}

@article{huCPUniGuardUnified2026,
  title = {{{{C}{P}-{U}ni{G}uard}}: {U}nified {M}alicious {A}gent {D}etection and {D}efense in {C}ollaborative {P}erception},
  author = {Hu, Senkang and Tao, Yihang and Xu, Guowen and Qian, Xinyuan and Deng, Yiqin and Chen, Xianhao and Kwong, Sam Tak Wu and Fang, Yuguang},
  journal = {IEEE Transactions on Mobile Computing},
  year = {2026},
  month = jan,
  note = {Early Access},
  doi = {10.1109/TMC.2026.3650980}
}

@article{huFullSceneDomainGeneralization2024,
  title = {{T}oward {{{F}ull-{S}cene {D}omain {G}eneralization}} in {{{M}ulti-{A}gent {C}ollaborative {B}ird}}'{S} {{{E}ye {V}iew {S}egmentation}} for {{{C}onnected}} and {{{A}utonomous {D}riving}}},
  author = {Hu, Senkang and Fang, Zhengru and Deng, Yiqin and Chen, Xianhao and Fang, Yuguang and Kwong, Sam},
  journal = {IEEE Transactions on Intelligent Transportation Systems},
  year = {2025},
  volume = {26},
  number = {2},
  pages = {1783--1796},
  issn = {1558-0016},
  doi = {10.1109/TITS.2024.3506284},
  urldate = {2025-04-11}
}

@inproceedings{huPlanningorientedAutonomousDriving2023,
  title = {{P}lanning-{O}riented {{{A}utonomous {D}riving}}},
  booktitle = {2023 {{IEEE}}/{{CVF Conference}} on {{Computer Vision}} and {{Pattern Recognition}} ({{CVPR}})},
  author = {Hu, Yihan and Yang, Jiazhi and Chen, Li and Li, Keyu and Sima, Chonghao and Zhu, Xizhou and Chai, Siqi and Du, Senyao and Lin, Tianwei and Wang, Wenhai and Lu, Lewei and Jia, Xiaosong and Liu, Qiang and Dai, Jifeng and Qiao, Yu and Li, Hongyang},
  year = {2023},
  month = jun,
  pages = {17853--17862},
  publisher = {IEEE},
  address = {Vancouver, BC, Canada},
  doi = {10.1109/CVPR52729.2023.01712},
  urldate = {2023-12-04},
  isbn = {979-8-3503-0129-8},
  langid = {english}
}

@article{huPragmaticCommunicationMultiAgent2024,
  title = {{P}ragmatic {{{C}ommunication}} in {{{M}ulti-{A}gent {C}ollaborative {P}erception}}},
  author = {Hu, Yue and Pang, Xianghe and Qin, Xiaoqi and Eldar, Yonina C. and Chen, Siheng and Zhang, Ping and Zhang, Wenjun},
  journal = {IEEE Transactions on Pattern Analysis and Machine Intelligence},
  year = {2026},
  urldate = {2024-06-15},
  langid = {english},
  journal = {arXiv preprint},
}

@incollection{huSTP3EndtoEndVisionBased2022,
  title = {{{{S}{T}-{P}3}}: {{{E}nd-to-{E}nd {V}ision-{B}ased {A}utonomous {D}riving}} via {{{S}patial-{T}emporal {F}eature {L}earning}}},
  shorttitle = {{{ST-P3}}},
  booktitle = {Computer {{Vision}} -- {{ECCV}} 2022},
  author = {Hu, Shengchao and Chen, Li and Wu, Penghao and Li, Hongyang and Yan, Junchi and Tao, Dacheng},
  editor = {Avidan, Shai and Brostow, Gabriel and Ciss{\'e}, Moustapha and Farinella, Giovanni Maria and Hassner, Tal},
  year = {2022},
  volume = {13698},
  pages = {533--549},
  publisher = {Springer Nature Switzerland},
  address = {Cham},
  doi = {10.1007/978-3-031-19839-7_31},
  urldate = {2024-06-07},
  isbn = {978-3-031-19838-0 978-3-031-19839-7},
  langid = {english}
}

@inproceedings{huWhere2commCommunicationefficientCollaborative2024,
  title = {{W}here2comm: {C}ommunication-{E}fficient {C}ollaborative {P}erception via {S}patial {C}onfidence {M}aps},
  shorttitle = {Where2comm},
  booktitle = {Proceedings of the 36th {{International Conference}} on {{Neural Information Processing Systems}}},
  author = {Hu, Yue and Fang, Shaoheng and Lei, Zixing and Zhong, Yiqi and Chen, Siheng},
  year = {2024},
  month = apr,
  series = {{{NIPS}} '22},
  pages = {4874--4886},
  publisher = {Curran Associates Inc.},
  address = {Red Hook, NY, USA},
  urldate = {2024-06-03},
  isbn = {978-1-7138-7108-8}
}

@article{itoCoordinationConnectedVehicles2019,
  title = {{C}oordination of {{{C}onnected {V}ehicles}} on {{{M}erging {R}oads {U}sing {P}seudo-{P}erturbation-{B}ased {B}roadcast {C}ontrol}}},
  author = {Ito, Yuji and Kamal, Md Abdus Samad and Yoshimura, Takayoshi and Azuma, Shun-ichi},
  year = {2019},
  month = sep,
  journal = {IEEE Transactions on Intelligent Transportation Systems},
  volume = {20},
  number = {9},
  pages = {3496--3512},
  issn = {1524-9050, 1558-0016},
  doi = {10.1109/TITS.2018.2876905},
  urldate = {2024-03-17},
  langid = {english}
}

@article{jiaAMPAutoregressiveMotion2024,
  title = {{{{A}{M}{P}}}: {{{A}utoregressive {M}otion {P}rediction {R}evisited}} with {{{N}ext {T}oken {P}rediction}} for {{{A}utonomous {D}riving}}},
  shorttitle = {{{AMP}}},
  author = {Jia, Xiaosong and Shi, Shaoshuai and Chen, Zijun and Jiang, Li and Liao, Wenlong and He, Tao and Yan, Junchi},
  year = {2024},
  month = mar,
  eprint = {2403.13331},
  primaryclass = {cs},
  publisher = {arXiv},
  urldate = {2024-04-10},
  archiveprefix = {arXiv},
  journal = {arXiv preprint},
}

@article{jiaHDGTHeterogeneousDriving2023,
  title={{{{H}{D}{G}{T}}}: {{{H}eterogeneous {D}riving {G}raph {T}ransformer}} for {{{M}ulti-{A}gent {T}rajectory {P}rediction}} via {{{S}cene {E}ncoding}}},
  author={Jia, Xiaosong and Wu, Penghao and Chen, Li and Liu, Yu and Li, Hongyang and Yan, Junchi},
  journal={IEEE Transactions on Pattern Analysis and Machine Intelligence},
  year={2023},
}

@article{karbalaiealiDynamicAdaptiveAlgorithm2020,
  title = {{A} {{{D}ynamic {A}daptive {A}lgorithm}} for {{{M}erging {I}nto {P}latoons}} in {{{C}onnected {A}utomated {E}nvironments}}},
  author = {Karbalaieali, Sogand and Osman, Osama A. and Ishak, Sherif},
  year = {2020},
  month = oct,
  journal = {IEEE Transactions on Intelligent Transportation Systems},
  volume = {21},
  number = {10},
  pages = {4111--4122},
  issn = {1524-9050, 1558-0016},
  doi = {10.1109/TITS.2019.2938728},
  urldate = {2024-03-18},
  langid = {english}
}

@inproceedings{kimDriveFuzzDiscoveringAutonomous2022,
  title = {{{{D}rive{F}uzz}}: {{{D}iscovering {A}utonomous {D}riving {B}ugs}} {T}hrough {{{D}riving {Q}uality-{G}uided {F}uzzing}}},
  shorttitle = {{{DriveFuzz}}},
  booktitle = {Proceedings of the 2022 {{ACM SIGSAC Conference}} on {{Computer}} and {{Communications Security}}},
  author = {Kim, Seulbae and Liu, Major and Rhee, Junghwan "John" and Jeon, Yuseok and Kwon, Yonghwi and Kim, Chung Hwan},
  year = {2022},
  month = nov,
  pages = {1753--1767},
  publisher = {ACM},
  address = {Los Angeles CA USA},
  doi = {10.1145/3548606.3560558},
  urldate = {2023-09-22},
  isbn = {978-1-4503-9450-5},
  langid = {english}
}

@inproceedings{kimPGFUZZPolicyGuidedFuzzing2021,
  title = {{{{P}{G}{F}{U}{Z}{Z}}}: {{{P}olicy-{G}uided {F}uzzing}} for {{{R}obotic {V}ehicles}}},
  shorttitle = {{{PGFUZZ}}},
  booktitle = {Proceedings 2021 {{Network}} and {{Distributed System Security Symposium}}},
  author = {Kim, Hyungsub and Ozmen, Muslum Ozgur and Bianchi, Antonio and Celik, Z. Berkay and Xu, Dongyan},
  year = {2021},
  publisher = {Internet Society},
  address = {Virtual},
  doi = {10.14722/ndss.2021.24096},
  urldate = {2023-09-22},
  isbn = {978-1-891562-66-2},
  langid = {english}
}

@article{kimVehicletoVehicleV2VMessage2019,
  title = {{V}ehicle-to-{{{V}ehicle}} ({{{V}2{V}}}) {{{M}essage {C}ontent {P}lausibility {C}heck}} for {{{P}latoons}} {T}hrough {{{L}ow-{P}ower {B}eaconing}}},
  author = {Kim, Hyogon and Kim, Taeho},
  year = {2019},
  month = dec,
  journal = {Sensors},
  volume = {19},
  number = {24},
  pages = {5493},
  issn = {1424-8220},
  doi = {10.3390/s19245493},
  urldate = {2024-08-07},
  copyright = {https://creativecommons.org/licenses/by/4.0/},
  langid = {english}
}

@inproceedings{kungRiskBenchScenariobasedBenchmark2024,
  title = {{{{R}isk{B}ench}}: {{{A} {S}cenario-based {B}enchmark}} for {{{R}isk {I}dentification}}},
  shorttitle = {{{RiskBench}}},
  booktitle = {2024 {{IEEE International Conference}} on {{Robotics}} and {{Automation}} ({{ICRA}})},
  author = {Kung, Chi-Hsi and Yang, Chieh-Chi and Pao, Pang-Yuan and Lu, Shu-Wei and Chen, Pin-Lun and Lu, Hsin-Cheng and Chen, Yi-Ting},
  year = {2024},
  month = may,
  pages = {14800--14807},
  doi = {10.1109/ICRA57147.2024.10610270},
  urldate = {2024-10-25}
}

@inproceedings{leiLatencyAwareCollaborativePerception2022a,
  title = {{L}atency-{{{A}ware {C}ollaborative {P}erception}}},
  booktitle = {Computer {{Vision}} -- {{ECCV}} 2022},
  author = {Lei, Zixing and Ren, Shunli and Hu, Yue and Zhang, Wenjun and Chen, Siheng},
  editor = {Avidan, Shai and Brostow, Gabriel and Ciss{\'e}, Moustapha and Farinella, Giovanni Maria and Hassner, Tal},
  year = {2022},
  pages = {316--332},
  publisher = {Springer Nature Switzerland},
  address = {Cham},
  doi = {10.1007/978-3-031-19824-3_19},
  isbn = {978-3-031-19824-3},
  langid = {english}
}

@article{liAttackingCooperativeMultiAgent2023,
  title={{A}ttacking {{{C}ooperative {M}ulti-{A}gent {R}einforcement {L}earning}} by {{{A}dversarial {M}inority {I}nfluence}}},
  author={Li, Simin and Guo, Jun and Xiu, Jingqiao and Feng, Pu and Yu, Xin and Liu, Aishan and Wu, Wenjun and Liu, Xianglong},
  journal={Neural Networks},
  year={2025},
}

@article{liAutomaticTargetlessLiDAR2023,
  title = {{A}utomatic {T}argetless {{{L}i{D}{A}{R}}}--{C}amera {C}alibration: {A} {S}urvey},
  shorttitle = {Automatic Targetless {{LiDAR}}--Camera Calibration},
  author = {Li, Xingchen and Xiao, Yuxuan and Wang, Beibei and Ren, Haojie and Zhang, Yanyong and Ji, Jianmin},
  year = {2023},
  month = sep,
  journal = {Artificial Intelligence Review},
  volume = {56},
  number = {9},
  pages = {9949--9987},
  issn = {1573-7462},
  doi = {10.1007/s10462-022-10317-y},
  urldate = {2023-11-26},
  langid = {english}
}

@incollection{liBEVFormerLearningBird2022,
  title = {{{{B}{E}{V}{F}ormer}}: {{{L}earning {B}ird}}'{S}-{{{E}ye-{V}iew {R}epresentation}} from {{{M}ulti-camera {I}mages}} via {{{S}patiotemporal {T}ransformers}}},
  shorttitle = {{{BEVFormer}}},
  booktitle = {Computer {{Vision}} -- {{ECCV}} 2022},
  author = {Li, Zhiqi and Wang, Wenhai and Li, Hongyang and Xie, Enze and Sima, Chonghao and Lu, Tong and Qiao, Yu and Dai, Jifeng},
  editor = {Avidan, Shai and Brostow, Gabriel and Ciss{\'e}, Moustapha and Farinella, Giovanni Maria and Hassner, Tal},
  year = {2022},
  volume = {13669},
  pages = {1--18},
  publisher = {Springer Nature Switzerland},
  address = {Cham},
  doi = {10.1007/978-3-031-20077-9_1},
  urldate = {2023-12-07},
  isbn = {978-3-031-20076-2 978-3-031-20077-9},
  langid = {english}
}

@article{liDelvingDevilsBird2024,
  title = {{D}elving {{{I}nto}} the {{{D}evils}} of {{{B}ird}}'{S}-{{{E}ye-{V}iew {P}erception}}: {{{A} {R}eview}}, {{{E}valuation}} and {{{R}ecipe}}},
  shorttitle = {Delving {{Into}} the {{Devils}} of {{Bird}}'s-{{Eye-View Perception}}},
  author = {Li, Hongyang and Sima, Chonghao and Dai, Jifeng and Wang, Wenhai and Lu, Lewei and Wang, Huijie and Zeng, Jia and Li, Zhiqi and Yang, Jiazhi and Deng, Hanming and Tian, Hao and Xie, Enze and Xie, Jiangwei and Chen, Li and Li, Tianyu and Li, Yang and Gao, Yulu and Jia, Xiaosong and Liu, Si and Shi, Jianping and Lin, Dahua and Qiao, Yu},
  year = {2024},
  month = apr,
  journal = {IEEE Transactions on Pattern Analysis and Machine Intelligence},
  volume = {46},
  number = {4},
  pages = {2151--2170},
  issn = {0162-8828, 2160-9292, 1939-3539},
  doi = {10.1109/TPAMI.2023.3333838},
  urldate = {2024-03-29},
  copyright = {https://creativecommons.org/licenses/by-nc-nd/4.0/},
  langid = {english}
}

@article{liKnowledgedrivenAutonomousDriving2023,
  title = {{T}owards {{{K}nowledge-driven {A}utonomous {D}riving}}},
  author = {Li, Xin and Bai, Yeqi and Cai, Pinlong and Wen, Licheng and Fu, Daocheng and Zhang, Bo and Yang, Xuemeng and Cai, Xinyu and Ma, Tao and Guo, Jianfei and Gao, Xing and Dou, Min and Shi, Botian and Liu, Yong and He, Liang and Qiao, Yu},
  year = {2023},
  month = dec,
  eprint = {2312.04316},
  primaryclass = {cs},
  publisher = {arXiv},
  urldate = {2023-12-09},
  archiveprefix = {arXiv},
  journal = {arXiv preprint},
}

@inproceedings{liLearningDistilledCollaboration2021,
  title = {{L}earning {{{D}istilled {C}ollaboration {G}raph}} for {{{M}ulti-{A}gent {P}erception}}},
  booktitle = {Advances in {{Neural Information Processing Systems}}},
  author = {Li, Yiming and Ren, Shunli and Wu, Pengxiang and Chen, Siheng and Feng, Chen and Zhang, Wenjun},
  year = {2021},
  volume = {34},
  pages = {29541--29552},
  publisher = {Curran Associates, Inc.},
  urldate = {2023-08-13}
}

@article{liLearningVehicletoVehicleCooperative2023,
  title = {{L}earning for {{{V}ehicle-to-{V}ehicle {C}ooperative {P}erception}} {U}nder {{{L}ossy {C}ommunication}}},
  author = {Li, Jinlong and Xu, Runsheng and Liu, Xinyu and Ma, Jin and Chi, Zicheng and Ma, Jiaqi and Yu, Hongkai},
  year = {2023},
  month = apr,
  journal = {IEEE Transactions on Intelligent Vehicles},
  volume = {8},
  number = {4},
  pages = {2650--2660},
  issn = {2379-8904, 2379-8858},
  doi = {10.1109/TIV.2023.3260040},
  urldate = {2023-09-13},
  langid = {english}
}

@article{liMobilityawareDynamicOffloading2021,
  title = {{M}obility-{A}ware {D}ynamic {O}ffloading {S}trategy for {{{C}-{V}2{X}}} {U}nder {M}ulti-{A}ccess {E}dge {C}omputing},
  author = {Li, Bo and Chen, Feilong and Peng, Ziyi and Hou, Peng and Ding, Hongwei},
  year = {2021},
  month = dec,
  journal = {Physical Communication},
  volume = {49},
  pages = {101446},
  issn = {1874-4907},
  doi = {10.1016/j.phycom.2021.101446},
  urldate = {2023-11-23}
}

@article{liPrivacyPreservedFederatedLearning2022,
  title = {{P}rivacy-{{{P}reserved {F}ederated {L}earning}} for {{{A}utonomous {D}riving}}},
  author = {Li, Yijing and Tao, Xiaofeng and Zhang, Xuefei and Liu, Junjie and Xu, Jin},
  year = {2022},
  month = jul,
  journal = {IEEE Transactions on Intelligent Transportation Systems},
  volume = {23},
  number = {7},
  pages = {8423--8434},
  issn = {1558-0016},
  doi = {10.1109/TITS.2021.3081560}
}

@inproceedings{liS2RViTMultiAgentCooperative2023,
  title={{{{S}2{R}-{V}i{T}}} for {{{M}ulti-{A}gent {C}ooperative {P}erception}}: {{{B}ridging}} the {{{G}ap}} from {{{S}imulation}} to {{{R}eality}}},
  author={Li, Jinlong and Xu, Runsheng and Liu, Xinyu and Li, Baolu and Zou, Qin and Ma, Jiaqi and Yu, Hongkai},
  booktitle={Proceedings of the IEEE International Conference on Robotics and Automation (ICRA)},
  year={2024},
}

@inproceedings{liScenarioNetOpenSourcePlatform2023,
  title={{{{S}cenario{N}et}}: {{{O}pen-{S}ource {P}latform}} for {{{L}arge-{S}cale {T}raffic {S}cenario {S}imulation}} and {{{M}odeling}}},
  author={Li, Quanyi and Peng, Zhenghao and Feng, Lan and Liu, Zhizheng and Duan, Chenda and Mo, Wenjie and Zhou, Bolei},
  booktitle={Advances in Neural Information Processing Systems (NeurIPS) Datasets and Benchmarks Track},
  year={2023},
}

@inproceedings{liuBEVFusionMultiTaskMultiSensor2023,
  title = {{{{B}{E}{V}{F}usion}}: {{{M}ulti-{T}ask {M}ulti-{S}ensor {F}usion}} with {{{U}nified {B}ird}}'{S}-{{{E}ye {V}iew {R}epresentation}}},
  shorttitle = {{{BEVFusion}}},
  booktitle = {2023 {{IEEE International Conference}} on {{Robotics}} and {{Automation}} ({{ICRA}})},
  author = {Liu, Zhijian and Tang, Haotian and Amini, Alexander and Yang, Xinyu and Mao, Huizi and Rus, Daniela L. and Han, Song},
  year = {2023},
  month = may,
  pages = {2774--2781},
  publisher = {IEEE},
  address = {London, United Kingdom},
  doi = {10.1109/ICRA48891.2023.10160968},
  urldate = {2023-12-31},
  isbn = {979-8-3503-2365-8},
  langid = {english}
}

@article{10979246,
  author={Liu, Genjia and Hu, Yue and Xu, Chenxin and Mao, Weibo and Ge, Junhao and Huang, Zhengxiang and Lu, Yifan and Xu, Yinda and Xia, Junkai and Wang, Yafei and Chen, Siheng},
  journal={IEEE Transactions on Pattern Analysis and Machine Intelligence},
  title={{T}oward {C}ollaborative {A}utonomous {D}riving: {S}imulation {P}latform and {E}nd-to-{E}nd {S}ystem},
  year={2025},
  volume={47},
  number={8},
  pages={6566--6584},
  doi={10.1109/TPAMI.2025.3560327}
}

@article{liuEcoFriendlyOnRampMerging2023,
  title = {{E}co-{{{F}riendly {O}n-{R}amp {M}erging {S}trategy}} for {{{C}onnected}} and {{{A}utomated {V}ehicles}} in {{{H}eterogeneous {T}raffic}}},
  author = {Liu, Jinqiang and Zhao, Wanzhong and Wang, Chunyan and Xu, Can and Li, Lin and Chen, Qingyun and Lian, Yubo},
  year = {2023},
  journal = {IEEE Transactions on Vehicular Technology},
  pages = {1--13},
  issn = {0018-9545, 1939-9359},
  doi = {10.1109/TVT.2023.3281367},
  urldate = {2024-03-18},
  langid = {english}
}

@inproceedings{liuMISOMisbehaviorDetection2021,
  title = {{{{M}{I}{S}{O}- {V}}}: {{{M}isbehavior {D}etection}} for {{{C}ollective {P}erception {S}ervices}} in {{{V}ehicular {C}ommunications}}},
  shorttitle = {{{MISO- V}}},
  booktitle = {2021 {{IEEE Intelligent Vehicles Symposium}} ({{IV}})},
  author = {Liu, Xiruo and Yang, Lily and Alvarez, Ignacio and Sivanesan, Kathiravetpillai and Merwaday, Arvind and Oboril, Fabian and Buerkle, Cornelius and Sastry, Manoj and Baltar, Leonardo Gomes},
  year = {2021},
  month = jul,
  pages = {369--376},
  doi = {10.1109/IV48863.2021.9575970},
  urldate = {2024-08-07}
}

@inproceedings{liUsAdversariallyRobust2023,
  title = {{A}mong {{{U}s}}: {{{A}dversarially {R}obust {C}ollaborative {P}erception}} by {{{C}onsensus}}},
  shorttitle = {Among {{Us}}},
  booktitle = {2023 {{IEEE}}/{{CVF International Conference}} on {{Computer Vision}} ({{ICCV}})},
  author = {Li, Yiming and Fang, Qi and Bai, Jiamu and Chen, Siheng and {Juefei-Xu}, Felix and Feng, Chen},
  year = {2023},
  month = oct,
  pages = {186--195},
  publisher = {IEEE},
  address = {Paris, France},
  doi = {10.1109/ICCV51070.2023.00024},
  urldate = {2024-07-03},
  copyright = {https://doi.org/10.15223/policy-029},
  isbn = {979-8-3503-0718-4},
  langid = {english}
}

@article{liuVehicletoeverythingAutonomousDriving2023,
  title = {{T}owards {{{V}ehicle-to-everything {A}utonomous {D}riving}}: {{{A} {S}urvey}} on {{{C}ollaborative {P}erception}}},
  shorttitle = {Towards {{Vehicle-to-everything Autonomous Driving}}},
  author = {Liu, Si and Gao, Chen and Chen, Yuan and Peng, Xingyu and Kong, Xianghao and Wang, Kun and Xu, Runsheng and Jiang, Wentao and Xiang, Hao and Ma, Jiaqi and Wang, Miao},
  year = {2023},
  month = aug,
  eprint = {2308.16714},
  primaryclass = {cs},
  publisher = {arXiv},
  urldate = {2023-11-20},
  archiveprefix = {arXiv},
  journal = {arXiv preprint},
}

@inproceedings{liuWhen2comMultiAgentPerception2020,
  title = {{W}hen2com: {{{M}ulti-{A}gent {P}erception}} via {{{C}ommunication {G}raph {G}rouping}}},
  shorttitle = {When2com},
  booktitle = {2020 {{IEEE}}/{{CVF Conference}} on {{Computer Vision}} and {{Pattern Recognition}} ({{CVPR}})},
  author = {Liu, Yen-Cheng and Tian, Junjiao and Glaser, Nathaniel and Kira, Zsolt},
  year = {2020},
  month = jun,
  pages = {4105--4114},
  publisher = {IEEE},
  address = {Seattle, WA, USA},
  doi = {10.1109/CVPR42600.2020.00416},
  urldate = {2023-08-11},
  isbn = {978-1-7281-7168-5},
  langid = {english}
}

@inproceedings{liuWho2comCollaborativePerception2020,
  title = {{W}ho2com: {{{C}ollaborative {P}erception}} via {{{L}earnable {H}andshake {C}ommunication}}},
  shorttitle = {Who2com},
  booktitle = {2020 {{IEEE International Conference}} on {{Robotics}} and {{Automation}} ({{ICRA}})},
  author = {Liu, Yen-Cheng and Tian, Junjiao and Ma, Chih-Yao and Glaser, Nathan and Kuo, Chia-Wen and Kira, Zsolt},
  year = {2020},
  month = may,
  publisher = {IEEE},
  address = {Paris, France},
  doi = {10.1109/icra40945.2020.9197364},
  urldate = {2024-09-17},
  copyright = {https://ieeexplore.ieee.org/Xplorehelp/downloads/license-information/IEEE.html},
  langid = {english}
}

@inproceedings{luExtensibleFrameworkOpen2024,
  title={{A}n {{{E}xtensible {F}ramework}} for {{{O}pen {H}eterogeneous {C}ollaborative {P}erception}}},
  author={Lu, Yifan and Hu, Yue and Zhong, Yiqi and Wang, Dequan and Wang, Yanfeng and Chen, Siheng},
  booktitle={International Conference on Learning Representations (ICLR)},
  year={2024},
}

@article{luoEdgeCooperNetworkAwareCooperative2023,
  title = {{{{E}dge{C}ooper}}: {{{N}etwork-{A}ware {C}ooperative {L}i{D}{A}{R} {P}erception}} for {{{E}nhanced {V}ehicular {A}wareness}}},
  shorttitle = {{{EdgeCooper}}},
  author = {Luo, Guiyang and Shao, Chongzhang and Cheng, Nan and Zhou, Haibo and Zhang, Hui and Yuan, Quan and Li, Jinglin},
  year = {2023},
  journal = {IEEE Journal on Selected Areas in Communications},
  pages = {1--1},
  issn = {0733-8716, 1558-0008},
  doi = {10.1109/JSAC.2023.3322764},
  urldate = {2023-10-16},
  langid = {english}
}

@article{malikCollaborativeAutonomousDriving2021,
  title = {{C}ollaborative {{{A}utonomous {D}riving}}---{{{A} {S}urvey}} of {{{S}olution {A}pproaches}} and {{{F}uture {C}hallenges}}},
  author = {Malik, Sumbal and Khan, Manzoor Ahmed and {El-Sayed}, Hesham},
  year = {2021},
  month = jan,
  journal = {Sensors},
  volume = {21},
  number = {11},
  pages = {3783},
  publisher = {Multidisciplinary Digital Publishing Institute},
  issn = {1424-8220},
  doi = {10.3390/s21113783},
  urldate = {2024-03-24},
  copyright = {http://creativecommons.org/licenses/by/3.0/},
  langid = {english}
}

@inproceedings{maoLanguageAgentAutonomous2023,
  title = {{A} {{{L}anguage {A}gent}} for {{{A}utonomous {D}riving}}},
  author = {Mao, Jiageng and Ye, Junjie and Qian, Yuxi and Pavone, Marco and Wang, Yue},
  booktitle = {First Conference on Language Modeling (COLM)},
  year = {2024},
}

@article{noor-a-rahim6GVehicletoEverythingV2X2022,
  title = {{{6{G}}} for {{{V}ehicle-to-{E}verything}} ({{{V}2{X}}}) {{{C}ommunications}}: {{{E}nabling {T}echnologies}}, {{{C}hallenges}}, and {{{O}pportunities}}},
  shorttitle = {{{6G}} for {{Vehicle-to-Everything}} ({{V2X}}) {{Communications}}},
  author = {{Noor-A-Rahim}, {\relax Md}. and Liu, Zilong and Lee, Haeyoung and Khyam, Mohammad Omar and He, Jianhua and Pesch, Dirk and Moessner, Klaus and Saad, Walid and Poor, H. Vincent},
  year = {2022},
  month = jun,
  journal = {Proceedings of the IEEE},
  volume = {110},
  number = {6},
  pages = {712--734},
  issn = {0018-9219, 1558-2256},
  doi = {10.1109/JPROC.2022.3173031},
  urldate = {2022-10-12},
  langid = {english}
}

@article{shalev-shwartzSafeMultiAgentReinforcement2016,
  title = {{S}afe, {{{M}ulti-{A}gent}}, {{{R}einforcement {L}earning}} for {{{A}utonomous {D}riving}}},
  author = {{Shalev-Shwartz}, Shai and Shammah, Shaked and Shashua, Amnon},
  year = {2016},
  month = oct,
  eprint = {1610.03295},
  primaryclass = {cs, stat},
  publisher = {arXiv},
  urldate = {2023-12-23},
  archiveprefix = {arXiv},
  journal = {arXiv preprint},
}

@inproceedings{shaoLMDriveClosedLoopEndtoEnd2023,
  title={{{{L}{M}{D}rive}}: {{{C}losed-{L}oop {E}nd-to-{E}nd {D}riving}} with {{{L}arge {L}anguage {M}odels}}},
  author={Shao, Hao and Hu, Yuxuan and Wang, Letian and Waslander, Steven L. and Liu, Yu and Li, Hongsheng},
  booktitle={Proceedings of the IEEE/CVF Conference on Computer Vision and Pattern Recognition (CVPR)},
  year={2024},
}

@inproceedings{shiVIPSRealtimePerception2022,
  author = {Shi, Shuyao and Cui, Jiahe and Jiang, Zhehao and Yan, Zhenyu and Xing, Guoliang and Niu, Jianwei and Ouyang, Zhenchao},
  title = {{{V}{I}{P}{S}}: {R}eal-{T}ime {P}erception {F}usion for {I}nfrastructure-{A}ssisted {A}utonomous {D}riving},
  year = {2022},
  isbn = {9781450391818},
  publisher = {Association for Computing Machinery},
  address = {New York, NY, USA},
  doi = {10.1145/3495243.3560539},
  booktitle = {Proceedings of the 28th Annual International Conference on Mobile Computing and Networking},
  pages = {133--146},
  numpages = {14},
  location = {Sydney, NSW, Australia},
  series = {MobiCom '22}
}

@inproceedings{simaDriveLMDrivingGraph2023,
  author = {Sima, Chonghao and Renz, Katrin and Chitta, Kashyap and Chen, Li and Zhang, Hanxue and Xie, Chengen and Bei{\ss}wenger, Jens and Luo, Ping and Geiger, Andreas and Li, Hongyang},
  title = {{{D}rive{L}{M}}: {D}riving with {G}raph {V}isual {Q}uestion {A}nswering},
  booktitle = {Computer Vision - ECCV 2024: 18th European Conference, Milan, Italy, September 29-October 4, 2024, Proceedings, Part LII},
  pages = {256--274},
  year = {2024},
  publisher = {Springer},
  address = {Berlin, Heidelberg},
  doi = {10.1007/978-3-031-72943-0_15},
}

@inproceedings{tianDriveVLMConvergenceAutonomous2024,
  title={{{{D}rive{V}{L}{M}}}: {{{T}he {C}onvergence}} of {{{A}utonomous {D}riving}} and {{{L}arge {V}ision-{L}anguage {M}odels}}},
  author={Tian, Xiaoyu and Gu, Junru and Li, Bailin and Liu, Yicheng and Hu, Chenxu and Wang, Yang and Zhan, Kun and Jia, Peng and Lang, Xianpeng and Zhao, Hang},
  booktitle={Proceedings of the Conference on Robot Learning (CoRL)},
  year={2024},
}

@inproceedings{wangDrivingFutureMultiview2023,
  title = {{D}riving {I}nto the {F}uture: {M}ultiview {V}isual {F}orecasting and {P}lanning with {W}orld {M}odel for {A}utonomous {D}riving},
  author = {Wang, Yuqi and He, Jiawei and Fan, Lue and Li, Hongxin and Chen, Yuntao and Zhang, Zhaoxiang},
  booktitle = {Proceedings of the IEEE/CVF Conference on Computer Vision and Pattern Recognition},
  pages = {14749--14759},
  year = {2024},
  doi = {10.1109/CVPR52733.2024.01397}
}

@inproceedings{wangOmniDriveHolisticLLMAgent2024,
  title={{{{O}mni{D}rive}}: {{{A} {H}olistic {L}{L}{M}-{A}gent {F}ramework}} for {{{A}utonomous {D}riving}} with {{3{D} {P}erception}}, {{{R}easoning}} and {{{P}lanning}}},
  author={Wang, Shihao and Yu, Zhiding and Jiang, Xiaohui and Lan, Shiyi and Shi, Min and Chang, Nadine and Kautz, Jan and Li, Ying and Alvarez, Jose M.},
  booktitle={Proceedings of the IEEE/CVF Conference on Computer Vision and Pattern Recognition (CVPR)},
  year={2025},
}

@inproceedings{wangV2VNetVehicletoVehicleCommunication2020a,
  title = {{{{V}2{V}{N}et}}: {{{V}ehicle-to-{V}ehicle {C}ommunication}} for {{{J}oint {P}erception}} and {{{P}rediction}}},
  shorttitle = {{{V2VNet}}},
  booktitle = {Computer {{Vision}} -- {{ECCV}} 2020},
  series = {Lecture Notes in Computer Science},
  volume = {12347},
  author = {Wang, Tsun-Hsuan and Manivasagam, Sivabalan and Liang, Ming and Yang, Bin and Zeng, Wenyuan and Urtasun, Raquel},
  editor = {Vedaldi, Andrea and Bischof, Horst and Brox, Thomas and Frahm, Jan-Michael},
  year = {2020},
  pages = {605--621},
  publisher = {Springer International Publishing},
  address = {Cham},
  doi = {10.1007/978-3-030-58536-5_36},
  isbn = {978-3-030-58536-5},
  langid = {english}
}

@article{weiAsynchronyRobustCollaborativePerception2023b,
  title = {{A}synchrony-{{{R}obust {C}ollaborative {P}erception}} via {{{B}ird}}'{S} {{{E}ye {V}iew {F}low}}},
  author = {Wei, Sizhe and Wei, Yuxi and Hu, Yue and Lu, Yifan and Zhong, Yiqi and Chen, Siheng and Zhang, Ya},
  year = {2023},
  journal = {NeurIPS},
  langid = {english}
}

@inproceedings{wenDiLuKnowledgeDrivenApproach2023,
  title = {{{{D}i{L}u}}: {{{A} {K}nowledge-{D}riven {A}pproach}} to {{{A}utonomous {D}riving}} with {{{L}arge {L}anguage {M}odels}}},
  shorttitle = {{{DiLu}}},
  booktitle = {{{ICLR}}'24},
  author = {Wen, Licheng and Fu, Daocheng and Li, Xin and Cai, Xinyu and Ma, Tao and Cai, Pinlong and Dou, Min and Shi, Botian and He, Liang and Qiao, Yu},
  year = {2023},
  month = oct,
  publisher = {arXiv},
  urldate = {2023-12-04}
}

@inproceedings{xiangHMViTHeteroModalVehicletoVehicle,
  title = {{{{H}{M}-{V}i{T}}}: {{{H}etero-{M}odal {V}ehicle-to-{V}ehicle {C}ooperative {P}erception}} with {{{V}ision {T}ransformer}}},
  author = {Xiang, Hao and Xu, Runsheng and Ma, Jiaqi},
  booktitle = {Proceedings of the IEEE/CVF International Conference on Computer Vision},
  pages = {284--295},
  year = {2023},
  langid = {english}
}

@inproceedings{xuCoBEVTCooperativeBird2023,
  title = {{{{C}o{B}{E}{V}{T}}}: {{{C}ooperative {B}ird}}'{S} {{{E}ye {V}iew {S}emantic {S}egmentation}} with {{{S}parse {T}ransformers}}},
  shorttitle = {{{CoBEVT}}},
  booktitle = {Proceedings of {{The}} 6th {{Conference}} on {{Robot Learning}}},
  author = {Xu, Runsheng and Tu, Zhengzhong and Xiang, Hao and Shao, Wei and Zhou, Bolei and Ma, Jiaqi},
  year = {2023},
  month = mar,
  pages = {989--1000},
  publisher = {PMLR},
  issn = {2640-3498},
  urldate = {2024-06-03},
  langid = {english}
}

@article{xuDriveGPT4InterpretableEndtoend2023,
  title={{{{D}rive{G}{P}{T}4}}: {{{I}nterpretable {E}nd-to-end {A}utonomous {D}riving}} via {{{L}arge {L}anguage {M}odel}}},
  author={Xu, Zhenhua and Zhang, Yujia and Xie, Enze and Zhao, Zhen and Guo, Yong and Wong, Kwan-Yee K. and Li, Zhenguo and Zhao, Hengshuang},
  journal={IEEE Robotics and Automation Letters},
  year={2024},
}

@inproceedings{xuOpenCDAOpenCooperative2021,
  title={{{{O}pen{C}{D}{A}}}:{{{A}n {O}pen {C}ooperative {D}riving {A}utomation {F}ramework {I}ntegrated}} with {{{C}o-{S}imulation}}},
  author={Xu, Runsheng and Guo, Yi and Han, Xu and Xia, Xin and Xiang, Hao and Ma, Jiaqi},
  booktitle={IEEE International Intelligent Transportation Systems Conference (ITSC)},
  year={2021},
}

@inproceedings{xuOPV2VOpenBenchmark2022,
  title = {{{{O}{P}{V}2{V}}}: {{{A}n {O}pen {B}enchmark {D}ataset}} and {{{F}usion {P}ipeline}} for {{{P}erception}} with {{{V}ehicle-to-{V}ehicle {C}ommunication}}},
  shorttitle = {{{OPV2V}}},
  booktitle = {2022 {{International Conference}} on {{Robotics}} and {{Automation}} ({{ICRA}})},
  author = {Xu, Runsheng and Xiang, Hao and Xia, Xin and Han, Xu and Li, Jinlong and Ma, Jiaqi},
  year = {2022},
  month = may,
  pages = {2583--2589},
  publisher = {IEEE},
  address = {Philadelphia, PA, USA},
  doi = {10.1109/ICRA46639.2022.9812038},
  urldate = {2024-05-30},
  copyright = {https://doi.org/10.15223/policy-029},
  isbn = {978-1-7281-9681-7},
  langid = {english}
}

@inproceedings{xuV2V4RealRealWorldLargeScale2023,
  title = {{{{V}2{V}4{R}eal}}: {{{A} {R}eal-{W}orld {L}arge-{S}cale {D}ataset}} for {{{V}ehicle-to-{V}ehicle {C}ooperative {P}erception}}},
  shorttitle = {{{V2V4Real}}},
  booktitle = {2023 {{IEEE}}/{{CVF Conference}} on {{Computer Vision}} and {{Pattern Recognition}} ({{CVPR}})},
  author = {Xu, Runsheng and Xia, Xin and Li, Jinlong and Li, Hanzhao and Zhang, Shuo and Tu, Zhengzhong and Meng, Zonglin and Xiang, Hao and Dong, Xiaoyu and Song, Rui and Yu, Hongkai and Zhou, Bolei and Ma, Jiaqi},
  year = {2023},
  month = jun,
  publisher = {IEEE},
  address = {Vancouver, BC, Canada},
  doi = {10.1109/cvpr52729.2023.01318},
  urldate = {2024-09-17},
  copyright = {https://doi.org/10.15223/policy-029},
  langid = {english}
}

@inproceedings{xuV2XViTVehicletoEverythingCooperative2022a,
  title = {{{{V}2{X}-{V}i{T}}}: {{{V}ehicle-to-{E}verything {C}ooperative {P}erception}} with~{{{V}ision {T}ransformer}}},
  shorttitle = {{{V2X-ViT}}},
  booktitle = {Computer {{Vision}} -- {{ECCV}} 2022},
  author = {Xu, Runsheng and Xiang, Hao and Tu, Zhengzhong and Xia, Xin and Yang, Ming-Hsuan and Ma, Jiaqi},
  editor = {Avidan, Shai and Brostow, Gabriel and Ciss{\'e}, Moustapha and Farinella, Giovanni Maria and Hassner, Tal},
  year = {2022},
  pages = {107--124},
  publisher = {Springer Nature Switzerland},
  address = {Cham},
  doi = {10.1007/978-3-031-19842-7_7},
  isbn = {978-3-031-19842-7},
  langid = {english}
}

@article{yangLLM4DriveSurveyLarge2023,
  title = {{{{L}{L}{M}4{D}rive}}: {{{A} {S}urvey}} of {{{L}arge {L}anguage {M}odels}} for {{{A}utonomous {D}riving}}},
  shorttitle = {{{LLM4Drive}}},
  author = {Yang, Zhenjie and Jia, Xiaosong and Li, Hongyang and Yan, Junchi},
  year = {2023},
  month = nov,
  eprint = {2311.01043},
  primaryclass = {cs},
  publisher = {arXiv},
  urldate = {2023-12-04},
  archiveprefix = {arXiv},
  journal = {arXiv preprint},
}

@inproceedings{yangWhat2commCommunicationefficientCollaborative2023,
  title = {{W}hat2comm: {{{T}owards {C}ommunication-efficient {C}ollaborative {P}erception}} via {{{F}eature {D}ecoupling}}},
  shorttitle = {What2comm},
  booktitle = {Proceedings of the 31st {{ACM International Conference}} on {{Multimedia}}},
  author = {Yang, Kun and Yang, Dingkang and Zhang, Jingyu and Wang, Hanqi and Sun, Peng and Song, Liang},
  year = {2023},
  month = oct,
  volume = {29},
  pages = {7686--7695},
  publisher = {ACM},
  address = {Ottawa ON Canada},
  doi = {10.1145/3581783.3611699},
  urldate = {2024-09-17}
}

@article{yinV2VFormerMultiModalVehicletoVehicle2023,
  title = {{{{V}2{V}{F}ormer}}++: {{{M}ulti-{M}odal {V}ehicle-to-{V}ehicle {C}ooperative {P}erception}} via {{{G}lobal-{L}ocal {T}ransformer}}},
  shorttitle = {{{V2VFormer}}\$++\$},
  author = {Yin, Hongbo and Tian, Daxin and Lin, Chunmian and Duan, Xuting and Zhou, Jianshan and Zhao, Dezong and Cao, Dongpu},
  year = {2023},
  journal = {IEEE Transactions on Intelligent Transportation Systems},
  pages = {1--14},
  issn = {1524-9050, 1558-0016},
  doi = {10.1109/TITS.2023.3314919},
  urldate = {2023-10-10},
  langid = {english}
}

@article{youV2XVLMEndtoEndV2X2024,
  title={{{{V}2{X}-{V}{L}{M}}}: {{{E}nd-to-{E}nd {V}2{X} {C}ooperative {A}utonomous {D}riving {T}hrough {L}arge {V}ision-{L}anguage {M}odels}}},
  author={You, Junwei and Shi, Haotian and Jiang, Zhuoyu and Huang, Zilin and Gan, Rui and Wu, Keshu and Cheng, Xi and Li, Xiaopeng and Ran, Bin},
  journal={Transportation Research Part C: Emerging Technologies},
  year={2025},
}

@inproceedings{yuDAIRV2XLargeScaleDataset2022,
  title = {{{{D}{A}{I}{R}-{V}2{X}}}: {{{A} {L}arge-{S}cale {D}ataset}} for {{{V}ehicle-{I}nfrastructure {C}ooperative 3{D} {O}bject {D}etection}}},
  shorttitle = {{{DAIR-V2X}}},
  booktitle = {2022 {{IEEE}}/{{CVF Conference}} on {{Computer Vision}} and {{Pattern Recognition}} ({{CVPR}})},
  author = {Yu, Haibao and Luo, Yizhen and Shu, Mao and Huo, Yiyi and Yang, Zebang and Shi, Yifeng and Guo, Zhenglong and Li, Hanyu and Hu, Xing and Yuan, Jirui and Nie, Zaiqing},
  year = {2022},
  month = jun,
  publisher = {IEEE},
  address = {New Orleans, LA, USA},
  doi = {10.1109/cvpr52688.2022.02067},
  urldate = {2024-09-17},
  copyright = {https://doi.org/10.15223/policy-029},
  langid = {english}
}

@inproceedings{yuEndtoEndAutonomousDriving2024,
  title = {{E}nd-to-{E}nd {A}utonomous {D}riving {T}hrough {V}2{X} {C}ooperation},
  author = {Yu, Haibao and Yang, Wenxian and Zhong, Jiaru and Yang, Zhenwei and Fan, Siqi and Luo, Ping and Nie, Zaiqing},
  booktitle = {Proceedings of the AAAI Conference on Artificial Intelligence},
  volume = {39},
  number = {9},
  pages = {9598--9606},
  year = {2025},
  doi = {10.1609/aaai.v39i9.33040},
}

@article{yuRobustCommunicativeMultiAgent2024,
  title = {{R}obust {{{C}ommunicative {M}ulti-{A}gent {R}einforcement {L}earning}} with {{{A}ctive {D}efense}}},
  author = {Yu, Lebin and Qiu, Yunbo and Yao, Quanming and Shen, Yuan and Zhang, Xudong and Wang, Jian},
  year = {2024},
  month = mar,
  journal = {Proceedings of the AAAI Conference on Artificial Intelligence},
  volume = {38},
  number = {16},
  pages = {17575--17582},
  issn = {2374-3468, 2159-5399},
  doi = {10.1609/aaai.v38i16.29708},
  urldate = {2024-07-07},
  langid = {english}
}

@inproceedings{zhangChatSceneKnowledgeEnabledSafetyCritical2024,
  title = {{{{C}hat{S}cene}}: {{{K}nowledge-{E}nabled {S}afety-{C}ritical {S}cenario {G}eneration}} for {{{A}utonomous {V}ehicles}}},
  shorttitle = {{{ChatScene}}},
  booktitle = {2024 {{IEEE}}/{{CVF Conference}} on {{Computer Vision}} and {{Pattern Recognition}} ({{CVPR}})},
  author = {Zhang, Jiawei and Xu, Chejian and Li, Bo},
  year = {2024},
  month = jun,
  pages = {15459--15469},
  publisher = {IEEE},
  address = {Seattle, WA, USA},
  doi = {10.1109/CVPR52733.2024.01464},
  urldate = {2024-10-25},
  copyright = {https://doi.org/10.15223/policy-029},
  isbn = {979-8-3503-5300-6},
  langid = {english}
}

@inproceedings{zhangCooperativePerceptionSafe2023,
  title={{C}ooperative {{{P}erception}} for {{{S}afe {C}ontrol}} of {{{A}utonomous {V}ehicles}} {U}nder {{{L}i{D}{A}{R} {S}poofing {A}ttacks}}},
  author={Zhang, Hongchao and Li, Zhouchi and Cheng, Shiyu and Clark, Andrew},
  booktitle={Symposium on Vehicles Security and Privacy (VehicleSec), NDSS},
  year={2023},
}

@inproceedings{zhangDataFabricationCollaborative2023,
  title={{O}n {{{D}ata {F}abrication}} in {{{C}ollaborative {V}ehicular {P}erception}}: {{{A}ttacks}} and {{{C}ountermeasures}}},
  author={Zhang, Qingzhao and Jin, Shuowei and Zhu, Ruiyang and Sun, Jiachen and Zhang, Xumiao and Chen, Qi Alfred and Mao, Z. Morley},
  booktitle={USENIX Security Symposium},
  year={2024},
}

@inproceedings{zhangEMPEdgeassistedMultivehicle2021,
  title = {{{{E}{M}{P}}}: {E}dge-{A}ssisted {M}ulti-{V}ehicle {P}erception},
  shorttitle = {{{EMP}}},
  booktitle = {Proceedings of the 27th {{Annual International Conference}} on {{Mobile Computing}} and {{Networking}}},
  author = {Zhang, Xumiao and Zhang, Anlan and Sun, Jiachen and Zhu, Xiao and Guo, Y. Ethan and Qian, Feng and Mao, Z. Morley},
  year = {2021},
  month = oct,
  pages = {545--558},
  publisher = {ACM},
  address = {New Orleans Louisiana},
  doi = {10.1145/3447993.3483242},
  urldate = {2023-11-28},
  isbn = {978-1-4503-8342-4},
  langid = {english}
}

@inproceedings{zhangERMVPCommunicationEfficientCollaborationRobust,
  title = {{{{E}{R}{M}{V}{P}}}: {{{C}ommunication-{E}fficient}} and {{{C}ollaboration-{R}obust {M}ulti-{V}ehicle {P}erception}} in {{{C}hallenging {E}nvironments}}},
  author = {Zhang, Jingyu and Yang, Kun and Wang, Yilei and Wang, Hanqi and Sun, Peng and Song, Liang},
  booktitle = {Proceedings of the IEEE/CVF Conference on Computer Vision and Pattern Recognition},
  pages = {12575--12584},
  year = {2024}
}

@inproceedings{zhengGenADGenerativeEndtoEnd2024,
  title={{{{G}en{A}{D}}}: {{{G}enerative {E}nd-to-{E}nd {A}utonomous {D}riving}}},
  author={Zheng, Wenzhao and Song, Ruiqi and Guo, Xianda and Chen, Long},
  booktitle={Proceedings of the European Conference on Computer Vision (ECCV)},
  year={2024},
}

@article{zhengPlanAgentMultimodalLarge2024,
  title={{P}lanagent: {A} {M}ulti-{M}odal {L}arge {L}anguage {A}gent for {C}losed-{L}oop {V}ehicle {M}otion {P}lanning},
  author={Zheng, Yupeng and Xing, Zebin and Zhang, Qichao and Jin, Bu and Li, Pengfei and Zheng, Yuhang and Xia, Zhongpu and Chen, Yaran and Zhao, Dongbin},
  journal={IEEE Transactions on Cognitive and Developmental Systems},
  year={2026},
  publisher={IEEE}
}

@ARTICLE{9316925,
  author={Liu, Jinqiang and Zhao, Wanzhong and Xu, Can},
  journal={IEEE Transactions on Intelligent Transportation Systems}, 
  title={{{{A}}}n {{{E}}}fficient {{{O}}}n-{{{R}}}amp {{{M}}}erging {{{S}}}trategy for {{{C}}}onnected and {{{A}}}utomated {{{V}}}ehicles in {{{M}}}ulti-{{{L}}}ane {{{T}}}raffic}, 
  year={2022},
  volume={23},
  number={6},
  pages={5056-5067},
  doi={10.1109/TITS.2020.3046643}}

@article{V2Xsim2022,
  title={{V}2{X}-{S}im: {M}ulti-{A}gent {C}ollaborative {P}erception {D}ataset and {B}enchmark for {A}utonomous {D}riving},
  author={Li, Yiming and Ma, Dekun and An, Ziyan and Wang, Zixun and Zhong, Yiqi and Chen, Siheng and Feng, Chen},
  journal={IEEE Robotics and Automation Letters},
  volume={7},
  number={4},
  pages={10914--10921},
  year={2022},
  publisher={IEEE}
}

@inproceedings{ahmed2022joint,
  title={{A} {J}oint {P}erception {S}cheme for {C}onnected {V}ehicles},
  author={Ahmed, Ahmed N and Ravijts, Ian and de Hoog, Jens and Anwar, Ali and Mercelis, Siegfried and Hellinckx, Peter},
  booktitle={2022 IEEE Sensors},
  pages={1--4},
  year={2022},
  organization={IEEE}
}

@inproceedings{luo2022complementarity,
  title={{C}omplementarity-{E}nhanced and {R}edundancy-{M}inimized {C}ollaboration {N}etwork for {M}ulti-{A}gent {P}erception},
  author={Luo, Guiyang and Zhang, Hui and Yuan, Quan and Li, Jinglin},
  booktitle={Proceedings of the 30th ACM International Conference on Multimedia},
  pages={3578--3586},
  year={2022}
}

@inproceedings{wang2023core,
  title={{C}ore: {C}ooperative {R}econstruction for {M}ulti-{A}gent {P}erception},
  author={Wang, Binglu and Zhang, Lei and Wang, Zhaozhong and Zhao, Yongqiang and Zhou, Tianfei},
  booktitle={Proceedings of the IEEE/CVF International Conference on Computer Vision},
  pages={8710--8720},
  year={2023}
}

@article{dao2024practical,
  title={{P}ractical {C}ollaborative {P}erception: {A} {F}ramework for {A}synchronous and {M}ulti-{A}gent 3{D} {O}bject {D}etection},
  author={Dao, Minh-Quan and Berrio, Julie Stephany and Fr{\'e}mont, Vincent and Shan, Mao and H{\'e}ry, Elwan and Worrall, Stewart},
  journal={IEEE Transactions on Intelligent Transportation Systems},
  year={2024},
  publisher={IEEE}
}

@inproceedings{hong2024multi,
  title={{M}ulti-{A}gent {C}ollaborative {P}erception via {M}otion-{A}ware {R}obust {C}ommunication {N}etwork},
  author={Hong, Shixin and Liu, Yu and Li, Zhi and Li, Shaohui and He, You},
  booktitle={Proceedings of the IEEE/CVF Conference on Computer Vision and Pattern Recognition},
  pages={15301--15310},
  year={2024}
}

@article{wang2025cmp,
  title={{C}mp: {C}ooperative {M}otion {P}rediction with {M}ulti-{A}gent {C}ommunication},
  author={Wang, Zehao and Wang, Yuping and Wu, Zhuoyuan and Ma, Hengbo and Li, Zhaowei and Qiu, Hang and Li, Jiachen},
  journal={IEEE Robotics and Automation Letters},
  year={2025},
  publisher={IEEE}
}

@article{arnold2020cooperative,
  title={{C}ooperative {P}erception for 3{D} {O}bject {D}etection in {D}riving {S}cenarios {U}sing {I}nfrastructure {S}ensors},
  author={Arnold, Eduardo and Dianati, Mehrdad and de Temple, Robert and Fallah, Saber},
  journal={IEEE Transactions on Intelligent Transportation Systems},
  volume={23},
  number={3},
  pages={1852--1864},
  year={2020},
  publisher={IEEE}
}

@inproceedings{xu2023model,
  title={{M}odel-{A}gnostic {M}ulti-{A}gent {P}erception {F}ramework},
  author={Xu, Runsheng and Chen, Weizhe and Xiang, Hao and Xia, Xin and Liu, Lantao and Ma, Jiaqi},
  booktitle={2023 IEEE International Conference on Robotics and Automation (ICRA)},
  pages={1471--1478},
  year={2023},
  organization={IEEE}
}

@article{glaser2023we,
  title={{W}e {N}eed to {T}alk: {I}dentifying and {O}vercoming {C}ommunication-{C}ritical {S}cenarios for {S}elf-{D}riving},
  author={Glaser, Nathaniel Moore and Kira, Zsolt},
  journal={arXiv preprint},
  year={2023},
  eprint={2305.04352},
  archivePrefix={arXiv},
  doi={10.48550/arXiv.2305.04352},
}

@article{huangVehicleEverythingCooperative2025,
  title={{V}ehicle-to-{E}verything {C}ooperative {P}erception for {A}utonomous {D}riving},
  author={Huang, Tao and Liu, Jianan and Zhou, Xi and Nguyen, Dinh C. and Azghadi, Mostafa Rahimi and Xia, Yuxuan and Han, Qing-Long and Sun, Sumei},
  journal={Proceedings of the IEEE},
  volume={113},
  number={5},
  pages={443--477},
  year={2025},
  doi={10.1109/JPROC.2025.3600903},
}

@article{bejarbanehExploringSharedPerception2024,
  title={{E}xploring {S}hared {P}erception and {C}ontrol in {C}ooperative {V}ehicle-{I}ntersection {S}ystems: {A} {R}eview},
  author={Bejarbaneh, Elham Yazdani and Du, Haiping and Naghdy, Fazel},
  journal={IEEE Transactions on Intelligent Transportation Systems},
  volume={25},
  number={11},
  pages={15247--15272},
  year={2024},
  doi={10.1109/TITS.2024.3432634},
}

@article{gao2024survey,
  title={{A} {S}urvey of {C}ollaborative {P}erception in {I}ntelligent {V}ehicles at {I}ntersections},
  author={Gao, Xin and Zhang, Xinyu and Lu, Yiguo and Huang, Yuning and Yang, Lei and Xiong, Yijin and Liu, Peng},
  journal={IEEE Transactions on Intelligent Vehicles},
  pages={1--20},
  year={2024},
  doi={10.1109/TIV.2024.3395783},
}

@article{tsukadaNetworkedRoadsidePerception2020,
  title={{N}etworked {R}oadside {P}erception {U}nits for {A}utonomous {D}riving},
  author={Tsukada, Manabu and Oi, Takaharu and Kitazawa, Masahiro and Esaki, Hiroshi},
  journal={Sensors},
  volume={20},
  number={18},
  pages={5320},
  year={2020},
  doi={10.3390/s20185320}
}

@inproceedings{fadiliLateCollaborativePerception2025,
  title={{A} {L}ate {C}ollaborative {P}erception {F}ramework for 3{D} {M}ulti-{O}bject and {M}ulti-{S}ource {A}ssociation and {F}usion},
  author={Fadili, Maryem and Ghaoui, Mohamed Anis and Lecrosnier, Louis and Pechberti, Steve and Khemmar, Redouane},
  booktitle={2025 9th International Conference on Robotics and Automation Sciences (ICRAS)},
  pages={259--266},
  year={2025},
  doi={10.1109/ICRAS65818.2025.11108781}
}

@article{castelinoTracktoTrackFusion2026,
  title={{T}rack-to-{T}rack {F}usion for {C}ooperative {P}erception {U}sing {C}ollective {P}erception {M}essages},
  author={Castelino, Redge Melroy and Pradhan, Shrijal and Hahn, Axel},
  journal={Sensors},
  volume={26},
  number={6},
  pages={2003},
  year={2026},
  doi={10.3390/s26062003}
}

@article{gao2024vehicle,
  title={{V}ehicle-{R}oad-{C}loud {C}ollaborative {P}erception {F}ramework and {K}ey {T}echnologies: {A} {R}eview},
  author={Gao, Bolin and Liu, Jiaxi and Zou, Hengduo and Chen, Jiaxing and He, Lei and Li, Keqiang},
  journal={IEEE Transactions on Intelligent Transportation Systems},
  year={2024},
  publisher={IEEE}
}

@article{shi2024mtr++,
  title={{M}tr++: {M}ulti-{A}gent {M}otion {P}rediction with {S}ymmetric {S}cene {M}odeling and {G}uided {I}ntention {Q}uerying},
  author={Shi, Shaoshuai and Jiang, Li and Dai, Dengxin and Schiele, Bernt},
  journal={IEEE Transactions on Pattern Analysis and Machine Intelligence},
  volume={46},
  number={5},
  pages={3955--3971},
  year={2024},
  publisher={IEEE}
}

@inproceedings{jiang2023motiondiffuser,
  title={{M}otiondiffuser: {C}ontrollable {M}ulti-{A}gent {M}otion {P}rediction {U}sing {D}iffusion},
  author={Jiang, Chiyu Max and Cornman, Andre and Park, Cheolho and Sapp, Benjamin and Zhou, Yin and Anguelov, Dragomir},
  booktitle={Proceedings of the IEEE/CVF conference on computer vision and pattern recognition},
  pages={9644--9653},
  year={2023}
}

@article{wu2024smart,
  title={{S}{M}{A}{R}{T}: {S}calable {M}ulti-{A}gent {R}eal-{T}ime {M}otion {G}eneration via {N}ext-{T}oken {P}rediction},
  author={Wu, Wei and Feng, Xiaoxin and Gao, Ziyan and Kan, Yuheng},
  journal={Advances in Neural Information Processing Systems},
  volume={37},
  pages={114048--114071},
  year={2024}
}

@inproceedings{rowe2023fjmp,
  title={{F}jmp: {F}actorized {J}oint {M}ulti-{A}gent {M}otion {P}rediction over {L}earned {D}irected {A}cyclic {I}nteraction {G}raphs},
  author={Rowe, Luke and Ethier, Martin and Dykhne, Eli-Henry and Czarnecki, Krzysztof},
  booktitle={Proceedings of the IEEE/CVF Conference on Computer Vision and Pattern Recognition},
  pages={13745--13755},
  year={2023}
}

@inproceedings{gilles2022thomas,
  title={{{T}{H}{O}{M}{A}{S}}: {T}rajectory {H}eatmap {O}utput with {L}earned {M}ulti-{A}gent {S}ampling},
  author={Gilles, Thomas and Sabatini, Stefano and Tsishkou, Dzmitry and Stanciulescu, Bogdan and Moutarde, Fabien},
  booktitle={International Conference on Learning Representations},
  year={2022},
}

@inproceedings{gu2021densetnt,
  title={{D}ensetnt: {E}nd-to-{E}nd {T}rajectory {P}rediction from {D}ense {G}oal {S}ets},
  author={Gu, Junru and Sun, Chen and Zhao, Hang},
  booktitle={Proceedings of the IEEE/CVF International Conference on Computer Vision},
  pages={15303--15312},
  year={2021}
}

@inproceedings{nayakanti2023wayformer,
  title={{W}ayformer: {M}otion {F}orecasting via {S}imple \& {E}fficient {A}ttention {N}etworks},
  author={Nayakanti, Nigamaa and Al-Rfou, Rami and Zhou, Aurick and Goel, Kratarth and Refaat, Khaled S and Sapp, Benjamin},
  booktitle={2023 IEEE International Conference on Robotics and Automation (ICRA)},
  pages={2980--2987},
  year={2023},
  organization={IEEE}
}

@inproceedings{zhou2022hivt,
  title={{H}ivt: {H}ierarchical {V}ector {T}ransformer for {M}ulti-{A}gent {M}otion {P}rediction},
  author={Zhou, Zikang and Ye, Luyao and Wang, Jianping and Wu, Kui and Lu, Kejie},
  booktitle={Proceedings of the IEEE/CVF Conference on Computer Vision and Pattern Recognition},
  pages={8823--8833},
  year={2022}
}

@article{huang2023decentralized,
  title={{D}ecentralized {I}{L}{Q}{R} for {C}ooperative {T}rajectory {P}lanning of {C}onnected {A}utonomous {V}ehicles via {D}ual {C}onsensus {A}{D}{M}{M}},
  author={Huang, Zhenmin and Shen, Shaojie and Ma, Jun},
  journal={IEEE Transactions on Intelligent Transportation Systems},
  volume={24},
  number={11},
  pages={12754--12766},
  year={2023},
  publisher={IEEE}
}

@article{liu2024improved,
  title={{I}mproved {C}onsensus {A}{D}{M}{M} for {C}ooperative {M}otion {P}lanning of {L}arge-{S}cale {C}onnected {A}utonomous {V}ehicles with {L}imited {C}ommunication},
  author={Liu, Haichao and Huang, Zhenmin and Zhu, Zicheng and Li, Yulin and Shen, Shaojie and Ma, Jun},
  journal={IEEE Transactions on Intelligent Vehicles},
  year={2024},
  publisher={IEEE}
}

@inproceedings{huang2024parallel,
  title={{P}arallel {O}ptimization with {H}ard {S}afety {C}onstraints for {C}ooperative {P}lanning of {C}onnected {A}utonomous {V}ehicles},
  author={Huang, Zhenmin and Liu, Haichao and Shen, Shaojie and Ma, Jun},
  booktitle={2024 IEEE International Conference on Robotics and Automation (ICRA)},
  pages={2238--2244},
  year={2024},
  organization={IEEE}
}

@article{pei2022fault,
  title={{F}ault-{T}olerant {C}ooperative {D}riving at {S}ignal-{F}ree {I}ntersections},
  author={Pei, Huaxin and Zhang, Jiawei and Zhang, Yi and Pei, Xin and Feng, Shuo and Li, Li},
  journal={IEEE Transactions on Intelligent Vehicles},
  volume={8},
  number={1},
  pages={121--134},
  year={2022},
  publisher={IEEE}
}

@article{wang2022multi,
  title={{M}ulti-{V}ehicle {C}onflict {M}anagement with {S}tatus and {I}ntent {S}haring {U}nder {T}ime {D}elays},
  author={Wang, Hao M and Avedisov, Sergei S and Altintas, Onur and Orosz, G{\'a}bor},
  journal={IEEE Transactions on Intelligent Vehicles},
  volume={8},
  number={2},
  pages={1624--1637},
  year={2022},
  publisher={IEEE}
}

@article{zheng2023opencda,
  title={{O}pen{C}{D}{A}-{R}{O}{S}: {E}nabling {S}eamless {I}ntegration of {S}imulation and {R}eal-{W}orld {C}ooperative {D}riving {A}utomation},
  author={Zheng, Zhaoliang and Han, Xu and Xia, Xin and Gao, Letian and Xiang, Hao and Ma, Jiaqi},
  journal={IEEE Transactions on Intelligent Vehicles},
  volume={8},
  number={7},
  pages={3775--3780},
  year={2023},
  publisher={IEEE}
}

@article{li2022cooperative,
  title={{C}ooperative {F}ormation of {A}utonomous {V}ehicles in {M}ixed {T}raffic {F}low: {B}eyond {P}latooning},
  author={Li, Keqiang and Wang, Jiawei and Zheng, Yang},
  journal={IEEE Transactions on Intelligent Transportation Systems},
  volume={23},
  number={9},
  pages={15951--15966},
  year={2022},
  publisher={IEEE}
}

@article{cai2022multi,
  title={{M}ulti-{L}ane {U}nsignalized {I}ntersection {C}ooperation with {F}lexible {L}ane {D}irection {B}ased on {M}ulti-{V}ehicle {F}ormation {C}ontrol},
  author={Cai, Mengchi and Xu, Qing and Chen, Chaoyi and Wang, Jiawei and Li, Keqiang and Wang, Jianqiang and Wu, Xiangbin},
  journal={IEEE transactions on vehicular technology},
  volume={71},
  number={6},
  pages={5787--5798},
  year={2022},
  publisher={IEEE}
}

@article{yanumula2023optimal,
  title={{O}ptimal {T}rajectory {P}lanning for {C}onnected and {A}utomated {V}ehicles in {L}ane-{F}ree {T}raffic with {V}ehicle {N}udging},
  author={Yanumula, Venkata Karteek and Typaldos, Panagiotis and Troullinos, Dimitrios and Malekzadeh, Milad and Papamichail, Ioannis and Papageorgiou, Markos},
  journal={IEEE Transactions on Intelligent Vehicles},
  volume={8},
  number={3},
  pages={2385--2399},
  year={2023},
  publisher={IEEE}
}

@inproceedings{wu2023intent,
  title={{I}ntent-{A}ware {P}lanning in {H}eterogeneous {T}raffic via {D}istributed {M}ulti-{A}gent {R}einforcement {L}earning},
  author={Wu, Xiyang and Chandra, Rohan and Guan, Tianrui and Bedi, Amrit and Manocha, Dinesh},
  booktitle={Conference on Robot Learning},
  pages={446--477},
  year={2023},
  organization={PMLR}
}

@inproceedings{dai2023socially,
  title={{S}ocially-{A}ttentive {P}olicy {O}ptimization in {M}ulti-{A}gent {S}elf-{D}riving {S}ystem},
  author={Dai, Zipeng and Zhou, Tianze and Shao, Kun and Mguni, David Henry and Wang, Bin and Hao, Jianye},
  booktitle={Conference on Robot Learning},
  pages={946--955},
  year={2023},
  organization={PMLR}
}

@inproceedings{peng2023curriculum,
  title={{C}urriculum {P}roximal {P}olicy {O}ptimization with {S}tage-{D}ecaying {C}lipping for {S}elf-{D}riving at {U}nsignalized {I}ntersections},
  author={Peng, Zengqi and Zhou, Xiao and Wang, Yubin and Zheng, Lei and Liu, Ming and Ma, Jun},
  booktitle={2023 IEEE 26th International Conference on Intelligent Transportation Systems (ITSC)},
  pages={5027--5033},
  year={2023},
  organization={IEEE}
}

@article{xiao2024decision,
  title={{D}ecision-{M}aking for {A}utonomous {V}ehicles in {R}andom {T}ask {S}cenarios at {U}nsignalized {I}ntersection {U}sing {D}eep {R}einforcement {L}earning},
  author={Xiao, Wenxuan and Yang, Yuyou and Mu, Xinyu and Xie, Yi and Tang, Xiaolin and Cao, Dongpu and Liu, Teng},
  journal={IEEE Transactions on Vehicular Technology},
  volume={73},
  number={6},
  pages={7812--7825},
  year={2024},
  publisher={IEEE}
}

@article{zhang2022multi,
  title={{M}ulti-{A}gent {D}{R}{L}-{B}ased {L}ane {C}hange with {R}ight-of-{W}ay {C}ollaboration {A}wareness},
  author={Zhang, Jiawei and Chang, Cheng and Zeng, Xianlin and Li, Li},
  journal={IEEE Transactions on Intelligent Transportation Systems},
  volume={24},
  number={1},
  pages={854--869},
  year={2022},
  publisher={IEEE}
}

@article{gutierrez2022reinforcement,
  title={{R}einforcement {L}earning-{B}ased {A}utonomous {D}riving at {I}ntersections in {C}arla {S}imulator},
  author={Guti{\'e}rrez-Moreno, Rodrigo and Barea, Rafael and L{\'o}pez-Guill{\'e}n, Elena and Araluce, Javier and Bergasa, Luis M},
  journal={Sensors},
  volume={22},
  number={21},
  pages={8373},
  year={2022},
  publisher={MDPI}
}

@inproceedings{zhang2023spatial,
  title={{S}patial-{T}emporal-{A}ware {S}afe {M}ulti-{A}gent {R}einforcement {L}earning of {C}onnected {A}utonomous {V}ehicles in {C}hallenging {S}cenarios},
  author={Zhang, Zhili and Han, Songyang and Wang, Jiangwei and Miao, Fei},
  booktitle={2023 IEEE International Conference on Robotics and Automation (ICRA)},
  pages={5574--5580},
  year={2023},
  organization={IEEE}
}

@article{liu2022graph,
  title={{G}raph {R}einforcement {L}earning-{B}ased {D}ecision-{M}aking {T}echnology for {C}onnected and {A}utonomous {V}ehicles: {F}ramework, {R}eview, and {F}uture {T}rends},
  author={Liu, Qi and Li, Xueyuan and Tang, Yujie and Gao, Xin and Yang, Fan and Li, Zirui},
  journal={Sensors},
  volume={23},
  number={19},
  pages={8229},
  year={2023},
  doi={10.3390/s23198229},
}

@article{dinnewethMultiagentReinforcementLearning2022,
  title={{M}ulti-{A}gent {R}einforcement {L}earning for {A}utonomous {V}ehicles: {A} {S}urvey},
  author={Dinneweth, Joris and Boubezoul, Abderrahmane and Mandiau, Rene and Espie, Stephane},
  journal={Autonomous Intelligent Systems},
  volume={2},
  number={1},
  year={2022},
  month={nov},
  doi={10.1007/s43684-022-00045-z},
}

@article{wang2023multi,
  title={{M}ulti-{A}gent {R}einforcement {L}earning {G}uided by {S}ignal {T}emporal {L}ogic {S}pecifications},
  author={Wang, Jiangwei and Yang, Shuo and An, Ziyan and Han, Songyang and Zhang, Zhili and Mangharam, Rahul and Ma, Meiyi and Miao, Fei},
  journal={arXiv preprint},
  year={2023},
  eprint={2306.06808},
  archivePrefix={arXiv},
  doi={10.48550/arXiv.2306.06808},
}

@inproceedings{wen2023bringing,
  title={{B}ringing {D}iversity to {A}utonomous {V}ehicles: {A}n {I}nterpretable {M}ulti-{V}ehicle {D}ecision-{M}aking and {P}lanning {F}ramework},
  author={Wen, Licheng and Cai, Pinlong and Fu, Daocheng and Mao, Song and Li, Yikang},
  booktitle={Proceedings of the International Conference on Autonomous Agents and Multiagent Systems (AAMAS)},
  year={2023},
}

@inproceedings{guo2024mappo,
  title={{M}{A}{P}{P}{O}-{P}{I}{S}: {A} {M}ulti-{A}gent {P}roximal {P}olicy {O}ptimization {M}ethod with {P}rior {I}ntent {S}haring for {C}{A}{V}s' {C}ooperative {D}ecision-{M}aking},
  author={Guo, Yicheng and Liu, Jiaqi and Yu, Rongjie and Hang, Peng and Sun, Jian},
  booktitle={Proceedings of the European Conference on Computer Vision Workshops (ECCVW)},
  year={2024}
}

@article{liu2024cooperative,
  title={{C}ooperative {D}ecision-{M}aking for {C}avs at {U}nsignalized {I}ntersections: {A} {M}arl {A}pproach with {A}ttention and {H}ierarchical {G}ame {P}riors},
  author={Liu, Jiaqi and Hang, Peng and Na, Xiaoxiang and Huang, Chao and Sun, Jian},
  journal={IEEE Transactions on Intelligent Transportation Systems},
  year={2024},
  publisher={IEEE}
}

@inproceedings{yao2024comal,
  title={{C}o{M}{A}{L}: {C}ollaborative {M}ulti-{A}gent {L}arge {L}anguage {M}odels for {M}ixed-{A}utonomy {T}raffic},
  author={Yao, Huaiyuan and Da, Longchao and Nandam, Vishnu and Turnau, Justin and Liu, Zhiwei and Pang, Linsey and Wei, Hua},
  booktitle={Proceedings of the 2025 SIAM International Conference on Data Mining (SDM)},
  pages={409--418},
  year={2025},
  doi={10.1137/1.9781611978520.43}
}

@article{yan2023multi,
  title={{A} {M}ulti-{V}ehicle {G}ame-{T}heoretic {F}ramework for {D}ecision {M}aking and {P}lanning of {A}utonomous {V}ehicles in {M}ixed {T}raffic},
  author={Yan, Yongjun and Peng, Lin and Shen, Tong and Wang, Jinxiang and Pi, Dawei and Cao, Dongpu and Yin, Guodong},
  journal={IEEE Transactions on Intelligent Vehicles},
  volume={8},
  number={11},
  pages={4572--4587},
  year={2023},
  publisher={IEEE}
}

@article{tang2022highway,
  title={{H}ighway {D}ecision-{M}aking and {M}otion {P}lanning for {A}utonomous {D}riving via {S}oft {A}ctor-{C}ritic},
  author={Tang, Xiaolin and Huang, Bing and Liu, Teng and Lin, Xianke},
  journal={IEEE Transactions on Vehicular Technology},
  volume={71},
  number={5},
  pages={4706--4717},
  year={2022},
  publisher={IEEE}
}
}

\end{document}